\pdfoutput=1
\documentclass[11pt]{article}
\usepackage{yinwang_technical_report}

\usepackage{amsmath,amsfonts,bm}

\def\eqref#1{equation~\ref{#1}}

\def\plaineqref#1{\ref{#1}}

\def\1{\bm{1}}

\DeclareMathAlphabet{\mathsfit}{\encodingdefault}{\sfdefault}{m}{sl}
\SetMathAlphabet{\mathsfit}{bold}{\encodingdefault}{\sfdefault}{bx}{n}

\usepackage{amsmath}
\usepackage{booktabs}
\usepackage{graphicx}
\usepackage{placeins}
\usepackage{needspace}
\usepackage{etoolbox}
\usepackage{hyperref}
\usepackage{url}
\usepackage{xcolor}
\usepackage{pgfplots}
\usepackage{enumitem}
\usepackage{longtable}
\usepackage{booktabs}
\usepackage{multirow}
\usepackage{array}
\usepackage{caption}

\usepgfplotslibrary{groupplots}
\pgfplotsset{compat=1.16}

\definecolor{SrcCond}{RGB}{77,121,167}
\definecolor{SrcFuture}{RGB}{89,161,79}
\definecolor{SrcAction}{RGB}{225,86,89}
\definecolor{SrcText}{RGB}{242,142,42}
\definecolor{SrcImage}{RGB}{175,121,161}
\definecolor{MyGreen}{RGB}{27,120,55}
\definecolor{MyBlue}{RGB}{33,102,172}
\definecolor{MyOrange}{RGB}{217,95,14}
\definecolor{MyPurple}{RGB}{140,81,160}
\definecolor{EWAMColorE}{HTML}{2F6FA3}
\definecolor{EWAMColorW}{HTML}{4B8F8C}
\definecolor{EWAMColorA}{HTML}{C97A58}
\definecolor{EWAMColorM}{HTML}{80649A}

\title{\mbox{\includegraphics[height=\fontcharht\font`E]{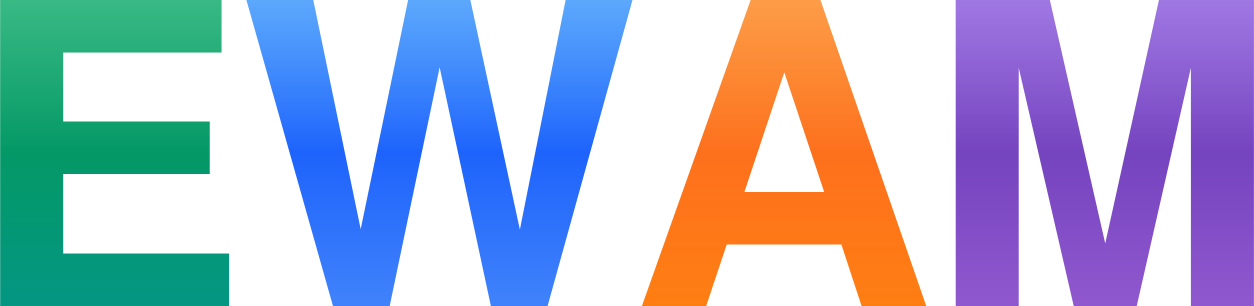}}:
Emergent Depth-Wise Specialization\\
in a Unified Embodied Model\\[0.15em]
{\fontsize{14}{17}\selectfont\mdseries From Semantic Understanding through Visual Foresight to Action}}

\author{Yinwang Intelligent Technology Co. Ltd.}

\begin{document}

\maketitle

\enlargethispage{0.85in}
\begin{center}
  \vspace{-0.4em}
  \includegraphics[width=\linewidth,trim={0.14in 0.16in 0.14in 0.02in},clip]{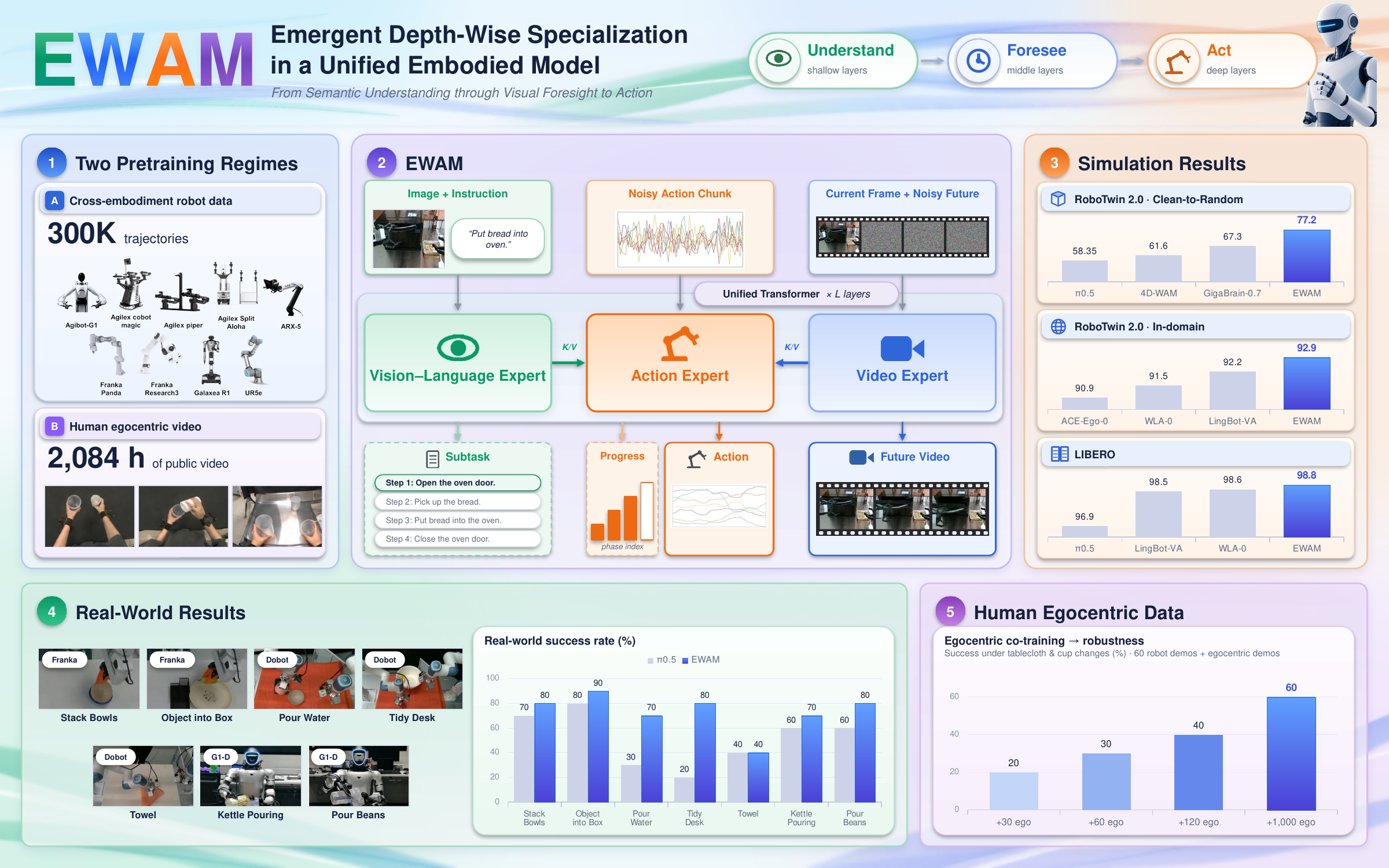}
\end{center}

\begin{abstract}
Vision-language-action (VLA) policies emphasize semantic understanding, whereas
world-action models (WAMs) learn predictive representations of environment
dynamics. Systems that expose a policy to both sources often still concentrate
action computation on a single expert. We present EWAM, an action-centric
unified embodied model whose asymmetric joint attention lets action tokens read
semantic, current-visual, predicted-future, and action information at every
layer while the perceptual experts retain their distinct roles. Without
layer-wise supervision, EWAM develops an \emph{emergent depth-wise
specialization}: action queries attend mainly to vision-language features in
shallow layers, to predicted future frames in intermediate layers, and to
action tokens themselves in deep layers. This handoff replicates across tasks
and is stable across denoising steps. Checkpoint tracking and causal interventions show that it is learned and that
action generation depends on it. EWAM is pretrained in two separate regimes, one on cross-embodiment
robot trajectories and one on human egocentric video. In simulation
and real-robot experiments, it surpasses existing
VLA, WAM, and hybrid baselines. Human egocentric data improve both
cross-embodiment transfer and real-robot robustness, and subtask-phase
supervision improves long-horizon completion. Together, these results suggest that unified embodied
learning can induce an ordered internal progression from semantic
understanding, through visual foresight, to action formation.

\vspace{0.4em}
\noindent{\sffamily\bfseries Project page:} \href{https://wanghao00pro.github.io/EWAM-project/}{\color{TRTeal}https://wanghao00pro.github.io/EWAM-project}
\end{abstract}

\section{Introduction}

Kenneth Craik's classic account of internal models proposed that an intelligent
organism carries a ``small-scale model'' of external reality, allowing it to
``react to future situations before they arise''~\citep{craik1943nature}.
This view casts purposeful action as an ordered computation: understanding the
present situation, anticipating how it may evolve, and then determining what
to do. Generalist robot policies face an analogous challenge. They must connect
at least three forms of computation: understanding what an instruction and the
current scene imply, anticipating how the physical world may evolve, and
producing precise actions.

Two dominant paradigms emphasize different parts of this pipeline. Vision-language-action
(VLA) models inherit rich semantic representations from vision-language models,
but their static pretraining provides limited explicit supervision for physical
dynamics~\citep{added_brohan2022rt1,added_driess2023palme,zitkovich2023rt2,
kim2024openvla,black2024pi0,black2025pi05}.
World-action models (WAMs), in contrast, learn from temporal visual data and can
acquire useful motion priors~\citep{ye2026dreamzero,li2026lingbot,
yuan2026fastwam}, yet predicting plausible futures does not by itself guarantee
semantic reasoning or fine-grained control~\citep{zhang2026wamrobustness}.

Prior attempts to obtain both capabilities generally follow one of two paths.
The first remains centered on a single paradigm, adding auxiliary outputs
while leaving action generation predominantly dependent on either
vision-language or video
features~\citep{wu2026lingbotvla2,zhang2025unijepa,sun2026vlajepa}. The second
attaches an independently operating world model or understanding module to a
policy~\citep{physicalintelligence2026pi07,long2026vista}. Such composition
adds capabilities, but does not necessarily induce
structured cooperation inside the action policy. Even a multi-expert
model~\citep{bi2026motus} can leave action computation concentrated on one
expert. This raises a more
fundamental question than how to add another module:
\emph{when semantic understanding, visual prediction, and action generation are
truly unified, do they continue to compete throughout the network, or does the
model develop an ordered progression reminiscent of purposeful human
action---first understanding the task, then imagining what should happen next,
and finally committing to an action?}

\begin{figure*}[t]
  \centering
  \includegraphics[width=\linewidth]{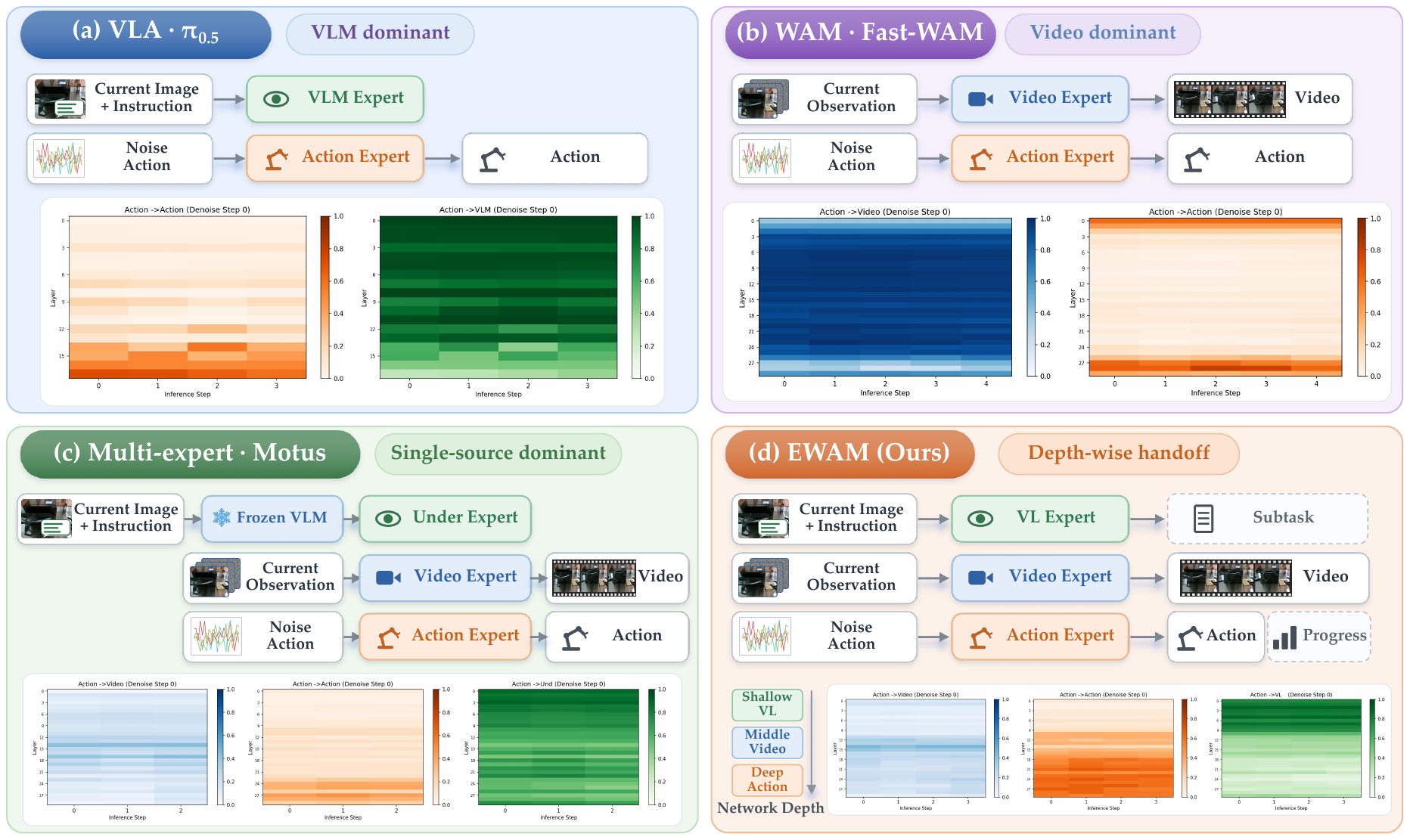}
  \caption{\textbf{Architectures and action-attention routing across
  embodied-model paradigms.} Each panel pairs a simplified architecture with
  action-query attention maps. Representative VLA, WAM,
  and multi-expert policies concentrate action attention on one primary
  source. In EWAM, the dominant source changes from vision-language features
  in shallow layers, through predicted-future features in intermediate
  layers, to action self-attention in deep layers. Colour encodes attention mass
  on an absolute zero-to-one scale, the horizontal axis is the policy inference
  step at denoising step~0, and the vertical axis is the layer index.
  Appendix~\ref{app:baseline_attention} gives further details.}
  \label{fig:motivation}
\end{figure*}

Figure~\ref{fig:motivation} makes this distinction concrete by visualizing
where action queries retrieve information across representative paradigms. In
$\pi_{0.5}$~\citep{black2025pi05}, action attention is consistently
concentrated on VLM tokens, reflecting a VLA-style computation in which
semantic representations are the primary source for control
(Figure~\ref{fig:motivation}(a)).
Fast-WAM~\citep{yuan2026fastwam} exhibits the complementary pattern: action
queries attend predominantly to video tokens, consistent with a WAM whose
control computation is centered on learned visual dynamics
(Figure~\ref{fig:motivation}(b)).
Motus~\citep{bi2026motus} integrates multiple expert modules through an
externally attached frozen VLM, yet its action attention remains dominated by
the understanding expert (Figure~\ref{fig:motivation}(c)). This observation
suggests that merely stacking
experts may still yield a single-expert bias resembling the VLA paradigm,
rather than inducing an internal division of labor.

To determine whether a unified policy can move beyond this single-expert bias,
we develop EWAM, a unified embodied model containing a vision-language (VL) expert,
a video expert, and an action expert. At every layer, action queries can attend
to semantic tokens, predicted visual tokens, and action tokens. An
asymmetric mask prevents video and vision-language queries from reading action
tokens or each other, making the action stream the integration site while
preserving the distinct predictive and semantic roles of the two perceptual
experts. Video and action outputs are trained jointly with flow-matching
objectives.

Despite receiving no explicit layer-wise supervision, EWAM develops a
consistent internal progression (Figure~\ref{fig:motivation}(d)). In shallow
layers, action tokens attend
primarily to vision-language tokens, including both language and current-image
content. In intermediate layers, attention to current-image tokens declines
while attention to predicted future-frame tokens rises. In deep layers, action
self-attention becomes dominant as the policy consolidates the information into
a control sequence. We refer to this depth-dependent handoff as
\emph{emergent depth-wise specialization}: an ordered progression from
semantic understanding, through visual foresight, to action formation.

We validate this claim along four dimensions: observational evidence,
developmental dynamics, causal interventions, and policy performance. Our
observational analysis confirms that the hierarchical handoff is stable across
tasks, while tracking training checkpoints shows that this internal
organization emerges during training. We then examine depth-specific
functional dependencies on vision-language, observed-frame, and
predicted-future features through layer-targeted training and fixed-checkpoint
interventions. Same-task counterfactual future-feature injection across five
tasks further tests whether actions depend on the specific content of
predicted futures. Finally, we evaluate the resulting policy in simulation and
real-world robot settings. We study two separate pretraining regimes, one on
cross-embodiment robot trajectories and the other on human egocentric video.
Initialized from cross-embodiment robot pretraining, EWAM achieves 72.1\% on
RoboTwin clean-to-random (C2R) evaluation and a 92.9\% average in the
in-domain setting. It also obtains 98.8\% on LIBERO and demonstrates
consistent performance across multiple real-world robot embodiments. Human
pretraining raises success on a held-out robot embodiment from 40.1\% to
66.9\%, and egocentric co-training raises real-robot success from 10\% to 60\%
under scene variations. Subtask-phase supervision during post-training further
improves long-horizon completion in both simulated and real-world tasks.

Our contributions are:
\begin{itemize}
  \item We develop EWAM, an action-centric unified embodied model that
  exposes semantic, predictive, and action representations to a common action
  stream through asymmetric joint attention while preserving the provenance
  of the perceptual experts.
  \item We identify and characterize emergent depth-wise
  specialization: a semantic-to-foresight-to-action handoff that replicates
  across diverse tasks and remains stable across denoising steps. Checkpoint
  tracking, layer-targeted interventions, and counterfactual future-feature
  injection show that it is learned and functionally consequential.
  \item We show that EWAM is competitive across complementary evaluation
  protocols, reaching 72.1\% on RoboTwin C2R, a 92.9\% in-domain average, and
  98.8\% on LIBERO, with consistent performance across three real-robot
  embodiments. Human egocentric data improve cross-embodiment transfer via
  pretraining and robustness under scene variations via co-training.
  Subtask-phase supervision further improves long-horizon completion.
\end{itemize}

\section{Related Work}

\paragraph{Vision-language-action models.}
VLA policies adapt vision-language representations to robot control, providing
a direct route for transferring semantic and web-scale knowledge into action
prediction. RT-1~\citep{added_brohan2022rt1} demonstrated large-scale tokenized
action prediction, while PaLM-E~\citep{added_driess2023palme} connected embodied
observations to a multimodal language model. RT-2~\citep{zitkovich2023rt2} casts
actions as language tokens and co-trains robot trajectories with vision-language
data. Octo~\citep{octo2024} learns a cross-embodiment generalist policy from
heterogeneous robot datasets, while OpenVLA~\citep{kim2024openvla} scales an open
VLA built from pretrained vision and language components. The $\pi$
family~\citep{black2024pi0} couples
a pretrained VLM to a flow-matching action expert, extending continuous
generative policy learning beyond the earlier Diffusion
Policy~\citep{added_chi2023diffusionpolicy}, and
$\pi_{0.5}$~\citep{black2025pi05} further combines multi-robot data, semantic
subtask prediction, and web data for open-world operation. These systems establish
the strength of semantic pretraining for instruction following and
generalization. However, when predictive visual dynamics are introduced
alongside VLM features, it remains unclear how the action stream coordinates
these complementary sources across network depth.

\paragraph{World-action models.}
WAMs instead build robot policies on predictive video representations and
jointly model future observations and actions. DreamZero~\citep{ye2026dreamzero} adapts a pretrained
video diffusion backbone into a large joint video-action policy, demonstrating
zero-shot transfer across embodiments. LingBot-VA~\citep{li2026lingbot}
learns causal frame prediction and policy execution in a shared latent space
with closed-loop rollout. Fast-WAM~\citep{yuan2026fastwam} separates video
co-training from test-time imagination, showing that predictive supervision can
benefit action learning even when future frames are not generated at
inference. A recent controlled robustness study~\citep{zhang2026wamrobustness}
reports complementary strengths between VLA and WAM policies, with their
relative performance shaped by architecture and training-data diversity. Our
work does not seek to determine which paradigm is universally superior;
instead, it examines how their semantic and predictive
representations are internally coordinated when exposed to one action stream.

\paragraph{Hybrid VLA--WAM systems.}
Recent systems begin to combine the semantic strengths of VLAs with the visual
foresight of world models, but most retain a modular pipeline. At inference,
$\pi_{0.7}$~\citep{physicalintelligence2026pi07} uses a high-level semantic policy to produce subtask instructions
and a separately trained world model based on BAGEL~\citep{added_deng2025bagel}
to generate subgoal images;
these external outputs then condition its VLA action policy. Cortex~2.0~\citep{aida2026cortex} similarly augments an existing
VLA with a distinct world-model planning module: the world model rolls out
candidate futures, a reward module selects one, and the selected trajectory is
passed to the downstream controller. Motus~\citep{bi2026motus} brings video
generation and action prediction into a Mixture-of-Transformers, but introduces
semantic understanding through an externally attached frozen VLM. These approaches establish the value of combining
semantics, foresight, and control, while preserving explicit module boundaries
or one-way intermediate interfaces. In contrast, our model couples semantic,
future-visual, and action representations within one layer-aligned architecture:
the action stream can interact with all three sources throughout depth, allowing
the understanding--foresight--action division of labor to emerge internally
rather than being prescribed by a serial pipeline.

\paragraph{Learning manipulation from human egocentric video.}
Human video provides manipulation supervision beyond robot demonstrations.
Corpora such as Ego4D~\citep{added_grauman2022ego4d},
EPIC-Kitchens~\citep{added_damen2018epickitchens}, and
Something-Something~v2~\citep{added_goyal2017something} support several learning routes.
R3M~\citep{added_nair2022r3m} and VC-1~\citep{added_majumdar2023vc1}
learn visual representations without action supervision, transferring knowledge
through the perceptual encoder. GR-1~\citep{added_wu2024gr1} and
LAPA~\citep{added_ye2025lapa} derive surrogate signals through future prediction
and latent action learning, respectively, without directly specifying metric
hand motion. Hand-centric approaches provide explicit motion targets:
VITRA~\citep{added_li2025vitra} constructs hand-centric vision-language-action episodes,
EgoDex~\citep{added_hoque2025egodex} pairs video with tracked hand motion, and
EgoMimic~\citep{added_kareer2025egomimic} co-trains human and robot
demonstrations captured through matched hardware. EWAM places three heterogeneous
human corpora and downstream robot data in a shared camera-relative end-effector
action space. Human pretraining uses the same joint video--action objective and
layer-aligned experts as robot pretraining, supplying shared perceptual and
predictive representations for downstream robot learning.

\section{Preliminaries and Problem Formulation}

Let $o_t$ denote the current image observation, $\ell$ a language instruction,
$v_{t+1:t+H}$ future visual observations, and
$a_{t:t+K-1}$ an action chunk. The policy models future-video and action
trajectories conditioned on $(o_t,\ell)$. Both output spaces are trained
with flow matching~\citep{added_lipman2023flowmatching}. For a clean target $x$ and Gaussian noise $\epsilon$, we
construct
\begin{equation}
  x_\tau = (1-\tau)x + \tau\epsilon,
  \qquad
  u^\star(x_\tau,\tau)=\epsilon-x.
\end{equation}
We regress a velocity field from the noisy sample to $u^\star$. Video and
action use independently sampled noise levels.

\section{A Unified Embodied Model}

\subsection{Architecture Overview}

\begin{figure}[t]
  \centering
  \includegraphics[width=\linewidth]{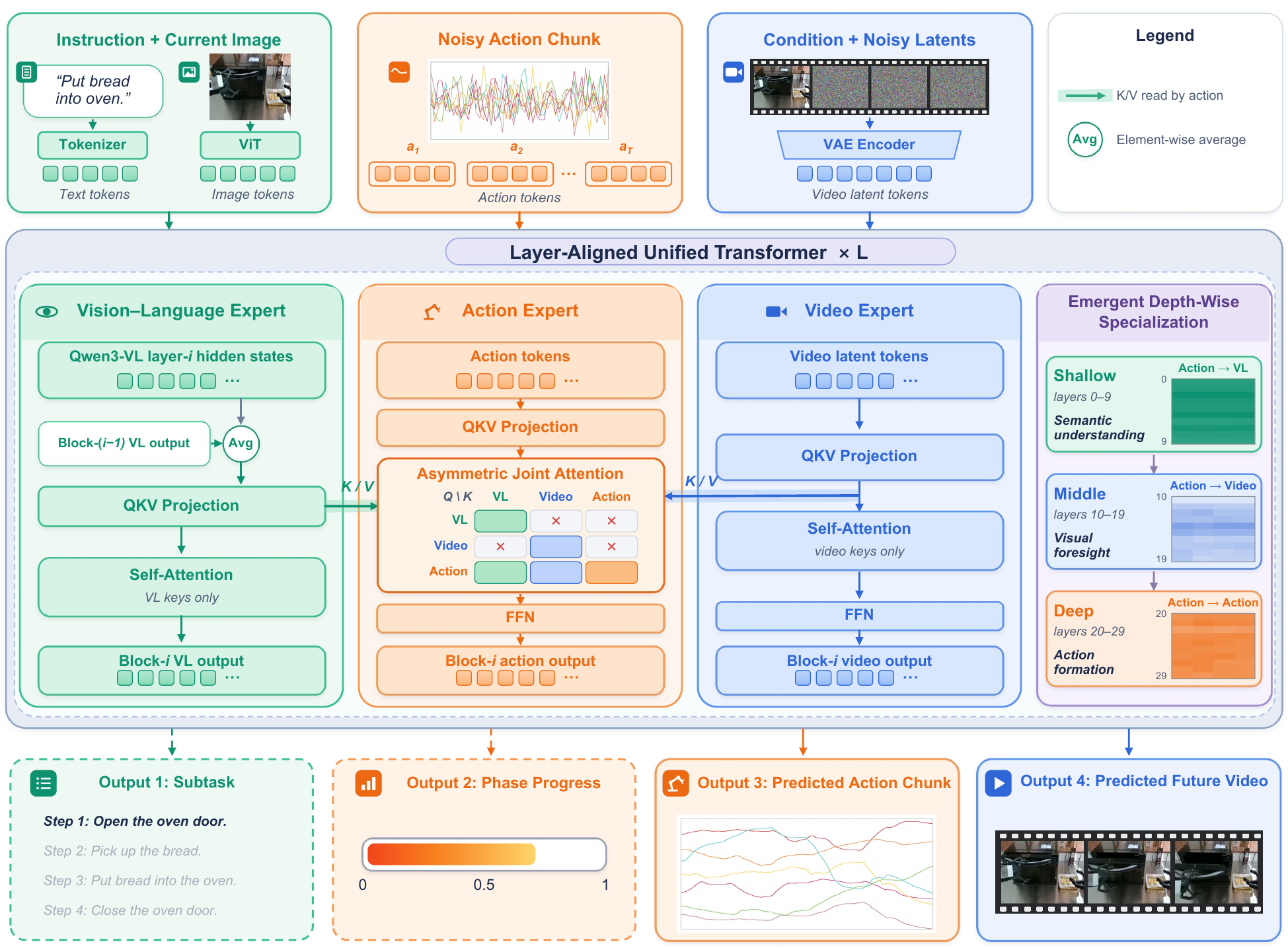}
  \caption{\textbf{Overview of our unified embodied model, EWAM.}
  Language and image observations, noisy action sequences, and noisy video
  latents are tokenized into layer-aligned VL, action, and video streams.
  Action queries attend to VL, video, and action keys, while VL and video
  queries remain within their own streams. The model predicts an action
  sequence and future video, and optionally predicts the active subtask
and its corresponding phase progress.}
  \label{fig:architecture}
\end{figure}

As illustrated in Figure~\ref{fig:architecture}, given an image observation
$o_t$ and a language instruction $\ell$, our model jointly
predicts a future visual trajectory
$v_{t+1:t+H}$ and an action chunk $a_{t:t+K-1}$. It contains three
layer-aligned experts. A vision-language expert encodes the instruction and
current image into semantic tokens $h^L$; a video expert processes the
condition-frame latent together with noisy future-video latents $h^V$; and an
action expert embeds the noisy action chunk into $h^A$. The video expert is initialized from
Wan2.2-TI2V-5B~\citep{wan2025}. The vision-language expert is initialized from
Qwen3-VL-2B~\citep{bai2025qwen3vl}. Unlike a pipeline in which a separately generated
future is passed to a downstream VLA, the three streams are executed in one
forward process and interact at every network depth through the action stream.

Training proceeds in two stages. During large-scale pretraining, we optimize
future prediction and action generation over cross-embodiment trajectories.
During task-specific post-training, we retain these objectives and introduce
subtask-phase supervision. Long-horizon demonstrations are partitioned into
semantic phases; the same partition supplies a textual label of the active
phase and a scalar phase index. The post-trained model therefore produces
\begin{equation}
  (\hat v_{t+1:t+H},\hat a_{t:t+K-1},\hat p_t,\hat y_t)
  = f_\theta(o_t,\ell),
\end{equation}
where $\hat y_t$ is the currently active subtask and $\hat p_t$ is the
normalized index of that phase in the subtask sequence.

\subsection{Action-Centric Asymmetric Attention}

Unrestricted trimodal attention can blur the pretrained roles of semantic and
predictive representations. We instead make action tokens the sole integration
interface. For video ($V$), action ($A$), and vision-language ($L$) tokens, the
attention mask permits
\begin{equation}
  A\leftarrow\{V,A,L\}, \qquad
  V\leftarrow V, \qquad
  L\leftarrow L.
  \label{eq:asymmetric_mask}
\end{equation}
Thus, action queries can retrieve semantic and predictive evidence, while the
video and vision-language streams retain modality-specific computation and cannot read one
another or the action stream. In implementation, the three query-side
computations in Figure~\ref{fig:architecture} are evaluated through a single
concatenated masked-attention operation, with each stream retaining its own
normalization and projection parameters.

The instruction and current image are first processed by
Qwen3-VL~\citep{bai2025qwen3vl}, yielding 28 hidden states. To expose these
semantic features across the 30 unified blocks, we align them to the unified
depth using the fixed mapping detailed in Appendix~\ref{app:model_details}. Let
$q_l^L\in\mathbb{R}^{N_L\times 2{,}048}$ denote the representation aligned to
block $l$, and let $z_{l-1}^L$ denote the VL state accumulated through the
preceding unified block. At unified block $l$, we form
\begin{equation}
  x_l^L =
  \begin{cases}
    q_l^L, & l=0,\\[2pt]
    \frac{1}{2}\left(q_l^L+z_{l-1}^L\right), & l>0.
  \end{cases}
  \label{eq:vlm_layer_fusion}
\end{equation}
Thus, $x_l^L$ combines the layer-matched Qwen3-VL feature with the VL
updates accumulated through unified depth; its VL-only attention output is
projected back to 2{,}048 dimensions and added residually to give $z_l^L$.
The video stream requires no such alignment, since the 30 unified blocks
follow the native depth of the Wan2.2 video expert. Both streams expose
their keys and values to action queries, making the action stream the sole
cross-modal reader; the action objective shapes semantic and predictive
representations only through these exposed keys and values. This separation
also makes action-query attention a well-defined measure of which source is
retrieved.

\subsection{Joint Video--Action Flow Matching}

Let $x_v$ and $x_a$ denote clean future-video and action targets. We independently
sample noise levels for the two modalities, construct their noisy interpolants,
and regress the corresponding flow velocities. The observed condition-frame
latent is kept fixed, while future-video and action latents are updated jointly.
The base objective is
\begin{equation}
  \mathcal{L}_{\mathrm{base}}
  = \lambda_v\lVert \hat{u}_v-u^\star_v\rVert_2^2
  + \lambda_a\lVert \hat{u}_a-u^\star_a\rVert_2^2,
\end{equation}
where $\hat{u}_v$ and $\hat{u}_a$ are the predicted video and action velocities,
$u^\star_v$ and $u^\star_a$ are the corresponding target flow velocities, and
$\lambda_v$ and $\lambda_a$ are loss weights. At inference, we jointly integrate the predicted video and action velocity
fields. The two trajectories are therefore denoised concurrently within the
same forward process, allowing evolving future-video features to participate
directly in action computation rather than serving as a separately generated
visual prompt.

\subsection{Subtask-Aware Post-Training}
\label{sec:progress_post_training}

\paragraph{Subtask-phase partition.}
Long-horizon demonstrations are divided into $N$ semantically meaningful
phases using annotated transition checkpoints. If the current observation lies
in phase $k_t\in\{0,\ldots,N-1\}$, we associate a short textual description
$y_t=(y_{t,1},\ldots,y_{t,M})$ of the active subtask and define a normalized
phase index
\begin{equation}
  p_t = \frac{k_t}{\max(N-1,1)}.
\end{equation}
The scalar $p_t$ is the position of the current phase in this partition, not
elapsed trajectory time. It is therefore derived from the same subtask
annotation rather than an independently defined progress target.

\paragraph{Subtask prediction.}
The vision-language stream receives only the high-level instruction, not the
subtask text. After the joint forward pass, a lightweight autoregressive
decoder predicts $y_t$ from a concatenated cross-attention memory, obtained by
linearly projecting the final action representation $h_t^A$ and the aligned
VL tokens $h_t^L$ with $W_A$ and $W_L$:
\begin{equation}
  \mathrm{Mem}_t
  =\bigl[W_A\,h_t^A;\; W_L\,h_t^L\bigr].
\end{equation}
The decoder reuses the Qwen3-VL token embedding and language-model output
head, and attends to $\mathrm{Mem}_t$ rather than to an independently executed
planner:
\begin{equation}
  p_\theta(y_t\mid \mathrm{Mem}_t)
  = \prod_{j=1}^{M}
    p_\theta(y_{t,j}\mid y_{t,<j},\,\mathrm{Mem}_t),
\end{equation}
where $p_\theta$ is the decoder distribution with parameters $\theta$ and
$y_{t,<j}$ denotes the preceding tokens, with token-level objective
\begin{equation}
  \mathcal{L}_{\mathrm{sub}}
  =-\sum_{j=1}^{M}m_{t,j}
    \log p_\theta(y_{t,j}\mid y_{t,<j},\,\mathrm{Mem}_t),
\end{equation}
where $m_{t,j}$ masks padding. We supervise the currently active subtask
throughout each annotated phase.

\paragraph{Phase-index head.}
We mean-pool the final action representation and regress $p_t$ as
\begin{equation}
  \bar h_t^A=\operatorname{MeanPool}(h_t^A), \qquad
  \hat p_t = W_2\,\operatorname{SiLU}(W_1\bar h_t^A),
\end{equation}
where $W_1$ and $W_2$ are the weights of a two-layer MLP head and
$\hat p_t$ is the predicted phase index, with
$\mathcal{L}_{\mathrm{prog}}=\lVert\hat p_t-p_t\rVert_2^2$.

\paragraph{Post-training objective.}
The full objective is
\begin{equation}
  \mathcal{L}_{\mathrm{post}}
  = \mathcal{L}_{\mathrm{base}}
  + \lambda_p\mathcal{L}_{\mathrm{prog}}
  + \lambda_s\mathcal{L}_{\mathrm{sub}}.
\end{equation}
Both terms come from the same phase partition and augment the base
video--action objective only during task-specific post-training; no external
high-level policy is required.

\section{Emergent Depth-Wise Specialization}
\label{sec:specialization}

\subsection{Measuring Modality Allocation}

We analyze how the information retrieved by action queries changes with model
depth. Let $\mathcal{Q}_A$ denote the set of action queries, $\mathcal{K}$ the
set of all keys visible to them, and $\mathcal{K}_m\subseteq\mathcal{K}$ the
keys from source $m$. With $\alpha_{qk}^{(l)}$ the post-softmax attention
weight from query $q$ to key $k$ in block $l$, the allocation to source $m$ is
\begin{equation}
  R_l^m =
  \frac{\sum_{q\in\mathcal{Q}_A}\sum_{k\in\mathcal{K}_m}
  \alpha_{qk}^{(l)}}
  {\sum_{q\in\mathcal{Q}_A}\sum_{k\in\mathcal{K}}\alpha_{qk}^{(l)}}.
  \label{eq:attention_allocation}
\end{equation}
We distinguish five sources: VL-text, VL-image, condition-frame,
future-frame, and action tokens. We aggregate over attention heads and action
queries, normalize within each layer, and average over samples and denoising
steps, using the robot-pretrained in-domain checkpoint of
Table~\ref{tab:robotwin_indomain}. This statistic shows where action tokens
retrieve information; Section~\ref{sec:causal_validation} tests functional
necessity.

\subsection{Emergent Specialization Across Depth}

Action queries exhibit three overlapping computational regimes rather than a
hard-routed pipeline. Unless stated otherwise, every value reported in this
section is a macro average over one representative sequence from each of the 50
RoboTwin~2.0 tasks under the randomized protocol, obtained by averaging the
five measured denoising steps
within a task before averaging across tasks with equal weight, using the
robot-pretrained in-domain checkpoint.
Appendix~\ref{app:cross_task_attention} reports the same quantities with 95\%
task-bootstrap confidence intervals, and
Appendix~\ref{app:token_normalization} reports their token-count-normalized
counterparts.

\paragraph{Shallow understanding.} In layers 0--9,
VL tokens receive 75.4\% of action attention, compared with 14.1\% for video
and 10.6\% for action tokens. Action computation therefore begins by retrieving
the instruction and current-scene information needed to identify and ground the
task.

\begin{figure*}[t]
  \centering
  \includegraphics[width=\linewidth]{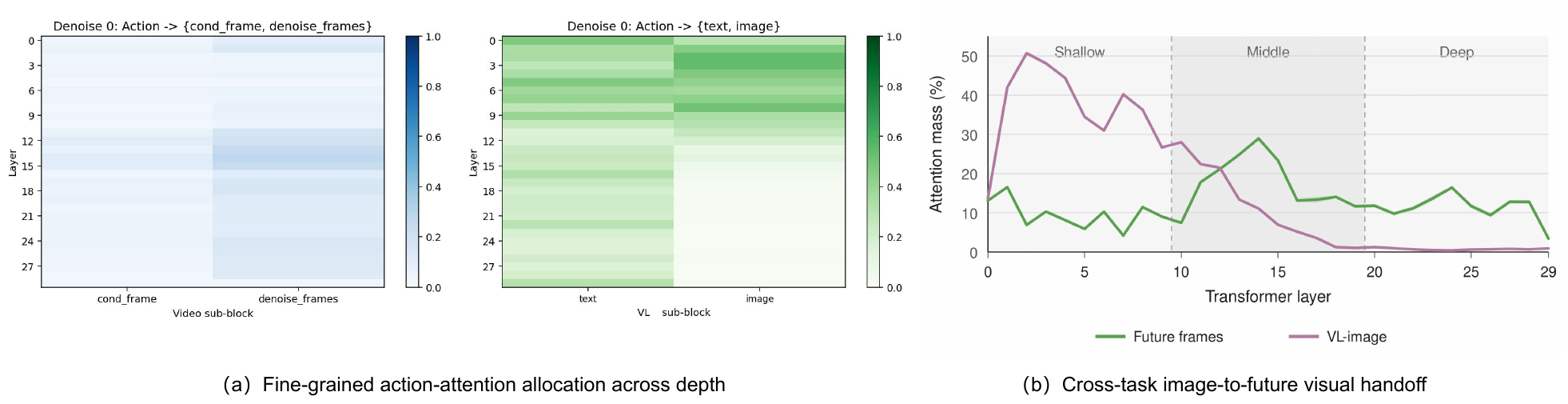}
  \caption{\textbf{Fine-grained visual handoff across depth.} (a) Action attention to
  observed and predicted video features and to VL text and image features on the
  \textsc{Adjust Bottle} task. (b) Cross-task VL-image and future-frame attention,
  macro-averaged over 50 tasks and five denoising steps.}
  \label{fig:depth_submodalities}
\end{figure*}

\paragraph{Intermediate visual foresight.}
Across the intermediate stage (layers 10--19), attention becomes more
distributed across VL, video, and action tokens, with allocations of 36.6\%,
24.3\%, and 39.1\%, respectively. The visual-source handoff is concentrated in
layers 11--15, with a cross-task median raw crossover at layer 13 (IQR 12--13).
Figure~\ref{fig:depth_submodalities} reveals the structure hidden
by the stage-level aggregates: attention to VL-image tokens drops sharply,
while attention to predicted future-frame tokens increases (see
Appendix~\ref{app:token_normalization} for token-count-normalized results);
attention to the observed condition frame remains comparatively weak. Thus,
the model retains linguistic guidance from the VL expert while shifting its
visual dependence from the current observation toward predictive
representations of future scene evolution.

\paragraph{Deep action formation.}
In layers 20--29, action self-attention rises to 59.5\%, while VL and video
attention fall to 25.2\% and 15.3\%, respectively. What the deep stage retains
from the two perceptual sources is selective rather than uniform: averaged over
layers 16--29, VL-image tokens hold only 1.3\% of action attention against
11.8\% for predicted future frames
(Table~\ref{tab:token_normalized_visual_handoff}). Deep layers therefore
primarily refine motor trajectories while preserving concise semantic and
predictive context. Together, these regimes constitute a gradual
understanding--foresight--action handoff rather than three isolated modules.

\paragraph{Cross-task replication and stability.}
We repeat the analysis on one representative sequence from each of the 50
RoboTwin~2.0 tasks, first averaging the five denoising steps within each task
and then macro-averaging across tasks.
Figure~\ref{fig:depth_submodalities}(b) shows that the image-to-future visual
handoff is not specific to \textsc{Adjust Bottle}: all 50 tasks exhibit the raw
handoff, with a stable future--image crossover at median layer 13 (IQR 12--13).
Across the complete five-source routing profile, VL attention is largest in
the shallow stage for all tasks and action attention is largest in the deep
stage for all tasks. After task-specific token-count normalization, the visual
handoff remains present in 49 of 50 tasks and its median crossover shifts to
layer 14. Attention profiles are also highly stable across denoising steps
(mean pairwise correlation $0.998$); complete raw curves, normalized curves,
and full statistics are reported in Appendix~\ref{app:cross_task_attention}.

Figure~\ref{fig:motivation} further shows that $\pi_{0.5}$, Fast-WAM, and Motus
concentrate action attention on one dominant source, whereas our model changes
its primary source with depth.

\subsection{Learning the Visual Handoff during Training}

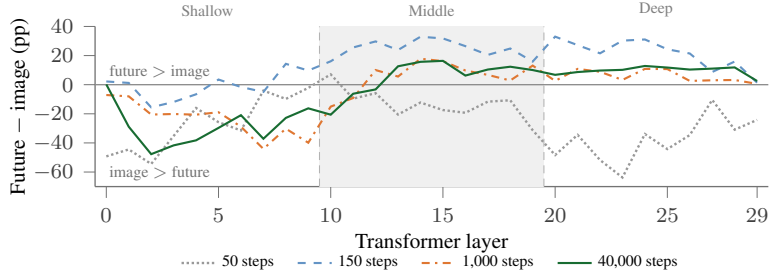
\begin{figure}[t]
  \centering
  \begin{minipage}{0.62\linewidth}
    \begin{tikzpicture}
\begin{axis}[
  width=\linewidth, height=3.7cm,
  scale only axis=false,
  xlabel={Transformer layer}, xmin=-0.5, xmax=29.5,
  xtick={0,5,10,15,20,25,29},
  ymin=-70, ymax=40, ytick={-60,-40,-20,0,20,40},
  ylabel={Future $-$ image (pp)},
  tick label style={font=\scriptsize, text=black!75},
  label style={font=\scriptsize},
  xlabel style={yshift=3pt}, ylabel style={yshift=-6pt},
  axis lines*=left, axis line style={black!55, line width=0.5pt},
  tick style={black!55}, tick align=outside,
  every axis plot/.append style={line width=0.9pt},
  legend style={font=\tiny, draw=none, fill=none,
                at={(0.5,-0.36)}, anchor=north, legend columns=4,
                /tikz/every even column/.append style={column sep=0.6em}},
  legend image code/.code={\draw[#1] (0cm,0cm) -- (0.45cm,0cm);},
  legend cell align=left,
  clip=false,
]
  \fill[black!6] (axis cs:9.5,-70) rectangle (axis cs:19.5,40);
  \draw[black!30, densely dashed, line width=0.4pt]
    (axis cs:9.5,-70) -- (axis cs:9.5,40)
    (axis cs:19.5,-70) -- (axis cs:19.5,40);
  \node[font=\tiny, text=black!50, anchor=south] at (axis cs:4.5,40) {Shallow};
  \node[font=\tiny, text=black!50, anchor=south] at (axis cs:14.5,40) {Middle};
  \node[font=\tiny, text=black!50, anchor=south] at (axis cs:24.5,40) {Deep};
  \addplot[black!45, line width=0.5pt, forget plot]
    coordinates {(-0.5,0) (29.5,0)};
  \node[font=\tiny, text=black!55, anchor=south west, inner sep=1pt]
    at (axis cs:0,1.5) {future $>$ image};
  \node[font=\tiny, text=black!55, anchor=south west, inner sep=1pt]
    at (axis cs:0,-68) {image $>$ future};
  \addplot[black!40, densely dotted, line width=0.9pt] coordinates {(0,-49.25) (1,-44.49) (2,-54.44) (3,-35.24) (4,-16.01) (5,-25.75) (6,-31.46) (7,-4.08) (8,-9.78) (9,-2.21) (10,7.08) (11,-9.57) (12,-5.84) (13,-20.49) (14,-12.26) (15,-17.43) (16,-19.29) (17,-11.65) (18,-10.71) (19,-31.30) (20,-48.34) (21,-34.46) (22,-51.74) (23,-63.74) (24,-33.89) (25,-44.22) (26,-34.61) (27,-10.46) (28,-30.90) (29,-24.20)};
  \addlegendentry{50 steps}
  \addplot[MyBlue!70, dashed, line width=0.8pt] coordinates {(0,2.21) (1,1.21) (2,-15.57) (3,-11.87) (4,-6.60) (5,3.53) (6,-1.57) (7,-5.00) (8,14.32) (9,9.73) (10,16.08) (11,25.57) (12,29.73) (13,23.83) (14,32.95) (15,31.77) (16,26.48) (17,20.47) (18,24.84) (19,15.87) (20,33.01) (21,27.17) (22,21.55) (23,30.26) (24,31.06) (25,24.31) (26,21.43) (27,8.73) (28,15.89) (29,1.41)};
  \addlegendentry{150 steps}
  \addplot[MyOrange!85, dash dot, line width=0.8pt] coordinates {(0,-7.10) (1,-8.01) (2,-20.51) (3,-20.14) (4,-20.59) (5,-18.93) (6,-28.84) (7,-44.11) (8,-30.20) (9,-39.87) (10,-15.06) (11,-9.16) (12,10.05) (13,5.58) (14,17.82) (15,16.18) (16,9.75) (17,6.73) (18,2.89) (19,13.12) (20,2.58) (21,10.85) (22,8.86) (23,3.28) (24,10.85) (25,10.74) (26,2.58) (27,3.03) (28,3.23) (29,0.54)};
  \addlegendentry{1{,}000 steps}
  \addplot[MyGreen!90!black, line width=0.8pt] coordinates {(0,0.18) (1,-28.88) (2,-47.72) (3,-41.69) (4,-38.23) (5,-29.68) (6,-20.95) (7,-37.13) (8,-22.82) (9,-16.25) (10,-20.60) (11,-6.22) (12,-3.25) (13,12.64) (14,15.70) (15,16.37) (16,6.26) (17,10.41) (18,12.34) (19,10.05) (20,6.77) (21,8.68) (22,9.72) (23,10.19) (24,12.86) (25,11.77) (26,10.39) (27,11.04) (28,11.86) (29,2.52)};
  \addlegendentry{40{,}000 steps}
\end{axis}
\end{tikzpicture}
  \end{minipage}
  \caption{\textbf{Emergence of the visual handoff.}
  Raw attention-mass difference between future-frame and VL-image tokens
  (percentage points) across depth at four checkpoints of the training run of
  Table~\ref{tab:checkpoint_evolution}. Positive values indicate that action
  queries place more attention mass on predicted future frames than on
  VL-image tokens. At 150 steps, raw future-frame mass first exceeds VL-image mass at
  layer 8; at 40{,}000 steps, it exceeds VL-image mass at every layer from 13
  onward.}
  \label{fig:checkpoint_handoff}
\end{figure}

We trace action-attention routing across twelve checkpoints spanning 50 to
40{,}000 optimizer steps of one training run without robot pretraining, to
determine whether the depth-wise division of labor is present from the start or
emerges through training. All checkpoints belong to the same thirty-two-GPU
training run and share the asymmetric mask and initialization regime. At 50
steps, action queries attend to VL-image tokens at every depth, and VL-text
receives only 2.9\% of shallow attention; no layer-wise division is yet
visible. A coarse organization appears within the first few hundred steps. By
150 steps, shallow text attention has risen to 22.6\%, image attention
concentrates in shallow layers, and raw future-frame attention mass first
exceeds VL-image mass at layer 8. Continued training refines this
organization until, at 40{,}000 steps, the three regimes of
Section~\ref{sec:specialization} are established: shallow layers read the
instruction and current image, and raw future-frame attention mass exceeds
VL-image mass at every layer from 13 onward
(Figure~\ref{fig:checkpoint_handoff}). This refinement is not monotonic: under
token-count normalization, shallow image enrichment rebounds to $1.79$ at
5{,}000 steps before deep image enrichment falls to $0.04$ at convergence.

The mask determines which sources action queries can access, but not the depth
at which each source is used. The layer-specific routing therefore emerges
during the training of our model. The trajectory is measured on one
representative sequence from one run; Appendix~\ref{app:checkpoint_evolution}
and Table~\ref{tab:checkpoint_evolution} give full values, including
token-normalized measurements.

\subsection{Causal Validation through Layer-Targeted Interventions}
\label{sec:causal_validation}

The attention profiles above are observational: they show where information is
routed, but not whether the routed sources affect behavior. We therefore use
two complementary tests in Table~\ref{tab:layer_intervention}. The left panel
reports structural ablations trained independently without pretraining; these
test whether the proposed layer-wise access pattern is beneficial when a model
can adapt during training. The right panel masks selected keys only at
inference time in one fixed cross-embodiment robot-pretrained checkpoint, thereby measuring its
learned dependence without allowing compensation through retraining. We
evaluate the inference interventions on the full RoboTwin Randomized protocol
of 50 tasks and 100 episodes per task, with matched episode seeds.

\begin{table*}[t]
  \caption{\textbf{Layer-targeted training and fixed-checkpoint interventions on RoboTwin Randomized.}
  The left panel reports independently trained models without pretraining.
  The right panel applies action-attention masks only at inference time to one
  fixed cross-embodiment robot-pretrained checkpoint, without retraining. Scores and deltas are
  therefore relative to the separate full-model baseline in each panel and
  should not be compared directly across panels. ``Action only'' removes both
  VL and video keys from action queries in the specified layers.}
  \label{tab:layer_intervention}
  \centering
  \small
  \begin{tabular}{@{}lcc@{\qquad}lcc@{}}
    \toprule
    \multicolumn{3}{c}{\textit{Training results (without pretraining)}} &
    \multicolumn{3}{c}{\textit{Inference results (robot-pretrained model)}} \\
    \cmidrule(r){1-3}\cmidrule(l){4-6}
    Configuration & Score & $\Delta$ &
    Configuration & Score & $\Delta$ \\
    \midrule
    Full unified model & 83 & -- &
    Full unified model & 92.80 & -- \\
    Symmetric full attention & 81 & $-2$ &
    Mask all video, layers 0--9 & 91.98 & $-0.82$ \\
    No VL access, layers 0--9 & 78 & $-5$ &
    Mask VL, layers 0--9 & 76.53 & $-16.27$ \\
    No video access, layers 10--19 & 78 & $-5$ &
    Mask condition, layers 10--19 & 81.40 & $-11.40$ \\
    Action only, layers 24--29 & 83 & $0$ &
    Mask future, layers 10--19 & 3.55 & $-89.25$ \\
    & & &
    Mask all video, layers 10--19 & 0.25 & $-92.55$ \\
    & & &
    Mask all video, layers 20--29 & 65.82 & $-26.98$ \\
    \bottomrule
  \end{tabular}
\end{table*}

\paragraph{Training-time structural ablations.}
Without pretraining, the asymmetric model reaches 83\%, compared with 81\% for
symmetric full attention. Removing VL access in layers 0--9 or video access in
layers 10--19 reduces success to 78\%, whereas restricting layers 24--29 to
action tokens preserves the 83\% score. These results are consistent with
early semantic grounding, intermediate visual integration, and late action
refinement. Because each variant is trained separately and may adapt to its
mask, however, they provide architectural support rather than direct evidence
that the pretrained policy uses each source at the corresponding depth.

\paragraph{Fixed-checkpoint depth interventions.}
Inference-time masking reveals a sharper stage-specific dependence. Relative
to the 92.80\% pretrained baseline, removing all video keys in shallow layers
0--9 changes success by only $-0.82$ points (91.98\%), while shallow VL masking
costs 16.27 points (76.53\%). In contrast, the
matched all-video intervention across depth yields 91.98\% in layers 0--9,
0.25\% in layers 10--19, and 65.82\% in layers 20--29. Thus, the policy has
little shallow dependence on video, maximal dependence in the intermediate
stage, and a smaller but still substantial dependence in the deep stage.

Decomposing the intermediate video pathway identifies which content drives
this effect. Masking condition-frame keys alone retains 81.40\% success, but
masking intermediate future features alone nearly collapses performance,
reducing success from 92.80\% to 3.55\%; masking both sources reduces it
further to 0.25\%. The dominant intermediate dependence is therefore on
predicted-future representations, not on visual tokens indiscriminately.
Together with the shallow VL intervention, this depth- and source-specific
pattern functionally corroborates the observational handoff.

\begin{table}[!t]
  \caption{\textbf{Counterfactual future-feature injection across five tasks.}}
  \label{tab:counterfactual_future_injection}
  \centering
  \small
  \setlength{\tabcolsep}{3pt}
  \begin{tabular}{@{}p{0.17\linewidth}cp{0.19\linewidth}p{0.34\linewidth}c@{}}
    \toprule
    Task & No injection & Injection setting & Observed behavior or outcome &
    Success \\
    \midrule
    \multirow{2}{*}{\raisebox{-1.2ex}{\scriptsize\textsc{Adjust Bottle}}} &
    \multirow{2}{*}{\raisebox{-1.2ex}{5/5}} & Left-facing (43, 46) &
    Donor-directed first grasp misses; both recover after cutoff & 2/2 \\
    & & Right-facing (44, 45, 47) &
    Donor-consistent right arm instead of required left arm; no recovery &
    0/3 \\
    \midrule
    {\scriptsize\textsc{Place Empty Cup}} & 5/5 & All targets (43--47) &
    All injected rollouts fail & 0/5 \\
    {\scriptsize\textsc{Stack Blocks Two}} & 5/5 & All targets (43--47) &
    All injected rollouts fail & 0/5 \\
    {\scriptsize\textsc{Click Bell}} & 5/5 & All targets (43--47) &
    Only seed 46 remains successful & 1/5 \\
    {\scriptsize\textsc{Grab Roller}} & 5/5 & All targets (43--47) &
    Only seed 45 remains successful & 1/5 \\
    \midrule
    \textbf{Overall} & \textbf{25/25} & All injected targets &
    Success decreases from 100\% to 16\% & \textbf{4/25} \\
    \bottomrule
  \end{tabular}
\end{table}

\paragraph{Counterfactual future-feature injection.}
Masking establishes necessity, but a collapse could arise merely because the
network expects the future-token pathway to be present. To test whether actions
depend on its content, we conduct same-task counterfactual interventions across
five tasks. For each task, we cache the future-frame keys and values from a
successful donor rollout (seed 100) at layers 10--19 during the first three
policy calls. We inject them during the corresponding calls of five target
rollouts (seeds 43--47), while preserving each target's condition-frame, VL,
state, and action representations. All 25 target rollouts succeed without
injection.

As summarized in Table~\ref{tab:counterfactual_future_injection}, injection
reduces aggregate success from 25/25 to 4/25. All injected rollouts fail on
\textsc{Place Empty Cup} and \textsc{Stack Blocks Two}. Only seed 46 succeeds
on \textsc{Click Bell}, and only seed 45 succeeds on \textsc{Grab Roller}.
On \textsc{Adjust Bottle}, success decreases from 5/5 to 2/5. These results
show that sensitivity to counterfactual future representations extends across
tasks rather than being specific to bottle adjustment.

\textsc{Adjust Bottle} additionally provides an interpretable view of how the
injected representations alter behavior. A bottle whose opening faces left
should be adjusted with the right arm, whereas a right-facing opening requires
the left arm. Injection causes all five targets to initially follow the donor's
right-arm behavior despite changes in bottle identity, orientation, and
position. For the two left-facing targets (seeds 43 and 46), the donor-directed
first grasp misses the target position; once injection stops, both policies
re-localize the bottle and succeed on a second grasp. For the three right-facing
targets (seeds 44, 45, and 47), the transferred right-arm choice conflicts with
the required left-arm strategy, and all three episodes fail.

\noindent
The consistent degradation across five tasks provides cross-task,
content-sensitive causal evidence that action generation depends on
intermediate future representations. The \textsc{Adjust Bottle} trajectories
further show that these representations can steer spatial targeting and arm
selection, while target-conditioned control can re-emerge after injection ends.

\FloatBarrier
\section{Human Egocentric Learning: Representation and Data Construction}
\label{sec:human_learning}

To extend the unified model to human demonstrations, we use two complementary
data interfaces, evaluated in Section~\ref{sec:human_data_study}.
Large-scale human pretraining uses
camera-relative wrist-motion targets from existing corpora, whereas G1-D
co-training uses task-specific recordings retargeted to the robot interface.
The final subsection assesses the retargeting quality of that offline
procedure and illustrates robot replay, separately from learned-policy
evaluation.

\subsection{Human Egocentric Pretraining}
\label{sec:human_pretraining}

The human-pretraining experiments in Section~\ref{sec:human_data_study}
use the same architecture and joint video--action objective as robot
pretraining. To provide consistent motion targets across the three egocentric
corpora, we express wrist displacement in the anchor camera frame and mask
unavailable action dimensions.

\paragraph{A camera-relative end-effector action space.}
Our starting point is that the egocentric camera is the one reference frame
that a human recording and a robot platform both provide. We therefore express
manipulation as wrist motion relative to that camera. Let $(R_{we},t_{we})$
denote the wrist pose in world coordinates at an anchor frame,
$(R_{we^\star},t_{we^\star})$ its pose at a later frame, and $R_{wc}$ the
orientation of the egocentric camera at the anchor. We first form the motion in
the wrist's own frame,
\begin{equation}
  R_{ee^\star}=R_{we}^{\top}R_{we^\star},
  \qquad
  t_{ee^\star}=R_{we}^{\top}\bigl(t_{we^\star}-t_{we}\bigr),
\end{equation}
where $R_{ee^\star}$ and $t_{ee^\star}$ are the rotation and translation of the
wrist from the anchor to the later frame, expressed in the anchor wrist frame;
we then map it into the camera frame through $R_{ce}=R_{wc}^{\top}R_{we}$:
\begin{equation}
  \Delta R_c=R_{ce}R_{ee^\star}R_{ce}^{\top},
  \qquad
  \Delta t_c=R_{ce}\,t_{ee^\star},
  \qquad
  \Delta\omega_c=\log_{SO(3)}\bigl(\Delta R_c\bigr),
  \label{eq:camera_delta}
\end{equation}
where $\Delta R_c$ and $\Delta t_c$ are the same wrist motion expressed in the
anchor camera frame and $\log_{SO(3)}$ maps $\Delta R_c$ to its axis--angle
vector $\Delta\omega_c$. Each hand contributes a translation and a rotation vector together with two
reserved slots, giving a sixteen-dimensional action
\begin{equation}
  a=\bigl[\,\Delta t_c^{L},\,\Delta\omega_c^{L},\,0,\,0,\;
            \Delta t_c^{R},\,\Delta\omega_c^{R},\,0,\,0\,\bigr],
\end{equation}
where superscripts $L$ and $R$ denote the left and right hands, and the two
reserved dimensions for each hand are excluded from the loss
using the same validity mask as for cross-embodiment padding. All actions in a chunk are referenced to
the same anchor pose and the same anchor camera, so the $i$-th action is the
displacement from the anchor to frame $i{+}1$ rather than a step-to-step
increment; the state is the current wrist pose expressed in that camera frame.
This construction removes dependence on the world-frame origin and axis
convention while preserving metric translation magnitudes; it does not
normalize embodiment scale or remove kinematic differences. It gives human
wrists and robot end-effectors a common task-space layout. On the robot side,
the reserved slots carry gripper commands.

\paragraph{Reducing heterogeneous sources to a common frame.}
The three corpora we pretrain on expose camera geometry in different forms, and
each requires its own route to the anchor camera pose. VITRA-1M~\citep{added_li2025vitra} publishes
world-to-camera extrinsics, which we invert; EgoDex~\citep{added_hoque2025egodex} stores a camera-to-world
transform directly; and Xperience~\citep{added_ropedia2026xperience} provides a SLAM body pose $T_{wb}$ together
with a stereo-rectified calibration $T_{cb}$, from which the camera pose
follows as $T_{wc}=T_{wb}T_{cb}^{-1}$. Once a source is reduced to world-frame
wrist poses and an anchor camera pose, all three enter the identical
computation in Eq.~\plaineqref{eq:camera_delta}, so a given dimension
carries the same physical meaning across the mixture. Per-hand tracking
validity, derived from each source's own confidence or visibility signal, masks
that hand's six dimensions whenever the estimate is unreliable, which keeps
invalid reconstructions from contributing gradients. No action normalization
is applied at this stage. The coordinate conversion preserves each source's
metric scale, so consistent units and calibration remain necessary.

\subsection{Robot-Compatible Demonstrations from Egocentric Capture}
\label{sec:ego_collection}

Wearable systems such as BifrostUMI~\citep{added_wang2026bifrostumi} combine
human motion sensing with instrumented manipulation interfaces to collect
demonstrations without the target robot. Video-based pipelines reconstruct hand
motion and camera trajectories to obtain metric action
supervision~\citep{added_macrodata2026pipeline}. These motions cannot be used
directly as robot actions: head motion changes the observed hand trajectory,
and human wrists and robot tools have different interaction frames.
We construct robot-compatible demonstrations by separating
task-space motion and grasp timing from the joint configurations that realize
them. Metric wrist tracking and visual hand geometry define the human-side
reference; interaction-aware tool alignment and sequence optimization convert
it into bimanual robot trajectories. Paired with egocentric images, these
references support task-specific human--robot co-training on G1-D, complementing
the wrist-motion
pretraining in Section~\ref{sec:human_pretraining}.

\begin{figure}[!t]
  \centering
  \includegraphics[width=0.90\linewidth]{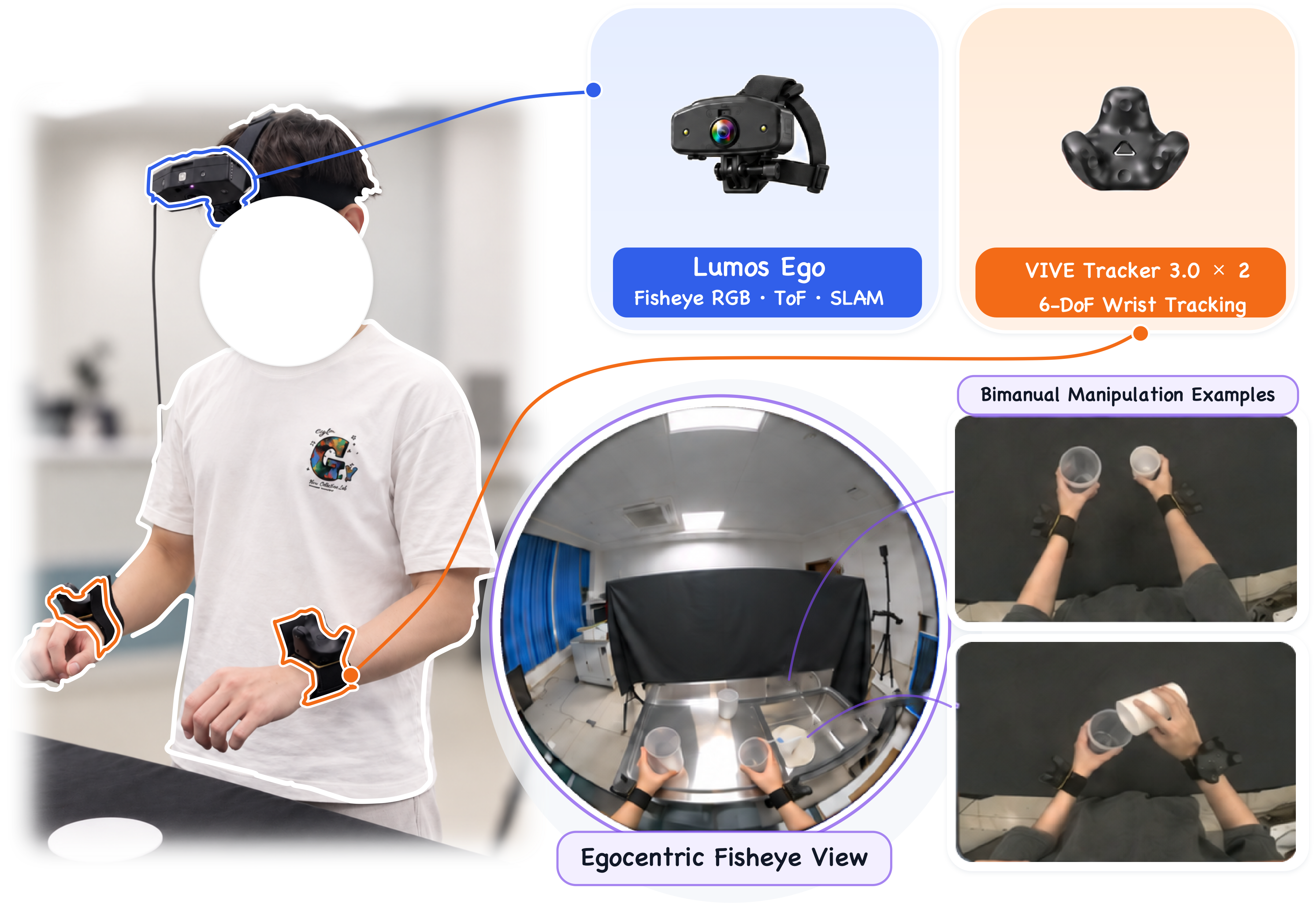}
  \caption{\textbf{Egocentric demonstration collection system.}
  A Lumos Ego headset records fisheye RGB, time-of-flight depth, and camera
  motion, while two wrist-mounted Vive Tracker~3.0 devices measure metric
  six-degree-of-freedom wrist motion.}
  \label{fig:ego_collection_system}
\end{figure}

\paragraph{Metric hand-motion reconstruction.}
Recovering manipulation motion requires separating wrist movement from camera
movement while retaining finger articulation. We combine a Lumos Ego headset,
which records fisheye RGB, time-of-flight depth, and camera poses from
simultaneous localization and mapping (SLAM), with two wrist-mounted Vive
Tracker~3.0 devices that measure metric six-degree-of-freedom motion
(Figure~\ref{fig:ego_collection_system}).

HaWoR~\citep{added_zhang2025hawor} estimates wrist poses and hand parameters
from the egocentric video; MANO~\citep{added_romero2017mano} decodes the hand
parameters into articulated geometry.
Camera poses map the visual estimates into the SLAM world frame, where the
registered Tracker measurements provide metric motion constraints. This
combines measured wrist motion with the finger geometry needed for tool
alignment and grasp inference. Appendix~\ref{app:ego_geometry} gives the
visual tracking and temporal processing details.

\paragraph{Interaction-aware action alignment.}
Robot action targets must specify tool motion and when to grasp or release.
We infer grasp timing from the relation between the hand and the
manipulated object. SAM~3 masks~\citep{added_carion2025sam3}, propagated by
Lucas--Kanade optical flow~\citep{added_lucas1981registration}, provide
image-space object geometry. Projected finger-to-mask distances, object
occupancy within the pinch region, and hand-aperture changes supply grasp
evidence. Distinct temporal conditions for grasp and release convert this
evidence into stable binary gripper commands.

EgoInfinity~\citep{added_wang2026egoinfinity} and
Do as I Do~\citep{added_paliwal2026doasido} recover hand--object interactions
and map reconstructed motion to robot embodiments. In our pipeline,
human-to-gripper alignment~\citep{added_qwen2026robotmanip} must also account
for the difference between the anatomical wrist and the tool center point (TCP).
We estimate a wrist-local interaction point from the
median thumb--index midpoint in valid closed-hand frames. Fixing this offset
separates the reference motion from instantaneous finger articulation: the
point follows wrist translation and rotation while the fingers open or close.

\begin{figure}[!htbp]
  \centering
  \includegraphics[width=\linewidth]{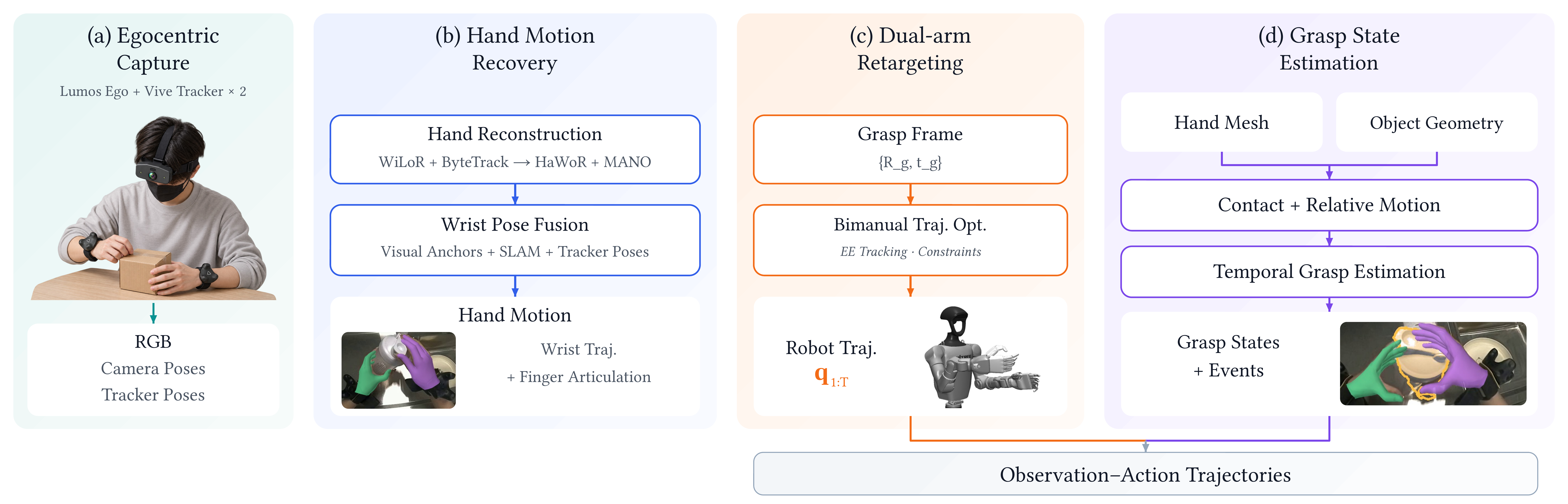}
  \caption{\textbf{Egocentric capture and observation--action trajectory construction.}
  (a) Capture setup.
  (b) Hand-motion reconstruction from RGB, SLAM, and Tracker observations.
  (c) Grasp-frame construction and bimanual trajectory optimization.
  (d) Mask-based grasp-state estimation.}
  \label{fig:ego_pipeline}
\end{figure}

A shared rigid transform places both hands in the robot workspace using a
real-robot reference trajectory, retaining metric scale and the relative
geometry of the two reference points. The median thumb--index direction and
the orthogonalized wrist-to-interaction direction define the tool's closing
and approach axes, with signs fixed across the episode. These targets refer
to the robot's nominal pinch-center TCP. Together
with the grasp schedule, they specify the task-space reference that the
retargeter must track. Figure~\ref{fig:ego_pipeline} summarizes this
construction; Appendix~\ref{app:ego_geometry} gives the coordinate definitions.

\paragraph{Kinematically constrained retargeting.}
The same tool motion can be realized by different robot joint configurations,
and independent framewise inverse kinematics can introduce abrupt joint
changes. We therefore optimize the 14 arm joints over the full sequence,
balancing tool tracking, reference-posture regularization, and temporal
smoothness. The tool targets and grasp schedule remain fixed, so optimization
changes how the robot realizes the reference without redefining that reference.

Grasp timing also determines orientation tolerance. Tighter tolerances during
closure prioritize the closing and approach directions; looser tolerances
when open allow posture adjustment. A real-robot reference regularizes
redundant shoulder--elbow and wrist configurations, while temporal penalties
control velocity, acceleration, and jerk on the original 30~Hz timeline.
Joint and derivative bounds constrain the solve, and workspace and collision
checks assess the resulting trajectory. Appendix~\ref{app:ego_geometry}
details the objective and optimization procedure.

\paragraph{Observation--action supervision.}
Each episode pairs the captured human images with robot end-effector
trajectories and inferred gripper commands on a common timeline. Forward
kinematics (FK) recomputes the end-effector trajectories from the optimized
joint reference using the robot data's tool convention. This pairing supplies
the correspondence between visual evolution and motion needed for EWAM's
joint video--action learning. In the co-training setting of
Section~\ref{sec:human_cotraining}, human and robot episodes use the same
sixteen-dimensional end-effector action interface.
All reconstruction and retargeting are performed offline.
Figure~\ref{fig:human_robot_replay} illustrates
demonstration replay, while Sections~\ref{sec:ego_pipeline_evaluation}
and~\ref{sec:human_cotraining} evaluate retargeting fidelity and learned-policy
performance, respectively.

\FloatBarrier

\subsection{Retargeting Quality and Robot Replay}
\label{sec:ego_pipeline_evaluation}

We assess the offline retargeting pipeline on five recorded human
demonstration sequences, examining compliance with the specified
kinematic checks and positional agreement with the human-derived tool
targets. TCP position error is computed between the positions obtained
by forward kinematics from the retargeted joint references and the
corresponding tool targets. We report the 95th percentile (P95) of this
distance separately for the left and right tools. These measurements
characterize the offline references rather than measured tracking error
during physical execution. As shown in
Table~\ref{tab:ego_alignment_results}, all five references pass the
constraint and consistency checks defined in
Appendix~\ref{app:ego_geometry}. Nevertheless, the largest per-sequence
P95 errors are 77.01~mm for the left tool and 19.69~mm for the right tool,
indicating residual target mismatch despite compliance with the
specified checks.

\begin{table}[!htbp]
  \centering
  \caption{\textbf{Kinematic checks and TCP position errors on five
  retargeted demonstrations.}
  TCP P95 measures the distance between FK-derived reference positions
  and human-derived tool targets. ``Pass'' indicates satisfaction of
  the specified constraint and consistency checks.}
  \label{tab:ego_alignment_results}
  \small
  \setlength{\tabcolsep}{6pt}
  \begin{tabular}{@{}lrrc@{}}
    \toprule
    Session & \multicolumn{2}{c}{TCP P95 (mm)} & Checks \\
    \cmidrule(lr){2-3}
            & Left & Right & \\
    \midrule
    013312 & 56.79 &  2.92 & Pass \\
    013326 & 77.01 & 14.65 & Pass \\
    013335 & 42.53 &  1.66 & Pass \\
    013352 & 18.63 &  1.30 & Pass \\
    222650 &  4.07 & 19.69 & Pass \\
    \bottomrule
  \end{tabular}
\end{table}

\begin{figure}[t]
  \centering
  \includegraphics[width=\linewidth]{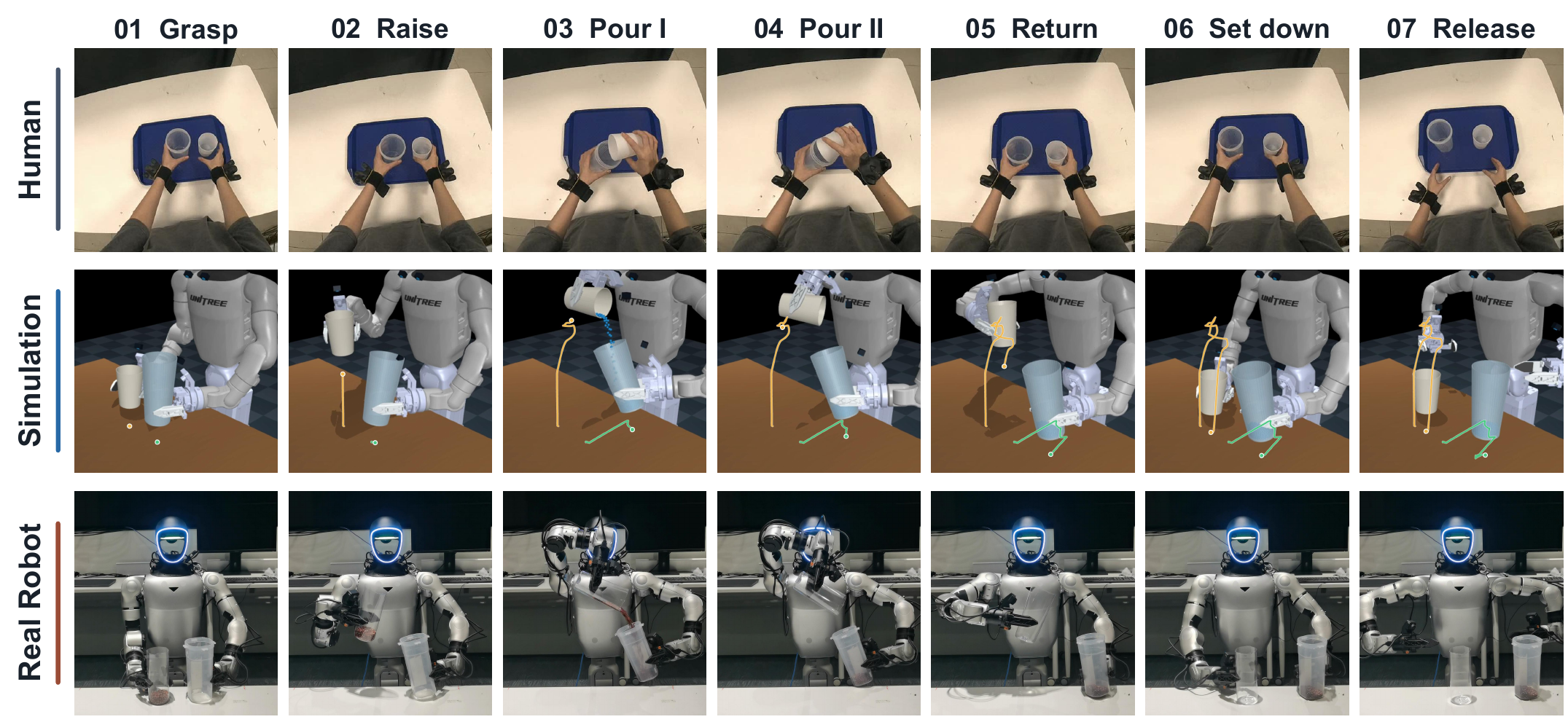}
  \caption{\textbf{Human demonstration and robot replay.}
  Rows show the human egocentric recording, simulation replay,
  and physical Unitree~G1-D replay. Columns correspond to seven
  visually matched action phases: grasp, raise, two pouring stages,
  return, placement, and release. Yellow and green curves show
  projected trajectories of the source and receiver cup bases
  in simulation, respectively. The simulation represents the
  transferred material using rigid particles.}
  \label{fig:human_robot_replay}
\end{figure}

As a qualitative illustration, Figure~\ref{fig:human_robot_replay}
shows a recorded bimanual pouring demonstration adapted for replay
in simulation and on the physical Unitree~G1-D robot. The comparison
aligns seven visible action phases, from grasping through pouring
to placement and release, providing a view of the demonstrated
action sequence across human and robot embodiments. The robot
sequences are replays of demonstration-derived trajectories.
The effect of human demonstrations on learned-policy performance
is evaluated separately in Section~\ref{sec:human_cotraining}.

\FloatBarrier
\section{Experiments}

\subsection{Experimental Setup}

We evaluate EWAM across simulation, cross-embodiment transfer, and real-world
deployment. In simulation, we use RoboTwin~2.0~\citep{chen2026robotwin2}
and LIBERO~\citep{added_liu2023libero}, reporting task success rate as the
primary metric. Baselines are grouped into VLA, WAM, and hybrid VLA--WAM
paradigms.

RoboTwin experiments follow three distinct evaluation protocols. First, the
clean-to-random protocol adapts a single policy on clean demonstrations of all
50 tasks and evaluates it under both clean and randomized conditions. Second,
the in-domain protocol adapts and evaluates each method on both clean and
randomized demonstrations. Third, the cross-embodiment protocol trains on four
robot platforms and evaluates on Aloha-Agilex-2 to isolate transfer from human
egocentric pretraining.

The checkpoints used for the C2R and in-domain evaluations are initialized
from cross-embodiment robot pretraining. The checkpoints used in the human-pretraining
experiments in Section~\ref{sec:human_transfer} are obtained from human
egocentric pretraining. Appendix~\ref{app:data_compute} provides details of
both data sources.
LIBERO evaluation covers its four standard task suites.
Real-world evaluation covers a Franka single arm, a Dobot dual-arm
platform, and a G1-D dual-arm humanoid.
We further evaluate subtask-phase supervision on three
instruction-conditioned block-manipulation tasks and a Unitree G1-D
\textsc{Dual-Basket Sorting} sequence in
Section~\ref{sec:subtask_progress}
(Appendix~\ref{app:long_horizon}).
Appendix~\ref{app:latency} reports end-to-end inference latency and the
optimizations behind it.

\subsection{RoboTwin Clean-to-Random Evaluation}
\label{sec:robotwin_c2r}

In our primary simulation comparison, a single policy is initialized from a
cross-embodiment robot-pretrained checkpoint, adapted on clean demonstrations
of all 50 tasks, and then evaluated under domain randomization. No randomized
data from the target embodiment are used during downstream adaptation.
Table~\ref{tab:robotwin_c2r} compares representative VLA, WAM, and hybrid
VLA--WAM policies under both clean-to-clean (C2C) and clean-to-random (C2R)
evaluation
~\citep{added_starvla2026,galaxea2025g0,
cai2026xiaomi,zheng2025xvla,added_yang2026abot,black2025pi05,
yuan2026fastwam,added_zhang2026imagewam,guo2026xwam,
yang2026fourdwam,gigabrainteam2026gigabrain07}.

EWAM reaches 82.2\% C2C and 72.1\% C2R, exceeding GigaBrain-0.7 by 4.2
C2R points and $\pi_{0.5}$ by 26.1. Overall, EWAM achieves the highest C2R
and average scores, generalizing well from clean to randomized scenes.
Because pretraining data differ across
methods, this comparison alone does not isolate the contribution of the
attention architecture.

\begin{table}[t]
  \caption{\textbf{RoboTwin~2.0 clean-to-random performance.} Policies are adapted on
  clean demonstrations only. Average is the mean of C2C and C2R.}
  \label{tab:robotwin_c2r}
  \centering
  \begin{tabular}{llccc}
    \toprule
    Paradigm & Model & C2C (\%) & C2R (\%) & Average (\%) \\
    \midrule
    \multirow{6}{*}{\textit{VLA}}
      & starVLA~\citep{added_starvla2026} & 46.52 & 3.16 & 24.84 \\
      & GalaxeaVLA~\citep{galaxea2025g0} & 62.70 & 12.72 & 37.71 \\
      & Xiaomi Robotics-0~\citep{cai2026xiaomi} & 62.90 & 18.20 & 40.55 \\
      & X-VLA~\citep{zheng2025xvla} & 68.00 & 20.90 & 44.45 \\
      & ABot-M0~\citep{added_yang2026abot} & 57.40 & 30.36 & 43.88 \\
      & $\pi_{0.5}$~\citep{black2025pi05} & 70.70 & 46.00 & 58.35 \\
    \midrule
    \multirow{4}{*}{\textit{WAM}}
      & Fast-WAM~\citep{yuan2026fastwam} & 77.80 & 1.90 & 39.85 \\
      & ImageWAM~\citep{added_zhang2026imagewam} & 84.4 & 18.3 & 51.35 \\
      & X-WAM~\citep{guo2026xwam} & 70.00 & 25.80 & 47.90 \\
      & 4D-WAM~\citep{yang2026fourdwam} & 81.5 & 41.8 & 61.6 \\
    \midrule
    \multirow{2}{*}{\textit{VLA+WAM}}
      & GigaBrain-0.7~\citep{gigabrainteam2026gigabrain07} & 66.8 & 67.9 & 67.3 \\
      & \textbf{EWAM (Ours)} & \textbf{82.2} & \textbf{72.1} & \textbf{77.2} \\
    \bottomrule
  \end{tabular}
\end{table}

\subsection{RoboTwin In-Domain Results}
\label{sec:robotwin_main}

We additionally report the in-domain setting, in which every method, including
ours, is trained on both clean and randomized demonstrations and evaluated under
each condition. We organize the baselines into VLA, WAM, and hybrid VLA+WAM
paradigms. The VLA comparisons
cover recent open and large-scale policies
~\citep{added_yang2026abot,added_wu2026lingbotvla,
added_zhang2026joyai,added_zhang2026hyvla,added_li2026aceego}. The WAM group
comprises policies centered on predictive pixel-space or latent dynamics
representations~\citep{added_gigaai2026gigaworld,added_chen2026lawam,
li2026lingbot,yuan2026fastwam}. The hybrid VLA+WAM group comprises systems that
combine VLA and WAM components
~\citep{bi2026motus,added_team2026internvla,added_yang2026wla0}.

EWAM achieves the highest average, 92.9\%, with 93.0\% on clean scenes and
92.8\% under randomization. Relative to the strongest baseline in each
paradigm, it improves average success by 2.0 points over ACE-Ego-0 among
VLAs, 0.7 points over LingBot-VA among WAMs, and 1.4 points over WLA-0 among
hybrid methods. Its 0.2-point gap between clean and randomized scores
indicates consistent performance across conditions. Together
with the clean-to-random results in Section~\ref{sec:robotwin_c2r}, these
results show that EWAM preserves strong in-domain performance while maintaining
randomized performance after clean-only downstream adaptation.

\begin{table}[t]
  \caption{\textbf{RoboTwin~2.0 in-domain results.} All methods are adapted on both
  clean and randomized demonstrations. Average is the mean of Clean and
  Randomized.}
  \label{tab:robotwin_indomain}
  \centering
  \begin{tabular}{llccc}
    \toprule
    Paradigm & Model & Clean (\%) & Randomized (\%) & Average (\%) \\
    \midrule
    \multirow{6}{*}{\textit{VLA}}
      & $\pi_{0.5}$~\citep{black2025pi05} & 82.7 & 76.8 & 79.8 \\
      & ABot-M0~\citep{added_yang2026abot} & 86.1 & 85.1 & 85.6 \\
      & LingBot-VLA~\citep{added_wu2026lingbotvla} & 88.6 & 86.7 & 87.7 \\
      & JoyAI-RA~\citep{added_zhang2026joyai} & 90.5 & 89.3 & 89.9 \\
      & HyVLA-0.5~\citep{added_zhang2026hyvla} & 90.9 & 90.1 & 90.5 \\
      & ACE-Ego-0~\citep{added_li2026aceego} & 91.1 & 90.6 & 90.9 \\
    \midrule
    \multirow{4}{*}{\textit{WAM}}
      & GigaWorld-Policy~\citep{added_gigaai2026gigaworld} & 86.4 & 85.0 & 85.7 \\
      & LaWAM~\citep{added_chen2026lawam} & 92.6 & 89.8 & 91.2 \\
      & LingBot-VA~\citep{li2026lingbot} & 92.9 & 91.5 & 92.2 \\
      & Fast-WAM~\citep{yuan2026fastwam} & 91.9 & 91.8 & 91.9 \\
    \midrule
    \multirow{4}{*}{\textit{VLA+WAM}}
      & Motus~\citep{bi2026motus} & 88.7 & 87.0 & 87.9 \\
      & InternVLA-A1~\citep{added_team2026internvla} & 89.4 & 89.6 & 89.5 \\
      & WLA-0~\citep{added_yang2026wla0} & 92.9 & 90.0 & 91.5 \\
      & \textbf{EWAM (Ours)} & \textbf{93.0} & \textbf{92.8} & \textbf{92.9} \\
    \bottomrule
  \end{tabular}
\end{table}

Without pretraining, the same architecture reaches 83.0\% under the
training-ablation protocol of Table~\ref{tab:layer_intervention}; the
robot-pretrained model reaches 92.8\% under randomization in the in-domain
evaluation. The architecture therefore remains effective from scratch while
benefiting substantially from cross-embodiment pretraining.

\subsection{LIBERO Simulation Results}

We additionally evaluate EWAM on the LIBERO manipulation benchmark
~\citep{added_liu2023libero}, covering spatial, object-centric,
goal-conditioned, and long-horizon task suites. As shown in
Table~\ref{tab:libero_main}, EWAM achieves success rates of 98.6\%, 99.8\%,
98.6\%, and 98.2\% on LIBERO-Spatial, LIBERO-Object, LIBERO-Goal, and
LIBERO-Long, respectively, yielding an overall average of 98.8\%. This is the
highest overall average among the compared policies.

\begin{table*}[t]
  \caption{\textbf{LIBERO simulation results.} Success rates (\%); baselines are from
  published evaluations, and dashes indicate unreported suite-level scores.}
  \label{tab:libero_main}
  \centering
  \begin{tabular}{llccccc}
    \toprule
    Paradigm & Model & Spatial & Object & Goal & Long & Average \\
    \midrule
    \multirow{3}{*}{\textit{VLA}}
      & $\pi_{0.5}$~\citep{black2025pi05} & 98.8 & 98.2 & 98.0 & 92.4 & 96.9 \\
      & ABot-M0~\citep{added_yang2026abot} & 98.8 & 99.8 & \textbf{99.0} & 96.6 & 98.6 \\
      & Qwen-VLA-Instruct~\citep{added_wang2026qwenvla} & -- & -- & -- & -- & 97.9 \\
    \midrule
    \multirow{3}{*}{\textit{WAM}}
      & Fast-WAM~\citep{yuan2026fastwam} & 98.2 & \textbf{100.0} & 97.0 & 95.2 & 97.6 \\
      & LingBot-VA~\citep{li2026lingbot} & 98.5 & 99.6 & 97.2 & \textbf{98.5} & 98.5 \\
      & LaWAM~\citep{added_chen2026lawam} & \textbf{99.4} & 99.6 & 98.4 & 97.0 & 98.6 \\
    \midrule
    \multirow{3}{*}{\textit{VLA+WAM}}
      & Motus~\citep{bi2026motus} & 96.8 & 99.8 & 96.6 & 97.6 & 97.7 \\
      & WLA-0~\citep{added_yang2026wla0} & 99.0 & \textbf{100.0} & 97.8 & 97.6 & 98.6 \\
      & \textbf{EWAM (Ours)} & 98.6 & 99.8 & 98.6 & 98.2 & \textbf{98.8} \\
    \bottomrule
  \end{tabular}
\end{table*}

\subsection{Real-World Evaluation}

We evaluate seven real-world tasks across a Franka single arm, a Dobot
dual-arm platform, and a Unitree~G1-D dual-arm humanoid. As shown in
Tables~\ref{tab:real_robot_franka} and~\ref{tab:real_robot_dual}, EWAM attains
the highest success rate on every task. Relative to
$\pi_{0.5}$~\citep{black2025pi05}, it is higher on six of the seven tasks and
ties on the Dobot towel task. Comparisons with the additional baselines show
the same trend where results are available: EWAM outperforms
Motus~\citep{bi2026motus}, LingBot-VA~\citep{li2026lingbot}, Xiaomi
Robotics-0~\citep{cai2026xiaomi}, and Fast-WAM~\citep{yuan2026fastwam} on
Franka, and outperforms Motus and
Fast-WAM~\citep{yuan2026fastwam} across all Dobot and G1-D tasks. These results demonstrate consistent
real-world performance across single-arm, dual-arm, and humanoid embodiments
and across rearrangement, deformable-object manipulation, and pouring tasks.

\begin{figure}[t]
  \centering
  \begin{minipage}[t]{0.37\linewidth}
    \vspace{0pt}
    \centering
    \captionof{table}{\textbf{Real-world success rates on Franka (\%).}}
    \label{tab:real_robot_franka}
    \vspace{0.6em}
    \resizebox{\linewidth}{!}{%
      \begin{tabular}{@{}lcc@{}}
        \toprule
        Model & Stack Bowls & Object into Box \\
        \midrule
        $\pi_{0.5}$~\citep{black2025pi05} & 70 & 80 \\
        Motus~\citep{bi2026motus} & 60 & 20 \\
        LingBot-VA~\citep{li2026lingbot} & 70 & 60 \\
        Xiaomi Robotics-0~\citep{cai2026xiaomi} & 70 & 70 \\
        Fast-WAM~\citep{yuan2026fastwam} & 60 & 30 \\
        \midrule
        \textbf{EWAM (Ours)} & \textbf{80} & \textbf{90} \\
        \bottomrule
      \end{tabular}%
    }
  \end{minipage}\hfill
  \begin{minipage}[t]{0.60\linewidth}
    \vspace{0pt}
    \centering
    \captionof{table}{\textbf{Real-world success rates on Dobot and Unitree
    G1-D (\%).}}
    \label{tab:real_robot_dual}
    \vspace{0.6em}
    \resizebox{\linewidth}{!}{%
      \begin{tabular}{@{}lccccc@{}}
      \toprule
        & \multicolumn{3}{c}{Dobot} & \multicolumn{2}{c}{Unitree G1-D} \\
        \cmidrule(lr){2-4}\cmidrule(lr){5-6}
        Model & Pour Water & Tidy Desk & Towel & Kettle Pouring & Pour Beans \\
      \midrule
        $\pi_{0.5}$~\citep{black2025pi05} & 30 & 20 & \textbf{40} & 60 & 60 \\
        Motus~\citep{bi2026motus} & 30 & 10 & 30 & 30 & 40 \\
        Fast-WAM~\citep{yuan2026fastwam} & 20 & 70 & 0 & 20 & 30 \\
        \midrule
        \textbf{EWAM (Ours)} & \textbf{70} & \textbf{80} & \textbf{40}
        & \textbf{70} & \textbf{80} \\
      \bottomrule
      \end{tabular}%
    }
  \end{minipage}
\end{figure}

\subsection{Real-Robot Generalization and Failure Localization}
\label{sec:pour_generalization}

We examine generalization and failure stages on G1-D pour-beans. The task
requires grasping both cups, transporting them, aligning the pouring cup over
the receiving cup, pouring, and replacing both cups. We vary cup instances,
object placement, or background and clutter one factor at a time relative to
an in-domain baseline, and score each stage separately. Each condition receives
10 physical trials, for 40 trials in total.

\begin{table}[t]
  \caption{\textbf{Stage-resolved generalization on the G1-D pour-beans task.}
  Ten trials per condition, one factor varied at a time against the in-domain
  baseline. S denotes a trial completed with no stage error, P a trial in which
  at least one stage fails but the episode still runs to completion, and F an
  outright failure. The success rate counts S only. The held-out aggregate
  excludes the in-domain baseline.}
  \label{tab:pour_generalization}
  \centering
  \small
  \begin{tabular}{lccccc}
    \toprule
    Condition & Trials & S & P & F & Success (\%) \\
    \midrule
    In-domain baseline & 10 & 8 & 2 & 0 & 80.0 \\
    \midrule
    Cup instance & 10 & 7 & 3 & 0 & 70.0 \\
    Placement & 10 & 7 & 2 & 1 & 70.0 \\
    Background and clutter & 10 & 6 & 4 & 0 & 60.0 \\
    \midrule
    All held-out conditions & 30 & 20 & 9 & 1 & \textbf{66.7} \\
    \bottomrule
  \end{tabular}
\end{table}

Table~\ref{tab:pour_generalization} reports 8/10 successes in-domain and
20/30 across the held-out conditions (66.7\%), a 13.3-point decrease from
the concurrent baseline. Cup-instance and placement changes each yield 7/10,
while background and clutter changes yield 6/10. Differences between these
conditions amount to one trial, so their ordering should not be overinterpreted.

Failures concentrate near pouring. Across all 40 trials, grasping succeeds
in 39, transport and replacement in 40, alignment in 29, and pouring in 28.
Ten of the twelve incomplete-success trials are annotated with spills
associated with inaccurate cup alignment. The single outright failure occurs
when the robot closes on both cups without grasping either and continues into
the pouring sequence. The stage metrics describe individual motion outcomes;
they are not nested measures of full-task success.

Spills also occur in two in-domain trials, indicating a precision bottleneck
that persists across the tested conditions. These observations motivate
improved alignment and feedback during pouring, while leaving open whether
the underlying limitation arises from perception, action prediction, or
controller tracking.

\subsection{Subtask-Phase Supervision}
\label{sec:subtask_progress}

We next test the subtask-phase prediction heads of
Section~\ref{sec:progress_post_training} as a current-goal signal for
long-horizon control. The same architecture is post-trained with and
without this supervision, then evaluated on three instruction-conditioned
block tasks and on a Unitree G1-D
dual-basket sorting sequence
(Appendix~\ref{app:long_horizon}).
Table~\ref{tab:subtask_success} shows higher later-stage completion and
higher full-episode success on all three simulation tasks:
instructed ranking rises from 83.5\% to 91.0\%,
instructed stacking from 44.5\% to 58.5\%, and
instructed ranking \& stacking from 1.5\% to 16.0\%.
The same supervision raises later-stage completion on
dual-basket sorting, from a mean of $1.0/4$ objects placed without the heads to
$2.2/4$ with them. The gains are consistent with training the policy to
recognize and complete the currently active subtask, rather than with
adding an independently executed planner.

\begin{table}[t]
  \caption{\textbf{Effect of subtask-phase supervision on long-horizon
block tasks.}
Grasp, Place, and Stack are stage-wise rates; Full is
complete-episode success. Dashes mark inapplicable stages.}
  \label{tab:subtask_success}
  \centering
  \small
  \setlength{\tabcolsep}{4.5pt}
  \begin{tabular}{@{}llcccc@{}}
    \toprule
    Task & Setting & Grasp & Place & Stack & Full \\
    \midrule
    \multirow{2}{*}{Instructed Ranking}
      & w/o subtask & 94.0 & 89.5 & -- & 83.5 \\
      & w/ subtask  & 93.5 & 97.5 & -- & \textbf{91.0} \\
    \midrule
    \multirow{2}{*}{Instructed Stacking}
      & w/o subtask & 91.5 & -- & 44.5 & 44.5 \\
      & w/ subtask  & 98.5 & -- & 58.5 & \textbf{58.5} \\
    \midrule
    \multirow{2}{*}{Instructed Ranking \& Stacking}
      & w/o subtask & 60.0 & 37.0 & 24.5 & 1.5 \\
      & w/ subtask  & 63.0 & 51.5 & 39.5 & \textbf{16.0} \\
    \bottomrule
  \end{tabular}
\end{table}

\subsection{Learning from Human Egocentric Data}
\label{sec:human_data_study}

We study human data as an additional source of supervision for EWAM through
three related comparisons. First, we test whether human-pretrained
initialization improves transfer to a held-out robot embodiment. Second, we
compare human- and robot-pretrained initialization under the same real-robot
post-training mixture. Third, without pretraining and with 60 robot episodes
fixed, we vary the number of added egocentric episodes and measure real-robot
robustness under tablecloth and cup changes. The first two comparisons examine
initialization; the third isolates the contribution of co-training data.
These are separate training regimes, rather than successive stages of one
recipe.

Human pretraining uses approximately 2,084 hours from VITRA-1M, EgoDex, and
Xperience, with wrist motion expressed as camera-relative end-effector
displacements.
For real-robot co-training, we instead use task-specific human recordings
retargeted to the G1-D action interface. Section~\ref{sec:human_pretraining}
details the pretraining representation, and Section~\ref{sec:ego_collection}
describes demonstration construction.

\subsubsection{Human Pretraining for Cross-Embodiment Transfer}
\label{sec:human_transfer}

We train on the 50 RoboTwin~2.0 tasks using ARX-X5, UR5, Franka, and Piper,
and evaluate on the held-out Aloha-Agilex-2. The compared policies share the
architecture, downstream robot data, and optimization, and differ only in
whether they use human-pretrained initialization.

\begin{figure}[!htbp]
  \centering
  \resizebox{0.55\linewidth}{!}{%
    \begin{minipage}[b]{0.46\linewidth}
      \centering
      \begin{tikzpicture}[baseline=(current axis.south)]
\begin{axis}[
  width=0.95\linewidth, height=5.3cm,
  title={},
  xlabel={Robot training step},
  ylabel={Cross-embodiment success (\%)},
  ybar,
  bar width=9pt,
  ymin=0, ymax=76,
  symbolic x coords={40K,50K,80K,120K},
  xtick=data,
  ytick={0,20,40,60},
  xmin={[normalized]-0.65}, xmax={[normalized]3.65},
  tick label style={font=\scriptsize}, label style={font=\small},
  axis lines=left,
  axis line style={black!55, line width=0.5pt},
  tick align=outside,
  ymajorgrids=true,
  grid style={black!9, line width=0.4pt},
  legend style={font=\tiny, draw=none, fill=none,
                at={(0.5,1.02)}, anchor=south, legend columns=2,
                /tikz/every even column/.append style={column sep=3pt}},
  legend image code/.code={
    \draw[#1] (0cm,-0.08cm) rectangle (0.16cm,0.08cm);
  },
  legend cell align=center,
  clip=false,
]
  \addplot[
    bar shift=0pt, bar width=16pt,
    fill=MyBlue!18, draw=MyBlue!45, line width=0.6pt,
    nodes near coords={\pgfmathprintnumber[fixed,precision=1]\pgfplotspointmeta},
    nodes near coords style={font=\tiny, scale=0.75,
                              text=MyBlue!80!black,
                              anchor=south, yshift=1pt},
  ] coordinates {(40K,44.3) (50K,53.4) (80K,58.9) (120K,66.9)};
  \addlegendentry{w/ human pretrain}

  \addplot[
    bar shift=0pt, bar width=10pt,
    fill=MyBlue!68, draw=MyBlue!90!black, line width=0.5pt,
    nodes near coords={\pgfmathprintnumber[fixed,precision=1]\pgfplotspointmeta},
    nodes near coords style={font=\tiny, scale=0.75, text=white,
                              anchor=north, yshift=-1pt},
  ] coordinates {(40K,33.5) (50K,37.2) (80K,40.1) (120K,36.2)};
  \addlegendentry{w/o human pretrain}
\end{axis}
\end{tikzpicture}
    \end{minipage}%
  }
  \caption{\textbf{Cross-embodiment transfer to Aloha-Agilex-2.} Success rate
  over the 50 RoboTwin~2.0 tasks at each evaluated robot-training checkpoint,
  with and without human-pretrained initialization. This protocol is not
  comparable to Tables~\ref{tab:robotwin_c2r} and~\ref{tab:robotwin_indomain}.}
  \label{fig:cross_embodiment}
\end{figure}
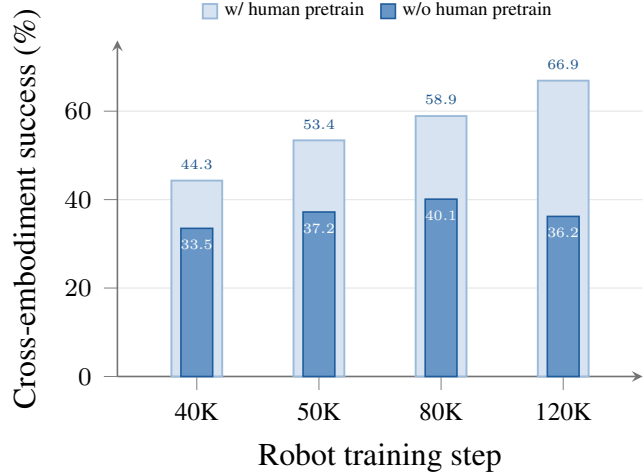

Human pretraining improves cross-embodiment success at every evaluated checkpoint
(Figure~\ref{fig:cross_embodiment}). At
120K robot-training steps, success increases from 36.2\% to 66.9\%, a
30.7-point gain under a matched training budget. Comparing each policy's best
evaluated checkpoint gives 40.1\% versus 66.9\%, a 26.8-point gain.

\subsubsection{Human Pretraining for Real-Robot Co-Training}
\label{sec:human_cotraining}

We next evaluate the same initialization question by co-training Unitree G1-D
on a long-horizon pouring task. Two policies are
post-trained on the same mixture of 60 teleoperated robot episodes and 30
retargeted human episodes. They share the schedule, seed, and evaluation
episodes, and differ only in human- versus robot-pretrained initialization.
A robot-pretrained policy using 90 robot episodes provides a robot-only
reference. All three are evaluated on 30 held-out robot trajectories for
action error and on the physical robot for task success.

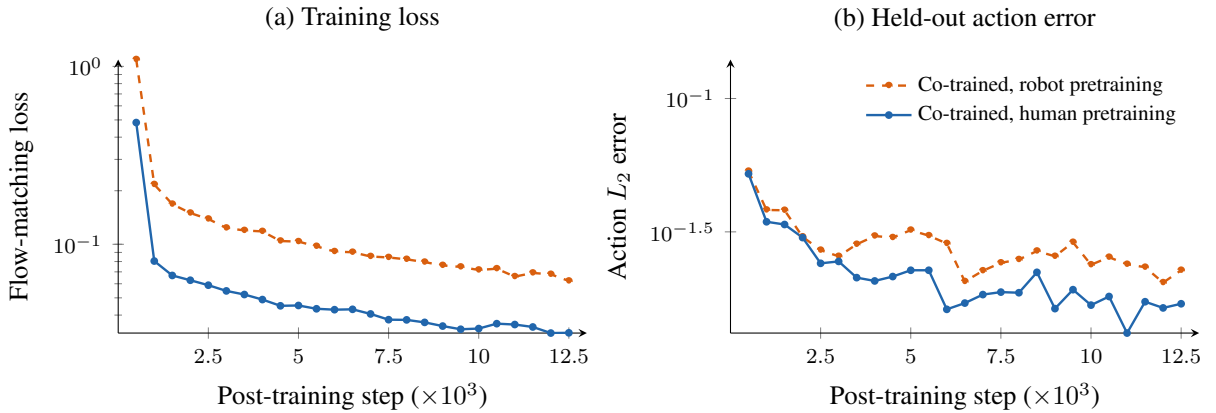
\begin{figure}[!t]
  \centering
  \begin{tikzpicture}[baseline=(current axis.south)]
\begin{groupplot}[
  group style={group size=2 by 1, horizontal sep=1.9cm},
  width=0.46\linewidth, height=5.2cm,
  xlabel={Post-training step ($\times 10^{3}$)},
  xmin=0, xmax=13, xtick={2.5,5,7.5,10,12.5},
  ymode=log, log basis y=10,
  tick label style={font=\scriptsize}, label style={font=\small},
  title style={font=\small, yshift=1pt}, axis lines=left,
  every axis plot/.append style={line width=0.9pt},
  legend style={font=\scriptsize, draw=none, fill=none,
                at={(0.98,0.98)}, anchor=north east},
  legend cell align=left,
]
\nextgroupplot[title={(a) Training loss}, ylabel={Flow-matching loss}]
  \addplot[MyOrange, dashed, mark=*, mark size=1.0pt] coordinates {(0.5,1.0925) (1,0.2168) (1.5,0.1681) (2,0.1498) (2.5,0.1387) (3,0.1235) (3.5,0.1196) (4,0.1181) (4.5,0.1043) (5,0.1035) (5.5,0.0974) (6,0.091) (6.5,0.0902) (7,0.0854) (7.5,0.0843) (8,0.0823) (8.5,0.0794) (9,0.0761) (9.5,0.0746) (10,0.0717) (10.5,0.0729) (11,0.0658) (11.5,0.069) (12,0.0678) (12.5,0.0622)};
  \addplot[MyBlue, mark=*, mark size=1.0pt] coordinates {(0.5,0.4832) (1,0.0805) (1.5,0.0668) (2,0.0628) (2.5,0.0589) (3,0.0547) (3.5,0.0522) (4,0.0489) (4.5,0.0451) (5,0.0453) (5.5,0.0434) (6,0.0429) (6.5,0.0431) (7,0.0406) (7.5,0.0377) (8,0.0376) (8.5,0.0364) (9,0.0347) (9.5,0.0333) (10,0.0336) (10.5,0.0358) (11,0.0354) (11.5,0.0343) (12,0.0317) (12.5,0.0318)};
\nextgroupplot[title={(b) Held-out action error},
              ylabel={Action $L_2$ error}, ymax=0.14]
  \addplot[MyOrange, dashed, mark=*, mark size=1.0pt] coordinates {(0.5,0.0534) (1,0.0382) (1.5,0.0381) (2,0.0303) (2.5,0.027) (3,0.0256) (3.5,0.0284) (4,0.0305) (4.5,0.0301) (5,0.0321) (5.5,0.0306) (6,0.0286) (6.5,0.0206) (7,0.0226) (7.5,0.0242) (8,0.0249) (8.5,0.0268) (9,0.0256) (9.5,0.0289) (10,0.0238) (10.5,0.0254) (11,0.0239) (11.5,0.0233) (12,0.0204) (12.5,0.0227)};
  \addlegendentry{Co-trained, robot pretraining}
  \addplot[MyBlue, mark=*, mark size=1.0pt] coordinates {(0.5,0.0522) (1,0.0345) (1.5,0.0337) (2,0.0301) (2.5,0.0241) (3,0.0245) (3.5,0.0213) (4,0.0207) (4.5,0.0215) (5,0.0227) (5.5,0.0227) (6,0.0162) (6.5,0.0171) (7,0.0184) (7.5,0.0188) (8,0.0187) (8.5,0.0223) (9,0.0163) (9.5,0.0192) (10,0.0168) (10.5,0.0181) (11,0.0132) (11.5,0.0173) (12,0.0164) (12.5,0.017)};
  \addlegendentry{Co-trained, human pretraining}
\end{groupplot}
\end{tikzpicture}
  \caption{\textbf{Effects of human pretraining and human co-training during
  G1-D real-robot post-training.} Both runs are co-trained on the same mixture
  of 60 robot and 30 retargeted human episodes and differ only in
  initialization. (a) Flow-matching loss on the shared post-training data.
  (b) Action $L_2$ error on the 30 held-out G1-D episodes. Both panels use
  logarithmic vertical axes. Human pretraining reduces both errors relative
  to robot pretraining under the same co-training mixture.}
  \label{fig:g1d_cotraining}
\end{figure}

Under the matched 60-robot/30-human mixture, human pretraining increases
real-robot success from 40\% to 60\%
(Table~\ref{tab:g1d_cotraining}). Training loss decreases from 0.068 to
0.034, and held-out action error decreases from 0.0231 to 0.0164
(Figure~\ref{fig:g1d_cotraining}). The 90-robot reference reaches 70\%
success, but changes the data composition and therefore does not isolate the
effect of adding human demonstrations. The next comparison holds the robot
set fixed and varies only the amount of egocentric co-training data.

\begin{table}[t]
  \caption{\textbf{Human pretraining and co-training on G1-D pouring.} Losses are means
  over the final five evaluations; success is measured over 10 physical trials.
  Training loss is comparable only between the two co-trained rows.}
  \label{tab:g1d_cotraining}
  \centering
  \small
  \begin{tabular}{llccc}
    \toprule
    Post-training data & Pretraining & Train loss & Held-out $L_2$
      & Success (\%) \\
    \midrule
    90 robot episodes & Robot & 0.055 & \textbf{0.0098} & \textbf{70} \\
    60 robot + 30 human & Robot & 0.068 & 0.0231 & 40 \\
    60 robot + 30 human & Human & 0.034 & 0.0164 & 60 \\
    \bottomrule
  \end{tabular}
\end{table}

\FloatBarrier

\subsubsection{Effect of Egocentric Co-Training Data}
\label{sec:ego_data_scaling}

Without pretraining and with 60 robot episodes fixed, adding 30 to 1{,}000
egocentric episodes raises
real-robot success under tablecloth and cup changes from 10\% to 60\%
without saturating (Figure~\ref{fig:ego_scaling};
Appendix~\ref{app:ego_generalization}).
Adding just 30 egocentric episodes reaches 20\%, matching robot-only training
with 90 episodes. These results show that low-cost
egocentric data can effectively improve robustness.

\par\smallskip
\noindent\begin{minipage}{\linewidth}
  \centering
  \begin{tikzpicture}[baseline=(current axis.south)]
\begin{axis}[
  width=0.82\linewidth, height=7.2cm,
  axis lines=left,
  axis line style={black!55, line width=0.5pt},
  tick align=outside, tick style={black!55},
  tick label style={font=\small, text=black!75},
  label style={font=\small},
  ylabel={Success rate (\%)},
  ybar, bar width=15pt,
  xtick={0,1,2,3.5,4.5,5.5,6.5},
  xticklabels={30,60,90,{+30},{+60},{+120},{+1k}},
  x tick label style={font=\small, text=black!75},
  xmin=-0.7, xmax=7.2,
  ymin=0, ymax=70, ytick={0,20,40,60},
  ymajorgrids=true, grid style={black!9, line width=0.4pt},
  clip=false,
  nodes near coords={\pgfmathprintnumber\pgfplotspointmeta},
  nodes near coords style={font=\small, anchor=south, yshift=1pt},
]
  \addplot[bar shift=0pt, fill=black!18, draw=black!45, line width=0.5pt,
    nodes near coords style={text=black!70}]
    coordinates {(0,0) (1,10) (2,20)};
  \addplot[bar shift=0pt, fill=MyBlue!30, draw=MyBlue!90!black, line width=0.5pt,
    nodes near coords style={text=MyBlue!90!black}] coordinates {(3.5,20)};
  \addplot[bar shift=0pt, fill=MyBlue!48, draw=MyBlue!90!black, line width=0.5pt,
    nodes near coords style={text=MyBlue!90!black}] coordinates {(4.5,30)};
  \addplot[bar shift=0pt, fill=MyBlue!66, draw=MyBlue!90!black, line width=0.5pt,
    nodes near coords style={text=MyBlue!90!black}] coordinates {(5.5,40)};
  \addplot[bar shift=0pt, fill=MyBlue!85, draw=MyBlue!90!black, line width=0.5pt,
    nodes near coords style={text=MyBlue!90!black}] coordinates {(6.5,60)};
  \draw[black!40, densely dashed, line width=0.6pt]
    (axis cs:-0.5,10) -- (axis cs:7.0,10);
  \node[font=\footnotesize, text=black!60, anchor=south,
    fill=white, inner sep=2pt]
    at (axis cs:2.75,10) {\textit{60 robot only}};
  \draw[black!65, line width=0.5pt]
    (axis cs:2,27) -- (axis cs:2,33) -- (axis cs:3.5,33) -- (axis cs:3.5,27);
  \node[font=\footnotesize, text=black!70, anchor=south, inner sep=2pt]
    at (axis cs:2.75,33) {equal};
  \draw[->, MyBlue!90!black, line width=0.6pt]
    (axis cs:7.15,11) -- (axis cs:7.15,59);
  \node[font=\footnotesize, text=MyBlue!90!black, anchor=west, inner sep=2pt]
    at (axis cs:7.15,35) {\textbf{+50\,pp}};
  \node[font=\small, text=black!70, anchor=north]
    at (axis cs:1,-10) {Robot only};
  \node[font=\small, text=MyBlue!90!black, anchor=north]
    at (axis cs:5,-10) {60 robot + ego};
\end{axis}
\end{tikzpicture}
  
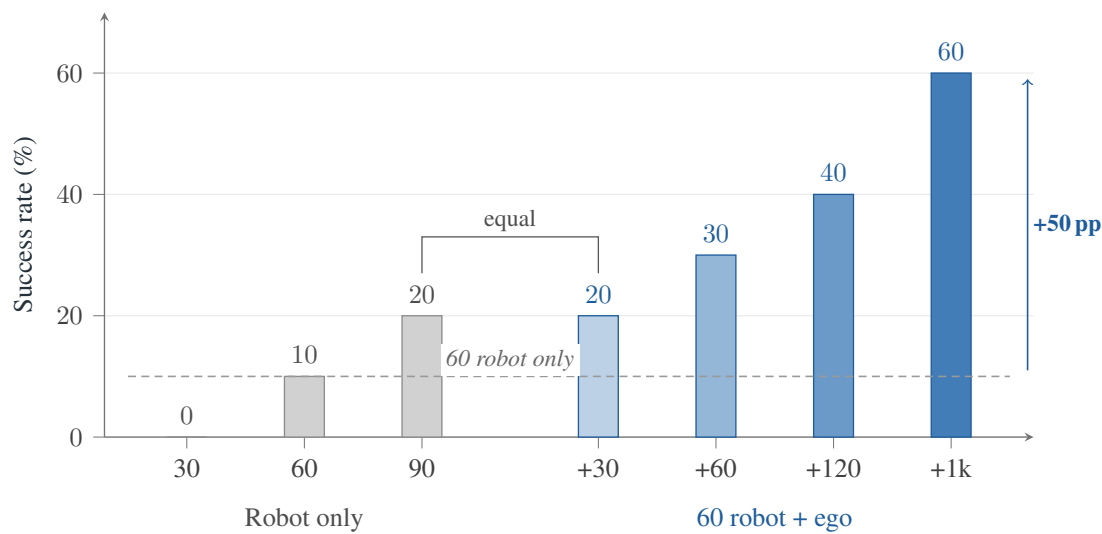
\captionof{figure}{\textbf{Egocentric data scaling under tablecloth and cup changes.}
  Real-robot success without (gray) and with (blue) egocentric data.
  Gray bars train on 30, 60, or 90 robot episodes only; blue bars add 30 to
  1{,}000 egocentric episodes to 60 robot episodes. All runs omit EWAM human
  and robot pretraining; each configuration is evaluated over 10 physical
  trials (Appendix~\ref{app:ego_generalization}).}
  \label{fig:ego_scaling}
\end{minipage}
\par\smallskip

\section{Discussion}

\paragraph{From a fixed handoff to adaptive computation.}
The emergent depth-wise handoff suggests that semantic grounding, visual
foresight, and action refinement need not follow a fixed global schedule.
Different tasks and execution phases may require different amounts of each
form of computation. Ambiguous instructions may demand prolonged semantic
grounding, uncertain interactions may require more predictive computation,
and well-specified motions may permit an earlier transition to action
refinement. Future models could therefore learn task- and state-conditioned
routing that dynamically shifts the transition between experts.

\paragraph{From visual foresight to counterfactual planning.}
The counterfactual future-feature intervention provides a further direction.
Changing the content of intermediate future representations alters spatial
targeting and arm selection, indicating that these representations carry
control-relevant information rather than serving only as an auxiliary
prediction target. The current model produces a future trajectory jointly
with its action prediction, but does not explicitly compare alternative
outcomes. A natural extension is to generate multiple action-conditioned
futures and evaluate their consequences before committing to an action.
This would move unified world--action modeling from a single predictive
trajectory toward counterfactual planning, while retaining semantic grounding
from the vision-language expert.

\paragraph{Limitations and future directions.}
Our analysis characterizes information routing rather than the complete
contents of the learned representations. Attention allocation identifies
where action queries retrieve information, and the interventions establish
functional dependence on selected sources, but neither provides a complete
causal decomposition of the computation. Representation probing, activation
patching, and cross-layer causal mediation could test whether semantic and
predictive information itself changes together with the observed routing.
Finally, although the deployed system achieves real-time operation through
chunked execution, maintaining three large experts remains computationally
expensive. The layer-wise division of labor observed in this work suggests a
route to structured optimization. Rather than compressing all three experts
uniformly, future work could optimize computation over the expert--depth grid:
model capacity could be allocated according to the measured contribution of
each expert at different depths, while structured pruning or parameter sharing
could target regions with consistently low functional contribution. These
improvements could reduce the model's computational overhead while preserving
its functional organization.

\section{Conclusion}

We present EWAM, a unified embodied model in which action queries read from
vision-language and video experts through asymmetric attention. Its
computation organizes into a depth-wise handoff from semantic grounding,
through predicted visual futures, to action refinement. This organization
emerges during training, and layer-targeted masking and counterfactual
injection show that action generation depends on it. Across simulation
benchmarks and real-robot embodiments, EWAM exceeds strong VLA, WAM, and hybrid
baselines. Low-cost human egocentric data further improve cross-embodiment
transfer and robustness, and subtask-phase supervision improves long-horizon
completion. These results suggest that how unified policies coordinate
semantic and predictive computation offers a principled path toward more
capable embodied models.

\clearpage
\section*{Authors}
\begingroup
\raggedright
\small
\textbf{Core Contributors}\\[0.25em]
Hao Wang\textsuperscript{1,2},
Jiajun Wen\textsuperscript{3},
Jingzhi Liu\textsuperscript{1},
Shuoshuo Xue\textsuperscript{1},
Zhiliang Chen\textsuperscript{2,4}

\vspace{0.55em}
\textbf{Contributors}\\[0.25em]
Min Lin\textsuperscript{1},
Yicheng Chang\textsuperscript{1},
Xiaoyu Guo\textsuperscript{5},
Yukang Zhuo\textsuperscript{1},
Zheng Chong\textsuperscript{1},
Yunshuang Nie\textsuperscript{3},
Jian Zhang\textsuperscript{6},
Weijia Liufu\textsuperscript{1},
Qingman Wu\textsuperscript{1},
Heming Xu\textsuperscript{7},
Bingchang Song\textsuperscript{8},
Dantong Wu\textsuperscript{9},
Zhiyuan Wang\textsuperscript{9}

\vspace{0.55em}
\textbf{Guidance Group}\\[0.25em]
Hang Xu\textsuperscript{9},
Jianhua Han\textsuperscript{9},
Bokui Chen\textsuperscript{3},
Shen Zhao\textsuperscript{1}

\vspace{0.55em}
\textbf{Project Leaders}\\[0.25em]
Rui Li\textsuperscript{9},
Xiaodan Liang\textsuperscript{1,9}

\vspace{0.8em}
\footnotesize
\textsuperscript{1}Sun Yat-sen University\\
\textsuperscript{2}Beijing Zhongguancun Academy\\
\textsuperscript{3}Tsinghua University\\
\textsuperscript{4}Fudan University\\
\textsuperscript{5}South China University of Technology\\
\textsuperscript{6}Mohamed bin Zayed University of Artificial Intelligence\\
\textsuperscript{7}Northwest University, Xi'an\\
\textsuperscript{8}Southern University of Science and Technology\\
\textsuperscript{9}Yinwang Intelligent Technology Co. Ltd.
\par
\endgroup

\bibliography{iclr2027_conference,references_added,ewam_references}
\bibliographystyle{unsrtnat}

\clearpage
\appendix
\pretocmd{\section}{\Needspace{8\baselineskip}}{}{}
\pretocmd{\subsection}{\Needspace{6\baselineskip}}{}{}
\renewcommand{\topfraction}{0.9}
\renewcommand{\textfraction}{0.08}
\renewcommand{\floatpagefraction}{0.85}

\section{Cross-Paradigm Attention Measurement}
\label{app:baseline_attention}

Figure~\ref{fig:motivation} compares action-query attention routing across four
policies. The measurement details are as follows.

\paragraph{Checkpoints.}
All baseline weights are publicly released. Figure~\ref{fig:motivation}(b) uses the released
Fast-WAM~\citep{yuan2026fastwam} weights and Figure~\ref{fig:motivation}(c) the released
Motus~\citep{bi2026motus} weights. Figure~\ref{fig:motivation}(a) uses the $\pi_{0.5}$ baseline
checkpoint distributed with Motus, that is, a $\pi_{0.5}$ policy trained on
RoboTwin~2.0 by the Motus authors. Figure~\ref{fig:motivation}(d) uses the in-domain EWAM checkpoint
of Table~\ref{tab:robotwin_indomain}, which is also the checkpoint used for the
inference-time interventions of Table~\ref{tab:layer_intervention}.

\paragraph{Evaluation condition.}
Every panel is measured on the same single RoboTwin~2.0 \textsc{Adjust Bottle}
episode under the randomized protocol, at denoising step~0. Attention
allocation is stable across tasks and denoising steps, so a single episode at
one denoising step is representative
(Appendix~\ref{app:cross_task_attention}). The horizontal axis
is the policy inference step, that is, the successive action-chunk queries
within the episode; the number of steps differs across panels because the four
policies complete the episode in different numbers of chunks.

\paragraph{Key grouping.}
For each model we assign every attended key to one of the sources that model
possesses and apply Eq.~\plaineqref{eq:attention_allocation} over those
groups. The available sources differ by architecture: $\pi_{0.5}$ has no video
stream, and Fast-WAM has no separate vision-language stream.

\section{Model and Training Details}
\label{app:model_details}

\subsection{Architecture}

EWAM contains three depth-aligned transformer experts for semantic
understanding, future-visual prediction, and action generation. The
vision-language expert receives the language instruction and current image. The
video expert receives a latent representation of the observed condition frame
and noisy future-video latents. The action expert receives a noisy action chunk. Corresponding blocks
from the three experts are executed at the same network depth, allowing the
action representation to aggregate information progressively rather than
waiting for a separately generated visual subgoal.

The attention mask is asymmetric. Vision-language queries attend only within
the vision-language stream, and video queries attend only within the video
stream. Action queries may attend to vision-language, video, and action
keys. This design prevents the auxiliary generative stream from overwriting
semantic representations while making both sources directly available to
action computation. It also gives a well-defined quantity for measuring how
action queries allocate attention among modalities at each depth.

\paragraph{VL layer alignment and recurrent fusion.}
The 28 Qwen3-VL~\citep{bai2025qwen3vl} hidden states are extracted in a single Qwen3-VL forward pass before
the 30-block unified computation. Writing the original Qwen3-VL outputs as
$(\bar q_0^L,\ldots,\bar q_{27}^L)$, the aligned sequence is
\begin{equation}
  (q_0^L,\ldots,q_{29}^L)
  =
  (\bar q_0^L,\ldots,\bar q_{27}^L,\bar q_{26}^L,\bar q_{27}^L).
  \label{eq:vlm_layer_alignment}
\end{equation}
At each unified block, Eq.~\plaineqref{eq:vlm_layer_fusion} combines the
aligned Qwen3-VL state with the accumulated VL state. The normalized fused
tokens are mapped by block-specific query, key, and value projections from
2{,}048 dimensions to 24 heads of width 128. Video, action, and VL tensors are
then concatenated and processed by one masked-attention call. For the VL
query rows, the mask retains only VL keys; the same VL keys and values remain
visible to action queries.

The VL-side attention result is mapped back to 2{,}048 dimensions by a
block-specific output projection and applied through a residual update,
\begin{equation}
  z_l^L =
  x_l^L +
  W_{O,l}^L\,\operatorname{Attn}_l^L(x_l^L).
  \label{eq:vlm_residual_update}
\end{equation}
Here $\operatorname{Attn}_l^L$ denotes the VL-query rows of the shared masked
attention operation. The resulting $z_l^L$ is carried to the next fusion step.
No additional VL feed-forward layer follows this update because each
$q_l^L$ already includes the native Qwen3-VL layer's feed-forward computation.

The implementation used in our analysis contains 30 aligned transformer
blocks, indexed from 0 to 29. Based on the observed attention transitions, we
refer to blocks 0--9 as the shallow stage, blocks 10--19 as the intermediate
stage, and blocks 20--29 as the deep stage. These equal-width ranges are
descriptive partitions rather than architecturally hard-coded modules.

\subsection{Training Objectives}

Training follows two stages. During cross-embodiment pretraining, EWAM jointly
optimizes future-video and action flow-matching losses. Independent noise levels
are sampled for video and action targets so that the model cannot solve one
modality by relying on an identical denoising state in the other. Padded action
dimensions introduced by cross-embodiment alignment are masked from the loss.
The base objective is
\begin{equation}
  \mathcal{L}_{\mathrm{base}}
  = \lambda_v \mathcal{L}_{\mathrm{video}}
  + \lambda_a \mathcal{L}_{\mathrm{action}}.
\end{equation}

Task-specific post-training retains both base losses and adds subtask-phase
supervision. Annotated phases of each demonstration supply a textual label of
the active subtask and a normalized phase index in $[0,1]$ derived from the
same partition:
\begin{equation}
  \mathcal{L}_{\mathrm{post}}
  = \mathcal{L}_{\mathrm{base}}
  + \lambda_p \mathcal{L}_{\mathrm{prog}}
  + \lambda_s \mathcal{L}_{\mathrm{sub}}.
\end{equation}
The scalar is regressed from the action-side representation; the textual
label identifies the currently active semantic phase. Both terms are applied
only during task-specific post-training.

\section{Token-Count-Normalized Attention Analysis}
\label{app:token_normalization}

The allocation $R_l^m$ in Eq.~\plaineqref{eq:attention_allocation} measures the total attention mass
received by source $m$. Because the five sources contain different numbers of
keys, we additionally measure attention enrichment relative to a uniform-token
reference:
\begin{equation}
  E_l^m =
  \frac{R_l^m}
  {|\mathcal{K}_m|/\sum_j|\mathcal{K}_j|},
\end{equation}
where $E_l^m>1$ indicates that a source receives more attention than expected
if attention were distributed uniformly over all available keys. We compute
this statistic on one representative sequence from each of the 50
RoboTwin~2.0 tasks, using the robot-pretrained in-domain checkpoint of
Table~\ref{tab:robotwin_indomain}. Every sequence contains 120 condition-frame, 240
future-frame, 21 action, and 120 VL-image tokens, while VL-text length varies
from 59 to 85 tokens. We therefore compute $E_l^m$ using each task's own token
counts, average over its five measured denoising steps, and then macro-average
with equal task weight.

Table~\ref{tab:token_normalized_visual_handoff} reports both total attention
mass and token-normalized enrichment for the two visual sources involved in
the observed handoff. VL-image tokens are strongly enriched in shallow layers
but decline rapidly with depth, while future-frame attention rises. In the
cross-task macro average, future-frame raw mass first exceeds VL-image mass at
layer 13, and token-normalized future enrichment first exceeds image enrichment
at layer 14. The handoff therefore persists after controlling for the
two-to-one difference in future-frame and VL-image token counts.

\begin{table*}[t]
  \caption{\textbf{Cross-task visual handoff before and after token-count
  normalization.}
  Each entry is the macro mean over 50 RoboTwin~2.0 tasks with its 95\%
  task-bootstrap confidence interval in brackets. We report individual layers
  around the transition and averages outside that interval. Enrichment is
  measured relative to a task-specific uniform-token reference.}
  \label{tab:token_normalized_visual_handoff}
  \centering
  \small
  \setlength{\tabcolsep}{4.5pt}
  \begin{tabular}{lcccc}
    \toprule
    & \multicolumn{2}{c}{VL-image}
    & \multicolumn{2}{c}{Future frame} \\
    \cmidrule(lr){2-3}\cmidrule(lr){4-5}
    Layers & Mass (\%) & Enrichment & Mass (\%) & Enrichment \\
    \midrule
    0--9 & 36.75 [36.45, 37.05] & 1.74 [1.73, 1.76] &
    9.58 [9.49, 9.66] & 0.23 [0.23, 0.23] \\
    10 & 27.99 [27.72, 28.26] & 1.33 [1.32, 1.34] &
    7.42 [7.34, 7.51] & 0.18 [0.17, 0.18] \\
    11 & 22.42 [22.22, 22.60] & 1.06 [1.06, 1.07] &
    17.79 [17.64, 17.94] & 0.42 [0.42, 0.43] \\
    12 & 21.48 [21.30, 21.66] & 1.02 [1.01, 1.03] &
    21.15 [20.89, 21.42] & 0.50 [0.50, 0.51] \\
    13 & 13.37 [13.15, 13.61] & 0.63 [0.62, 0.65] &
    24.85 [24.65, 25.05] & 0.59 [0.58, 0.59] \\
    14 & 11.09 [10.94, 11.24] & 0.53 [0.52, 0.53] &
    28.96 [28.47, 29.47] & 0.69 [0.67, 0.70] \\
    15 & 6.96 [6.80, 7.12] & 0.33 [0.32, 0.34] &
    23.37 [23.06, 23.72] & 0.55 [0.55, 0.56] \\
    16--29 &
    1.31 [1.28, 1.34] & 0.06 [0.06, 0.06] &
    11.79 [11.42, 12.19] & 0.28 [0.27, 0.29] \\
    \bottomrule
  \end{tabular}
\end{table*}

\subsection{Checkpoint Evolution of the Visual Handoff}
\label{app:checkpoint_evolution}

\begin{figure*}[t]
  \centering
  \begin{tikzpicture}
\begin{groupplot}[
  group style={group size=3 by 1, horizontal sep=1.55cm},
  width=0.33\linewidth, height=4.2cm,
  xmode=log, xlabel={Training step}, xmin=35, xmax=90000,
  xtick={100,1000,10000}, xticklabels={$10^2$,$10^3$,$10^4$},
  tick label style={font=\scriptsize}, label style={font=\scriptsize},
  title style={font=\scriptsize, yshift=1pt}, axis lines=left,
  unbounded coords=jump,
  every axis plot/.append style={line width=0.9pt},
]
\nextgroupplot[title={(a) Action stream, shallow}, ylabel={Attention mass (\%)}, ymin=0, ymax=70, ytick={0,20,40,60}, legend style={font=\scriptsize, draw=none, fill=none, at={(0.5,-0.32)}, anchor=north, legend columns=2, column sep=0.25cm}]
  \addplot[SrcCond, mark=*, mark size=1.1pt] coordinates {(50,18.58) (100,15.34) (150,10.78) (200,5.523) (250,2.856) (300,2.277) (350,1.862) (400,1.916) (600,1.861) (1000,1.887) (5000,1.255) (40000,4.291)};
  \addlegendentry{Condition frame}
  \addplot[SrcFuture, mark=*, mark size=1.1pt] coordinates {(50,16.88) (100,23.62) (150,22.29) (200,11.46) (250,6.137) (300,4.369) (350,3.677) (400,3.668) (600,3.516) (1000,3.745) (5000,3.043) (40000,9.075)};
  \addlegendentry{Future frames}
  \addplot[SrcAction, mark=*, mark size=1.1pt] coordinates {(50,17.44) (100,23.18) (150,21.06) (200,24.04) (250,23.19) (300,19.73) (350,18.83) (400,17.49) (600,10.97) (1000,7.086) (5000,4.125) (40000,10.59)};
  \addlegendentry{Action}
  \addplot[SrcText, mark=*, mark size=1.1pt] coordinates {(50,2.944) (100,14.32) (150,22.62) (200,37.11) (250,45.8) (300,49.76) (350,52.18) (400,54.81) (600,58.05) (1000,59.71) (5000,53.38) (40000,38.65)};
  \addlegendentry{VL-text}
  \addplot[SrcImage, mark=*, mark size=1.1pt] coordinates {(50,44.15) (100,23.55) (150,23.25) (200,21.87) (250,22.02) (300,23.87) (350,23.45) (400,22.12) (600,25.6) (1000,27.58) (5000,38.2) (40000,37.39)};
  \addlegendentry{VL-image}
\nextgroupplot[title={(b) VL-image enrichment}, ylabel={Enrichment $E$}, ymode=log, ymin=0.025, ymax=6, ytick={0.1,1}, yticklabels={$0.1$,$1$}, legend style={font=\scriptsize, draw=none, fill=none, at={(0.5,-0.32)}, anchor=north, legend columns=1, column sep=0.25cm}]
  \addplot[black!35, dotted, line width=0.7pt, forget plot] coordinates {(35,1) (90000,1)};
  \addplot[SrcImage, mark=*, mark size=1.1pt] coordinates {(50,2.071) (100,1.109) (150,1.097) (200,1.024) (250,1.042) (300,1.124) (350,1.11) (400,1.041) (600,1.205) (1000,1.296) (5000,1.792) (40000,1.751)};
  \addlegendentry{shallow (0--9)}
  \addplot[SrcImage!80, dashed, mark=*, mark size=1.1pt] coordinates {(50,2.326) (100,0.7532) (150,0.606) (200,0.2468) (250,0.3117) (300,0.3459) (350,0.2561) (400,0.2148) (600,0.1958) (1000,0.2142) (5000,0.4865) (40000,0.04101)};
  \addlegendentry{deep (20--29)}
\nextgroupplot[title={(c) Crossover persistence}, ylabel={Layers, future $>$ image (\%)}, ymin=0, ymax=105, ytick={0,25,50,75,100}, legend style={font=\scriptsize, draw=none, fill=none, at={(0.5,-0.32)}, anchor=north, legend columns=1, column sep=0.25cm}]
  \addplot[MyOrange, mark=*, mark size=1.1pt] coordinates {(50,nan) (100,71.43) (150,84.21) (200,94.74) (250,88.89) (300,77.78) (350,77.78) (400,88.89) (600,62.5) (1000,56.25) (5000,nan) (40000,100)};
  \addlegendentry{token-normalized}
\end{groupplot}
\end{tikzpicture}
  \caption{\textbf{Checkpoint evolution of depth-wise attention routing under
  the asymmetric mask.} Stage-level quantities across the trajectory of
  Table~\ref{tab:checkpoint_evolution}, laid out to match
  Figure~\ref{fig:symmetric_dynamics} so that the two attention designs can be
  compared panel by panel. (a) Shallow-stage action attention over the five key
  sources. (b) Token-normalized VL-image enrichment at the two ends of the
  network, with the dotted line marking the uniform-token reference $E=1$.
  (c) Fraction of layers at or
  beyond the stable crossover for which token-normalized future-frame enrichment
  exceeds VL-image enrichment, with the curve broken at checkpoints that have
  no stable crossover. All circles belong to the same thirty-two-GPU training
  trajectory.}
  \label{fig:checkpoint_attention_full}
\end{figure*}
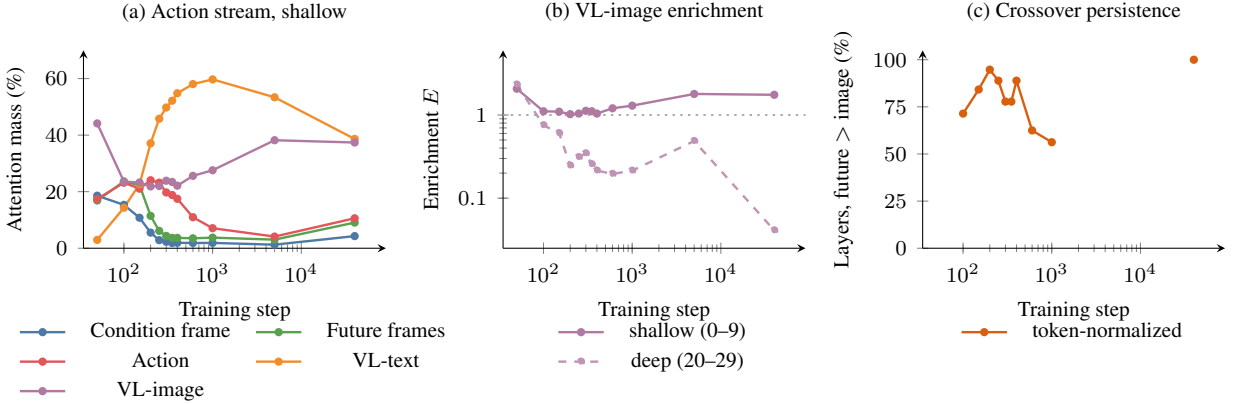

Table~\ref{tab:checkpoint_evolution} reports the checkpoint trajectory of the
asymmetric model over twelve checkpoints, and
Figure~\ref{fig:checkpoint_attention_full} plots it against training step in the
same layout as the bidirectional-mask trajectory of
Figure~\ref{fig:symmetric_dynamics}. The table resolves the concurrent
redistribution of attention from image to text tokens and separates the two
halves of the handoff.

Three phases are visible. At 50 steps, action queries attend to VL-image
tokens at every depth: image enrichment is $2.07$, $1.74$, and $2.33$ across
the three stages, and text accounts for only 2.9\% of shallow attention.
Between 100 and 400 steps the shallow stage rapidly acquires its linguistic
character. Shallow text mass rises from 14.3\% to 54.8\%, while deep image
enrichment falls from $0.75$ to $0.21$. A future-over-image crossover is
already detectable during this phase, but its onset shifts from layer 23 at
100 steps to layers 11--12 thereafter, and its persistence varies from 71\% to
95\%.

The subsequent refinement is not monotonic. At 600 and 1{,}000 steps, stable
crossovers begin at layer 14 but persist over only 62\% and 56\% of the
remaining layers. At 5{,}000 steps, shallow image enrichment rebounds to
$1.79$ and no stable crossover is detected. By 40{,}000 steps, shallow image
enrichment remains high at $1.75$, deep image enrichment falls to $0.04$, and
the layer-14 crossover persists through every subsequent layer. Thus, coarse
semantic-to-action depth organization appears within the first few hundred
steps, whereas the specific shallow-image-to-intermediate-future handoff
fluctuates before stabilizing at convergence. All checkpoints come from the
same thirty-two-GPU run, so this trajectory is not confounded by a change in
global batch size.

\begin{table}[t]
  \caption{\textbf{Checkpoint evolution of the visual handoff under the
  asymmetric mask.} Token-normalized enrichment $E$ of action-query attention to
  VL-image and future-frame tokens by stage, with the shallow-stage raw mass of
  text and image attention. The crossover column gives the first layer at which
  token-normalized future enrichment exceeds image enrichment for three
  consecutive layers, followed in parentheses by the percentage of subsequent
  layers that maintain the ordering; ``---'' marks checkpoints at which no such
  layer exists; this token-normalized criterion can lag the raw-mass crossover
  of Figure~\ref{fig:checkpoint_handoff}. Measured on one representative
  sequence averaged over five denoising steps, using the token counts recorded
  for each checkpoint.}
  \label{tab:checkpoint_evolution}
  \centering
  \small
  \setlength{\tabcolsep}{4pt}
  \begin{tabular}{rcccccccc c}
    \toprule
    & \multicolumn{3}{c}{VL-image $E$}
    & \multicolumn{3}{c}{Future frame $E$}
    & \multicolumn{2}{c}{Shallow mass (\%)} & \\
    \cmidrule(lr){2-4}\cmidrule(lr){5-7}\cmidrule(lr){8-9}
    Step & 0--9 & 10--19 & 20--29 & 0--9 & 10--19 & 20--29
    & Text & Image & Crossover \\
    \midrule
        50 & 2.07 & 1.74 & 2.33 & 0.40 & 0.56 & 0.28 & 2.9 & 44.2 & --- \\
    100 & 1.11 & 0.95 & 0.75 & 0.56 & 0.86 & 0.77 & 14.3 & 23.5 & 23 (71) \\
    150 & 1.10 & 0.61 & 0.61 & 0.53 & 0.89 & 0.81 & 22.6 & 23.3 & 11 (84) \\
    200 & 1.02 & 0.27 & 0.25 & 0.27 & 0.55 & 0.46 & 37.1 & 21.9 & 11 (95) \\
    250 & 1.04 & 0.21 & 0.31 & 0.15 & 0.46 & 0.45 & 45.8 & 22.0 & 12 (89) \\
    300 & 1.12 & 0.28 & 0.35 & 0.10 & 0.39 & 0.38 & 49.8 & 23.9 & 12 (78) \\
    350 & 1.11 & 0.27 & 0.26 & 0.09 & 0.40 & 0.34 & 52.2 & 23.4 & 12 (78) \\
    400 & 1.04 & 0.31 & 0.21 & 0.09 & 0.39 & 0.30 & 54.8 & 22.1 & 12 (89) \\
    600 & 1.21 & 0.50 & 0.20 & 0.08 & 0.35 & 0.24 & 58.1 & 25.6 & 14 (62) \\
    1{,}000 & 1.30 & 0.43 & 0.21 & 0.09 & 0.35 & 0.24 & 59.7 & 27.6 & 14 (56) \\
    5{,}000 & 1.79 & 0.64 & 0.49 & 0.07 & 0.38 & 0.20 & 53.4 & 38.2 & --- \\
    40{,}000 & 1.75 & 0.56 & 0.04 & 0.21 & 0.41 & 0.24 & 38.7 & 37.4 & 14 (100) \\
    \bottomrule

  \end{tabular}
\end{table}

\subsection{Cross-Task Mechanism Replication}
\label{app:cross_task_attention}

\begin{figure*}[!t]
  \centering
  \includegraphics[width=0.82\linewidth]
  {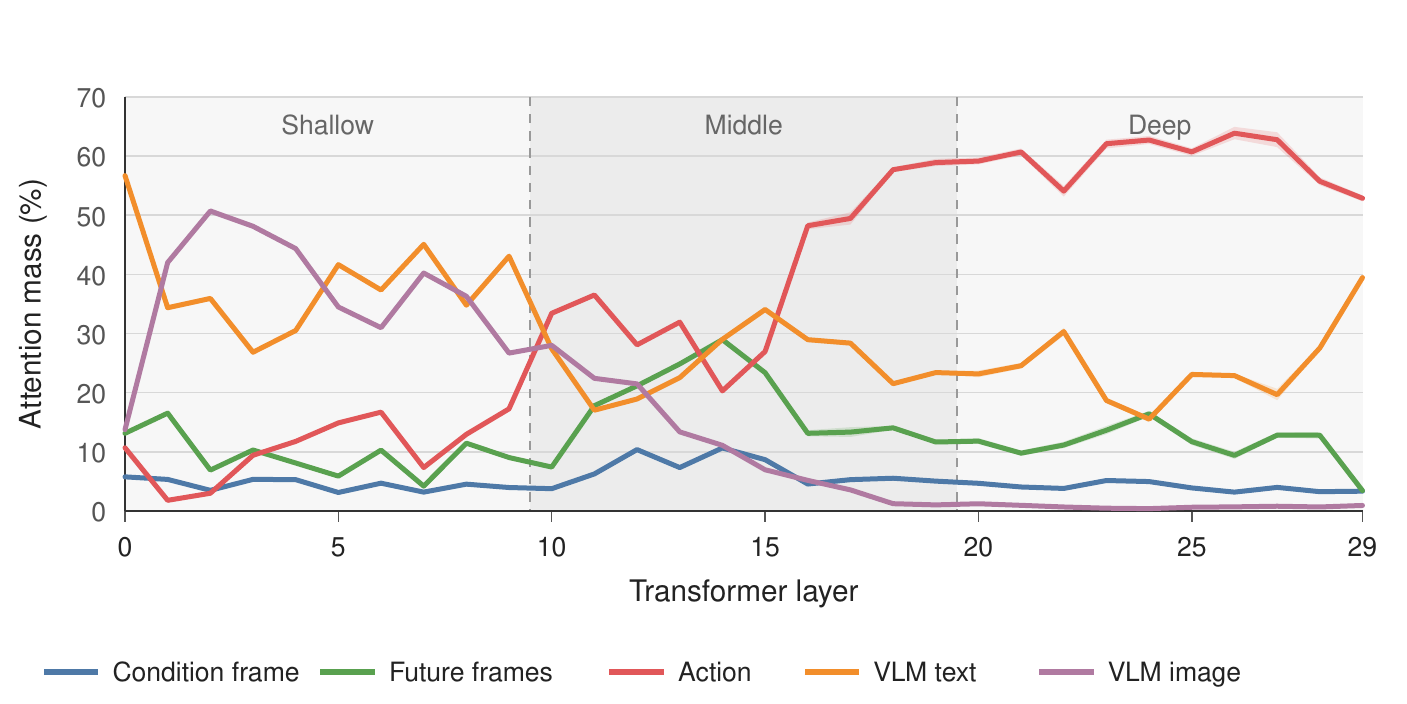}
  \caption{\textbf{Complete raw cross-task action-attention allocation.}
  Attention mass to all five sources is macro-averaged over 50 RoboTwin~2.0
  tasks after averaging five denoising steps within each task; shaded regions
  show 95\% task-bootstrap confidence intervals. Dashed vertical lines mark
  the shared shallow (0--9), intermediate (10--19), and deep (20--29) stages.}
  \label{fig:cross_task_attention_raw}
\end{figure*}

Using the same 50-task macro-averaging protocol and the same robot-pretrained
in-domain checkpoint, we next quantify how often the
depth-wise mechanism occurs. For stage-level allocations, uncertainty is
estimated with 95\% task-bootstrap confidence intervals (CIs), while mechanism
prevalence uses 95\% Wilson CIs. We define the stable crossover as the first of
three consecutive layers for which future-frame attention exceeds VL-image
attention, and summarize its location by the median and interquartile range
(IQR).
Figure~\ref{fig:cross_task_attention_raw} reports the complete raw five-source
routing profile. Table~\ref{tab:cross_task_attention} summarizes the
stage-level attention values, raw and token-normalized mechanism prevalence,
crossover locations, and denoising-step stability.

\begin{table*}[!t]
  \caption{\textbf{Cross-task replication of depth-wise specialization.}
  Stage allocations are macro means with 95\% task-bootstrap confidence
  intervals. Mechanism prevalence is measured across 50 tasks; its uncertainty
  uses 95\% Wilson intervals. Each task contributes one representative
  sequence.}
  \label{tab:cross_task_attention}
  \centering
  \small
  \setlength{\tabcolsep}{5pt}
  \begin{tabular}{lccc}
    \toprule
    \multicolumn{4}{l}{\textit{Stage-level action-attention mass (\%)}} \\
    Stage & VL & Video & Action \\
    \midrule
    Shallow (0--9) &
    75.38 [75.24, 75.51] & 14.05 [13.93, 14.16] & 10.58 [10.54, 10.61] \\
    Middle (10--19) &
    36.55 [36.28, 36.79] & 24.31 [23.90, 24.76] & 39.14 [38.86, 39.41] \\
    Deep (20--29) &
    25.22 [25.02, 25.40] & 15.32 [14.83, 15.84] & 59.46 [59.00, 59.90] \\
    \midrule
    \multicolumn{4}{l}{\textit{Mechanism prevalence and transition location}} \\
    Criterion & Coverage & Estimate & Uncertainty \\
    \midrule
    Shallow VL largest & 50/50 & 100\% & 95\% CI [92.9, 100.0] \\
    Raw image-to-future handoff & 50/50 & 100\% & 95\% CI [92.9, 100.0] \\
    Raw stable crossover & 50/50 & median layer 13 & IQR 12--13 \\
    Normalized image-to-future handoff & 49/50 & 98\% & 95\% CI [89.5, 99.6] \\
    Normalized stable crossover & 50/50 & median layer 14 & IQR 14--14 \\
    Deep action largest & 50/50 & 100\% & 95\% CI [92.9, 100.0] \\
    Denoising-step stability & 50 tasks & mean $r=0.998$ & minimum $r=0.995$ \\
    \bottomrule
  \end{tabular}
\end{table*}

The upper block of Table~\ref{tab:cross_task_attention} quantifies the
three-stage reallocation. VL tokens receive 75.38\% of shallow attention,
compared with 14.05\% for video and 10.58\% for action. The middle stage is
substantially more distributed (36.55\% VL, 24.31\% video, and 39.14\%
action), consistent with a gradual handoff rather than a hard module boundary.
In the deep stage, action attention rises to 59.46\%, while VL and video
attention fall to 25.22\% and 15.32\%, respectively.

The lower block measures how consistently this organization appears across
tasks. All 50 tasks exhibit shallow VL dominance, the intermediate raw
image-to-future handoff, and deep action dominance. The stable raw crossover is tightly
concentrated around layers 12--13. After controlling for each task's token
counts, 49 of 50 tasks still exhibit the handoff over layers 13--15, and all
tasks eventually reach a stable normalized crossover, with a median at layer
14. The mean correlation of 0.998 across denoising-step profiles further shows
that these depth trends are largely invariant to denoising time.

The only task without a late-middle normalized handoff is \textsc{Stamp Seal},
for which future-frame enrichment is slightly below VL-image enrichment when
averaged over layers 13--15 (margin $-0.020$). Its normalized stable crossover
occurs later, at layer 17, and it still exhibits the complete raw
understanding--foresight--action progression.

\FloatBarrier
\section{Attention Routing under a Fully Bidirectional Mask}
\label{app:symmetric_mask}

The asymmetric mask in Eq.~\plaineqref{eq:asymmetric_mask} restricts the two
perceptual streams to their own tokens. To determine what this restriction
contributes, we train a variant in which every stream may attend to every other
stream, so that $A\leftarrow\{V,A,L\}$, $V\leftarrow\{V,A,L\}$, and
$L\leftarrow\{V,A,L\}$, and apply the same depth-wise attention analysis. As for
the asymmetric trajectory of Appendix~\ref{app:checkpoint_evolution}, the variant
is trained without robot pretraining; all other components, objectives, and data
are unchanged. We report two
independently trained runs: a trajectory of checkpoints between 50 and 5{,}000
optimizer steps trained on four GPUs, and a converged 40{,}000-step run trained
on thirty-two GPUs. Because the two runs use different global batch sizes, we
treat the converged run as an independent replication rather than as a
continuation of the trajectory. Both are analysed on a single representative
sequence with ten denoising steps.

\begin{figure}[t]
  \centering
  \includegraphics[width=\linewidth]{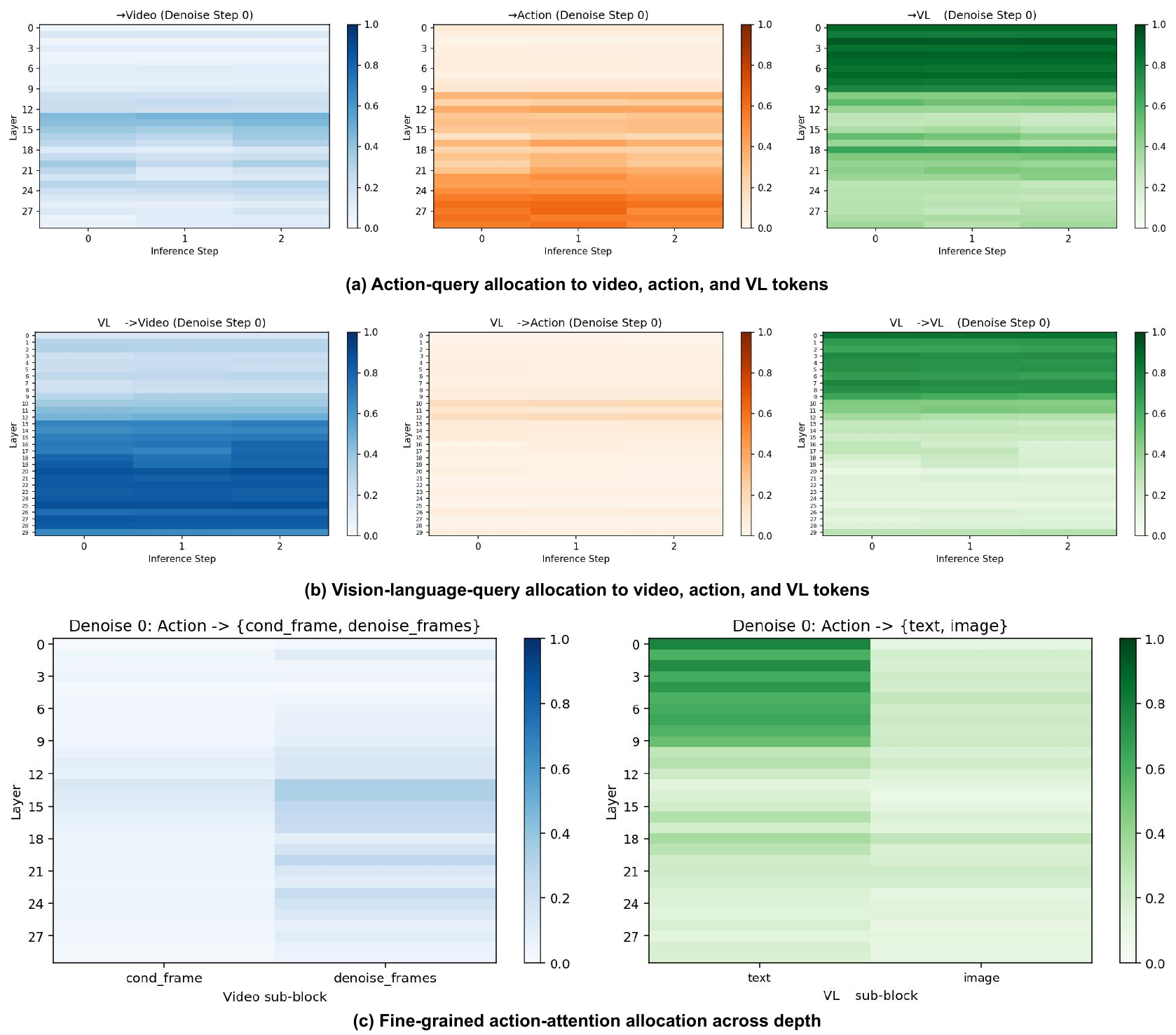}
  \caption{\textbf{Attention allocation under a fully bidirectional mask at
  40{,}000 steps.} Rows are the thirty transformer layers. All panels use an
  absolute colour scale from zero to one, so intensities are comparable across
  panels within this figure. (a) Action-query allocation over the three key
  sources, the bidirectional counterpart of Figure~\ref{fig:motivation}(d).
  (b) Vision-language-query allocation, which has no counterpart in the main text
  because the asymmetric mask fixes this stream to self-attention at every depth.
  (c) Action-query allocation resolved into the two video sub-blocks and the two
  vision-language sub-blocks, the bidirectional counterpart of
  Figure~\ref{fig:depth_submodalities}(a). The vision-language column of (a) is
  dark at every depth and the image column of (c) is uniformly pale, in contrast
  to the shallow-to-deep gradient that the asymmetric mask produces.}
  \label{fig:symmetric_heatmaps}
\end{figure}

\paragraph{The coarse depth organization does not depend on the mask.}
At convergence, the bidirectional variant reproduces the three-stage
action-attention profile reported in Section~\ref{sec:specialization}, which is
visible in the aggregate maps of Figure~\ref{fig:symmetric_heatmaps}(a). VL
tokens receive 85.6\% of shallow action attention, the intermediate stage is
distributed across VL, video, and action tokens (40.6\%, 29.0\%, and 30.5\%),
and action self-attention rises to 49.9\% in the deep stage. The concentration
of predicted-future usage in intermediate layers also recurs: future-frame
enrichment rises from 0.14 in the shallow stage to 0.50 in the middle stage
before falling to 0.31, peaking at layer 14, against a corresponding
0.21--0.41--0.24 profile for the asymmetric model at the same step. At the level of the
three-source aggregate, and on the future-frame side of the visual transition,
the depth-wise organization studied in this paper is therefore not produced by
the asymmetric mask and is not an artifact of one particular choice of
cross-stream connectivity. We regard this as a partial replication that
strengthens the generality of the phenomenon. As in
Section~\ref{sec:specialization}, however, the stage-level aggregate conceals
the sub-modality structure, and it is there that the two configurations diverge.

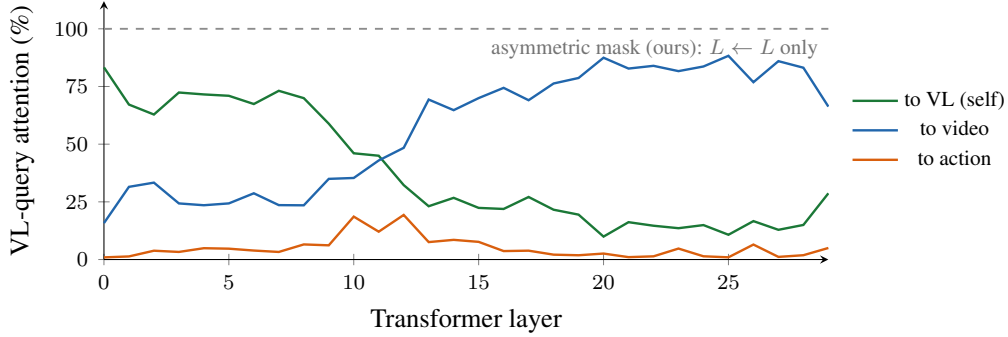
\begin{figure}[t]
  \centering
  \begin{tikzpicture}
\begin{axis}[
  width=0.66\linewidth, height=5.0cm,
  xlabel={Transformer layer}, xmin=0, xmax=29,
  ylabel={VL-query attention (\%)},
  ymin=0, ymax=112, ytick={0,25,50,75,100},
  tick label style={font=\scriptsize}, label style={font=\small},
  axis lines=left,
  every axis plot/.append style={line width=0.9pt},
  legend style={font=\scriptsize, draw=none, fill=none,
                at={(1.02,0.5)}, anchor=west, legend columns=1},
]
  \addplot[black!45, dashed, line width=0.7pt, forget plot]
    coordinates {(0,100) (29,100)};
  \node[font=\scriptsize, text=black!55, anchor=north east] at (axis cs:29,100)
    {asymmetric mask (ours): $L\leftarrow L$ only};
  \addplot[MyGreen] coordinates {(0,83.29) (1,67.17) (2,62.85) (3,72.39) (4,71.56) (5,70.97) (6,67.4) (7,73.13) (8,69.95) (9,58.88) (10,46.05) (11,44.99) (12,32.26) (13,23.1) (14,26.74) (15,22.38) (16,21.92) (17,27.11) (18,21.59) (19,19.46) (20,9.942) (21,16.19) (22,14.65) (23,13.58) (24,14.93) (25,10.74) (26,16.65) (27,12.88) (28,14.99) (29,28.7)};
  \addlegendentry{to VL (self)}
  \addplot[MyBlue] coordinates {(0,15.78) (1,31.5) (2,33.33) (3,24.32) (4,23.53) (5,24.33) (6,28.71) (7,23.58) (8,23.5) (9,34.96) (10,35.35) (11,42.94) (12,48.41) (13,69.34) (14,64.74) (15,69.97) (16,74.41) (17,69.05) (18,76.3) (19,78.69) (20,87.47) (21,82.77) (22,83.96) (23,81.66) (24,83.67) (25,88.3) (26,76.86) (27,85.98) (28,83.14) (29,66.29)};
  \addlegendentry{to video}
  \addplot[MyOrange] coordinates {(0,0.931) (1,1.339) (2,3.821) (3,3.293) (4,4.908) (5,4.703) (6,3.89) (7,3.286) (8,6.544) (9,6.158) (10,18.6) (11,12.07) (12,19.33) (13,7.56) (14,8.526) (15,7.658) (16,3.667) (17,3.838) (18,2.113) (19,1.853) (20,2.588) (21,1.036) (22,1.384) (23,4.754) (24,1.401) (25,0.9631) (26,6.488) (27,1.143) (28,1.871) (29,5.009)};
  \addlegendentry{to action}
\end{axis}
\end{tikzpicture}
  \caption{\textbf{Vision-language-query attention allocation across depth
  under a fully bidirectional mask.} The vision-language stream retains
  self-attention only in shallow layers; from roughly layer 11 onward video
  tokens become its dominant key source, reaching 82.0\% in the deep stage.
  The dashed line marks the asymmetric mask of
  Eq.~\plaineqref{eq:asymmetric_mask}, under which this curve is fixed at
  100\% self-attention by construction. Curves are averaged over ten denoising
  steps at 40{,}000 steps.}
  \label{fig:vlm_query_routing}
\end{figure}

\paragraph{The role of the semantic stream does depend on the mask.}
The two configurations differ in what the vision-language stream itself
computes. Under the asymmetric mask the vision-language stream attends only to
its own tokens at every depth, by construction. Under the bidirectional mask it
does not, as Figure~\ref{fig:vlm_query_routing} and
Figure~\ref{fig:symmetric_heatmaps}(b) show: its shallow layers
retain self-attention, but from roughly layer 11
onward video tokens become the dominant key source, and in the deep stage the
vision-language stream directs 82.0\% of its attention to video tokens while
retaining only 15.3\% on its own. In other words, the semantic stream ceases to
act primarily as a source of semantic keys and becomes a consumer of the
generative stream.

The consequence on the action side is that the visual handoff itself does not
survive, even though its future-frame half does. A handoff requires two things:
an early reliance on the observed image, and its subsequent release in favour of
predicted future frames. Table~\ref{tab:mask_image_enrichment} shows that the
bidirectional variant has neither. Under the asymmetric mask, action queries use
VL-image tokens in a sharply depth-specific way: image enrichment is 1.75 in
the shallow stage, well above the uniform-token reference, and falls to 0.04 in
the deep stage, a shallow-to-deep ratio of roughly 44. The model reads the
current image where it grounds the instruction and gives it up once actions are
being refined. Under the bidirectional mask this differentiation is absent.
Figure~\ref{fig:symmetric_heatmaps}(c) shows the pattern directly: the image
column is pale at every depth, whereas the text column is dark throughout the
shallow stage. Image enrichment is flat across depth (0.81, 0.70, and 0.67 for the three
stages, a ratio of 1.2), never reaches the uniform-token reference at the stage
level, and attains its per-layer maximum at layer 18 rather than in the shallow
stage. Its shallow stage is instead almost purely linguistic, with text
enrichment of 6.29 against image enrichment of 0.81, so the shallow visual
grounding that the asymmetric model performs has no counterpart here.

The bidirectional variant does exhibit layers at which future-frame enrichment
numerically exceeds image enrichment, at layers 13--15. That inequality,
however, holds between two signals that both remain below the uniform-token
reference throughout the network, and it is produced entirely by the rise of
future-frame attention rather than by any release of the observed image: image
enrichment recovers to 0.67 in the deep stage rather than collapsing. We
therefore do not regard it as the same phenomenon. Consistent with this, the
inequality is not sustained. Measuring the fraction of layers at or beyond the
stable crossover for which token-normalized future enrichment exceeds image
enrichment, the bidirectional variant reaches only 24\%, whereas the asymmetric
model maintains the ordering at every layer from the crossover to layer 29.

\begin{table}[t]
  \caption{\textbf{Depth-differentiated use of VL-image tokens under the two
  masks.} Token-normalized enrichment of action-query attention to VL-image
  tokens, by stage. Values above one indicate that a source receives more
  attention than a uniform allocation over all keys would give. Both rows are
  converged 40{,}000-step runs without robot pretraining; asymmetric values are
  the 40{,}000-step row of Table~\ref{tab:checkpoint_evolution}.}
  \label{tab:mask_image_enrichment}
  \centering
  \begin{tabular}{lcccc}
    \toprule
    Mask & Shallow (0--9) & Middle (10--19) & Deep (20--29) & Shallow/Deep \\
    \midrule
    Asymmetric (ours) & 1.75 & 0.56 & 0.04 & 43.8 \\
    Fully bidirectional & 0.81 & 0.70 & 0.67 & 1.2 \\
    \bottomrule
  \end{tabular}
\end{table}

\begin{figure}[t]
  \centering
  \begin{tikzpicture}
\begin{groupplot}[
  group style={group size=3 by 1, horizontal sep=1.55cm},
  width=0.33\linewidth, height=4.2cm,
  xmode=log, xlabel={Training step}, xmin=35, xmax=90000,
  xtick={100,1000,10000}, xticklabels={$10^2$,$10^3$,$10^4$},
  tick label style={font=\scriptsize}, label style={font=\scriptsize},
  title style={font=\scriptsize, yshift=1pt}, axis lines=left,
  every axis plot/.append style={line width=0.9pt},
  legend style={font=\scriptsize, draw=none, fill=none,
                at={(0.5,-0.32)}, anchor=north, legend columns=1},
]
\nextgroupplot[title={(a) VL stream}, ylabel={VL-query attn.\ (\%)}, ymin=0, ymax=100]
  \draw[black!35, dashed] (axis cs:14000,0) -- (axis cs:14000,100);
  \addplot[MyGreen, mark=*, mark size=1.1pt] coordinates {(50,31.19) (100,31.11) (150,29.84) (200,30.61) (250,34.15) (300,36.66) (350,39.58) (400,38.22) (600,41.45) (1000,36.12) (5000,27.02)};
  \addlegendentry{deep: to self}
  \addplot[MyGreen, only marks, mark=star, mark size=3pt, forget plot] coordinates {(40000,15.33)};
  \addplot[MyBlue, mark=*, mark size=1.1pt] coordinates {(50,64.87) (100,65.32) (150,66.02) (200,64.24) (250,59.88) (300,55.94) (350,52.23) (400,52.79) (600,49.41) (1000,57.69) (5000,68.75)};
  \addlegendentry{deep: to video}
  \addplot[MyBlue, only marks, mark=star, mark size=3pt, forget plot] coordinates {(40000,82.01)};
  \addplot[MyGreen!75, dashed, mark=*, mark size=1.1pt] coordinates {(50,27.05) (100,27.83) (150,43.01) (200,60.11) (250,66.41) (300,68.01) (350,67.22) (400,69.79) (600,69.97) (1000,75.59) (5000,83.76)};
  \addlegendentry{shallow: to self}
  \addplot[MyGreen!75, dashed, only marks, mark=star, mark size=3pt, forget plot] coordinates {(40000,69.76)};
  \addplot[MyBlue!75, dashed, mark=*, mark size=1.1pt] coordinates {(50,69.3) (100,67.56) (150,51.23) (200,33.46) (250,26.89) (300,25.21) (350,25.59) (400,23.35) (600,22.98) (1000,20.08) (5000,13.72)};
  \addlegendentry{shallow: to video}
  \addplot[MyBlue!75, dashed, only marks, mark=star, mark size=3pt, forget plot] coordinates {(40000,26.35)};
\nextgroupplot[title={(b) Action stream, shallow}, ylabel={Action-query attn.\ (\%)}, ymin=0, ymax=80]
  \draw[black!35, dashed] (axis cs:14000,0) -- (axis cs:14000,80);
  \addplot[MyPurple, mark=*, mark size=1.1pt] coordinates {(50,5.517) (100,6.869) (150,13.12) (200,27.99) (250,37.7) (300,42.21) (350,46.54) (400,47.89) (600,56.84) (1000,55.21) (5000,51.17)};
  \addlegendentry{to VL-text}
  \addplot[MyPurple, only marks, mark=star, mark size=3pt, forget plot] coordinates {(40000,68.24)};
  \addplot[MyGreen, mark=*, mark size=1.1pt] coordinates {(50,24.01) (100,27.81) (150,19.64) (200,25.2) (250,22.37) (300,23.54) (350,25.22) (400,24.72) (600,23.91) (1000,24.69) (5000,30.43)};
  \addlegendentry{to VL-image}
  \addplot[MyGreen, only marks, mark=star, mark size=3pt, forget plot] coordinates {(40000,17.39)};
  \addplot[MyBlue, mark=*, mark size=1.1pt] coordinates {(50,58.87) (100,54.83) (150,60.9) (200,37.32) (250,24.03) (300,15.97) (350,10.53) (400,10.46) (600,9.248) (1000,7.9) (5000,6.763)};
  \addlegendentry{to video}
  \addplot[MyBlue, only marks, mark=star, mark size=3pt, forget plot] coordinates {(40000,8.38)};
\nextgroupplot[title={(c) Crossover persistence}, ylabel={Layers, future $>$ image (\%)}, ymin=0, ymax=100]
  \draw[black!35, dashed] (axis cs:14000,0) -- (axis cs:14000,100);
  \addplot[MyOrange, mark=*, mark size=1.1pt] coordinates {(50,34.62) (100,35) (150,83.33) (200,61.9) (250,55) (300,60) (350,55) (400,55) (600,45) (1000,31.25) (5000,31.25)};
  \addlegendentry{token-normalized}
  \addplot[MyOrange, only marks, mark=star, mark size=3pt, forget plot] coordinates {(40000,23.53)};
\end{groupplot}
\end{tikzpicture}
  \caption{\textbf{Attention routing under a fully bidirectional mask across
  training.} (a) Vision-language-query allocation, split into self-attention and
  attention to video, for the deep stage (layers 20--29, solid) and the shallow
  stage (layers 0--9, dashed). (b) Shallow-stage action-query allocation.
  (c) Fraction of layers at or beyond the stable crossover for which
  token-normalized future-frame enrichment exceeds VL-image enrichment. Circles
  joined by lines
  are the four-GPU checkpoint trajectory; stars to the right of the dashed rule
  are the independently trained thirty-two-GPU run at 40{,}000 steps, which uses
  a different global batch size and is therefore shown as a separate
  measurement rather than as a continuation of the trajectory.}
  \label{fig:symmetric_dynamics}
\end{figure}
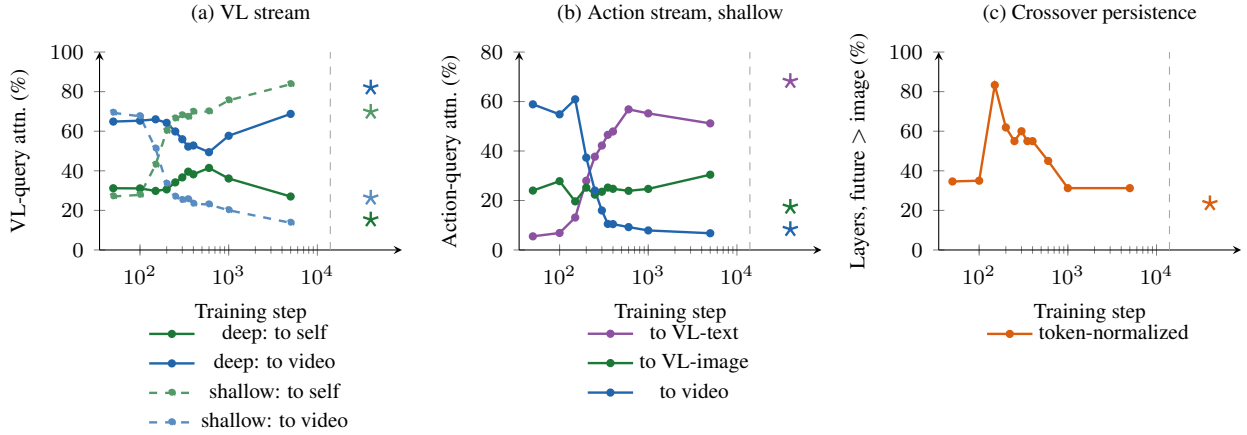

\paragraph{Training dynamics.}
Figure~\ref{fig:symmetric_dynamics} traces these quantities across training.
Three observations follow. First, the vision-language stream moves in opposite
directions at the two ends of the network. Its shallow layers reclaim their own
tokens early, with self-attention rising from 27.1\% at 50 steps to 43.0\% at
150 and 66.4\% at 250. Its deep layers do not: deep self-attention recovers only
to 39.6\% near 350 steps and then declines steadily, reaching 27.0\% at 5{,}000
steps in the four-GPU run and 15.3\% in the independently trained converged run.
Attention to video moves oppositely at the two ends over the same span: the
shallow share falls from 69.3\% to 13.7\% between 50 and 5{,}000 steps and
stands at 26.4\% in the converged run, whereas the deep share rises from 64.9\%
to 68.8\% over the same interval and reaches 82.0\% at convergence. The deep-layer capture is therefore progressive rather than a property of
initialization. Second, the action stream reorganizes over the same short
interval: between 150 and 250 steps, shallow action attention to VL-text rises
from 13.1\% to 37.7\% while shallow action attention to video falls from 60.9\%
to 24.0\%. Because the shallow vision-language recovery occupies exactly this
window, the two streams reorganize together rather than in a resolvable order.
Third, crossover persistence declines monotonically over training, from
83\% at 150 steps to 55\% at 400 steps and 31\% at 5{,}000 steps, with the
converged run at 24\%. Under a bidirectional mask the future-over-image ordering
therefore becomes less rather than more sustained as training proceeds, and does
not consolidate into a handoff.

\paragraph{Scope.}
These measurements characterize where each stream retrieves information, not
what its representations encode. Attention to video keys does not by itself
establish that semantic content has been displaced from the vision-language
representation, since residual connections can carry shallow-layer semantics
through deeper blocks irrespective of where those blocks attend. Establishing
whether the routing change is accompanied by a representational change requires
decoding semantic targets directly from the deep vision-language activations of
both configurations, which we leave to future work. We therefore restrict the
claim to routing: the asymmetric mask is not what produces the coarse depth-wise
organization, but it is what makes the observed image a source that shallow
layers rely on and deeper layers release, and what keeps the vision-language
stream a pure semantic source. It is therefore also what makes token identity an
interpretable indicator of information source in the analysis of
Section~\ref{sec:specialization}.

\FloatBarrier
\section{Experimental Details}
\label{app:simulation}

\subsection{RoboTwin~2.0}

RoboTwin~2.0 contains 50 bimanual manipulation tasks and applies structured
randomization to clutter, illumination, background, tabletop height, and
language~\citep{chen2026robotwin2}. For the standard multi-task experiment,
we evaluate one policy across all tasks in both Clean and Randomized scenes.
Each reported aggregate is the unweighted mean of per-task success rates, with
100 evaluation episodes per task and setting.

For clean-to-random evaluation, no target-embodiment randomized data are used
during downstream adaptation; the adapted policy is evaluated directly in the
fully randomized Hard setting. In contrast, the in-domain results in
Table~\ref{tab:robotwin_indomain} use both clean and randomized data during
downstream adaptation.
Table~\ref{tab:app_robotwin_protocol} summarizes the RoboTwin evaluation
settings detailed above. Table~\ref{tab:app_robotwin_per_task} reports the
complete task-level Clean and Randomized results for the in-domain multi-task
evaluation.

\begin{table}[h]
  \caption{\textbf{RoboTwin evaluation protocols used in this work.}}
  \label{tab:app_robotwin_protocol}
  \centering
  \begin{tabular}{lccc}
    \toprule
    Protocol & Tasks & Evaluation/task & Test domain \\
    \midrule
    Standard Clean & 50 & 100 & Clean \\
    Standard Randomized & 50 & 100 & Randomized \\
    Clean-to-Random & 50 & 100 & Randomized Hard \\
    \bottomrule
  \end{tabular}
\end{table}

\begin{table*}[p]
  \caption{\textbf{RoboTwin~2.0 per-task success rates (\%).}
  Each baseline is the strongest Randomized-setting representative of its
  paradigm.
  EWAM is evaluated on 100 episodes per task and setting.}
  \label{tab:app_robotwin_per_task}
  \centering
  \small
  \renewcommand{\arraystretch}{0.95}
  \begin{tabular}{lcccccccc}
    \toprule
    & \multicolumn{2}{c}{ACE-Ego-0~\citep{added_li2026aceego}}
    & \multicolumn{2}{c}{Fast-WAM~\citep{yuan2026fastwam}}
    & \multicolumn{2}{c}{WLA-0~\citep{added_yang2026wla0}}
    & \multicolumn{2}{c}{EWAM} \\
    \cmidrule(lr){2-3}\cmidrule(lr){4-5}\cmidrule(lr){6-7}\cmidrule(lr){8-9}
    Task & Clean & Rand. & Clean & Rand. & Clean & Rand. & Clean & Rand. \\
    \midrule
    Adjust Bottle & 100 & 100 & 100 & 100 & 100 & 100 & 100 & 100 \\
    Beat Block Hammer & 98 & 92 & 99 & 97 & 95 & 87 & 94 & 95 \\
    Blocks Ranking RGB & 98 & 97 & 100 & 100 & 98 & 98 & 99 & 97 \\
    Blocks Ranking Size & 89 & 91 & 94 & 98 & 93 & 85 & 76 & 85 \\
    Click Alarmclock & 52 & 38 & 100 & 100 & 99 & 100 & 100 & 100 \\
    Click Bell & 66 & 71 & 100 & 100 & 100 & 100 & 100 & 100 \\
    Dump Bin Bigbin & 100 & 97 & 97 & 96 & 90 & 94 & 97 & 97 \\
    Grab Roller & 100 & 100 & 100 & 100 & 100 & 100 & 100 & 100 \\
    Handover Block & 96 & 85 & 95 & 81 & 96 & 87 & 94 & 92 \\
    Handover Mic & 91 & 94 & 99 & 100 & 92 & 93 & 98 & 98 \\
    Hanging Mug & 29 & 31 & 58 & 62 & 69 & 47 & 72 & 69 \\
    Lift Pot & 100 & 100 & 100 & 100 & 100 & 100 & 100 & 99 \\
    Move Can Pot & 100 & 98 & 90 & 88 & 98 & 99 & 96 & 97 \\
    Move Pillbottle Pad & 100 & 100 & 100 & 99 & 100 & 97 & 96 & 99 \\
    Move Playingcard Away & 100 & 98 & 100 & 100 & 99 & 100 & 100 & 100 \\
    Move Stapler Pad & 90 & 89 & 77 & 64 & 92 & 75 & 68 & 66 \\
    Open Laptop & 100 & 98 & 98 & 100 & 99 & 100 & 97 & 99 \\
    Open Microwave & 91 & 85 & 62 & 45 & 97 & 92 & 100 & 100 \\
    Pick Diverse Bottles & 84 & 86 & 80 & 85 & 95 & 79 & 85 & 91 \\
    Pick Dual Bottles & 89 & 88 & 100 & 96 & 100 & 83 & 95 & 89 \\
    Place A2B Left & 95 & 96 & 95 & 93 & 77 & 76 & 95 & 86 \\
    Place A2B Right & 90 & 94 & 93 & 99 & 75 & 75 & 96 & 91 \\
    Place Bread Basket & 92 & 93 & 91 & 93 & 91 & 91 & 95 & 99 \\
    Place Bread Skillet & 94 & 89 & 90 & 93 & 94 & 85 & 92 & 96 \\
    Place Burger Fries & 98 & 100 & 96 & 99 & 95 & 98 & 99 & 97 \\
    Place Can Basket & 78 & 82 & 71 & 69 & 87 & 78 & 82 & 89 \\
    Place Cans Plasticbox & 100 & 98 & 99 & 96 & 100 & 98 & 100 & 100 \\
    Place Container Plate & 98 & 100 & 96 & 100 & 99 & 99 & 99 & 98 \\
    Place Dual Shoes & 95 & 96 & 94 & 88 & 94 & 92 & 95 & 95 \\
    Place Empty Cup & 100 & 100 & 100 & 100 & 99 & 100 & 100 & 99 \\
    Place Fan & 94 & 93 & 96 & 96 & 94 & 94 & 96 & 95 \\
    Place Mouse Pad & 96 & 95 & 83 & 89 & 89 & 88 & 86 & 74 \\
    Place Object Basket & 93 & 89 & 89 & 88 & 82 & 84 & 89 & 83 \\
    Place Object Scale & 95 & 92 & 90 & 97 & 99 & 96 & 93 & 95 \\
    Place Object Stand & 95 & 94 & 90 & 94 & 99 & 92 & 97 & 97 \\
    Place Phone Stand & 91 & 98 & 97 & 99 & 95 & 98 & 94 & 82 \\
    Place Shoe & 100 & 100 & 96 & 99 & 100 & 99 & 98 & 99 \\
    Press Stapler & 98 & 98 & 90 & 97 & 99 & 97 & 94 & 100 \\
    Put Bottles Dustbin & 94 & 93 & 95 & 90 & 89 & 85 & 84 & 81 \\
    Put Object Cabinet & 82 & 79 & 94 & 89 & 82 & 84 & 61 & 71 \\
    Rotate QRcode & 94 & 95 & 93 & 89 & 91 & 91 & 96 & 89 \\
    Scan Object & 95 & 97 & 89 & 92 & 96 & 95 & 76 & 90 \\
    Shake Bottle & 100 & 100 & 100 & 100 & 99 & 100 & 100 & 100 \\
    Shake Bottle Horizontally & 100 & 100 & 100 & 100 & 99 & 100 & 100 & 100 \\
    Stack Blocks Three & 87 & 82 & 95 & 97 & 95 & 91 & 100 & 99 \\
    Stack Blocks Two & 100 & 100 & 100 & 100 & 100 & 100 & 100 & 100 \\
    Stack Bowls Three & 80 & 85 & 80 & 81 & 86 & 84 & 91 & 88 \\
    Stack Bowls Two & 96 & 98 & 92 & 98 & 96 & 99 & 97 & 97 \\
    Stamp Seal & 94 & 100 & 90 & 94 & 95 & 84 & 99 & 99 \\
    Turn Switch & 59 & 57 & 61 & 59 & 39 & 32 & 78 & 78 \\
    \midrule
    Average & 91.12 & 90.62 & 91.88 & 91.78 & 92.94 & 90.02 &
    \textbf{92.98} & \textbf{92.80} \\
    \bottomrule
  \end{tabular}
\end{table*}

\subsection{LIBERO}

LIBERO evaluates knowledge transfer across spatial, object-centric,
goal-conditioned, and long-horizon manipulation suites
~\citep{added_liu2023libero}. We train and evaluate a single EWAM policy under
the same observation and action interface used by the comparison methods. The
primary metric is average task success, reported separately for LIBERO-Spatial,
LIBERO-Object, LIBERO-Goal, and LIBERO-Long in addition to the overall mean.

\subsection{Real-World Evaluation Details}
\label{app:real_world}

\begin{figure}[htbp]
  \centering
  \includegraphics[width=0.90\linewidth]{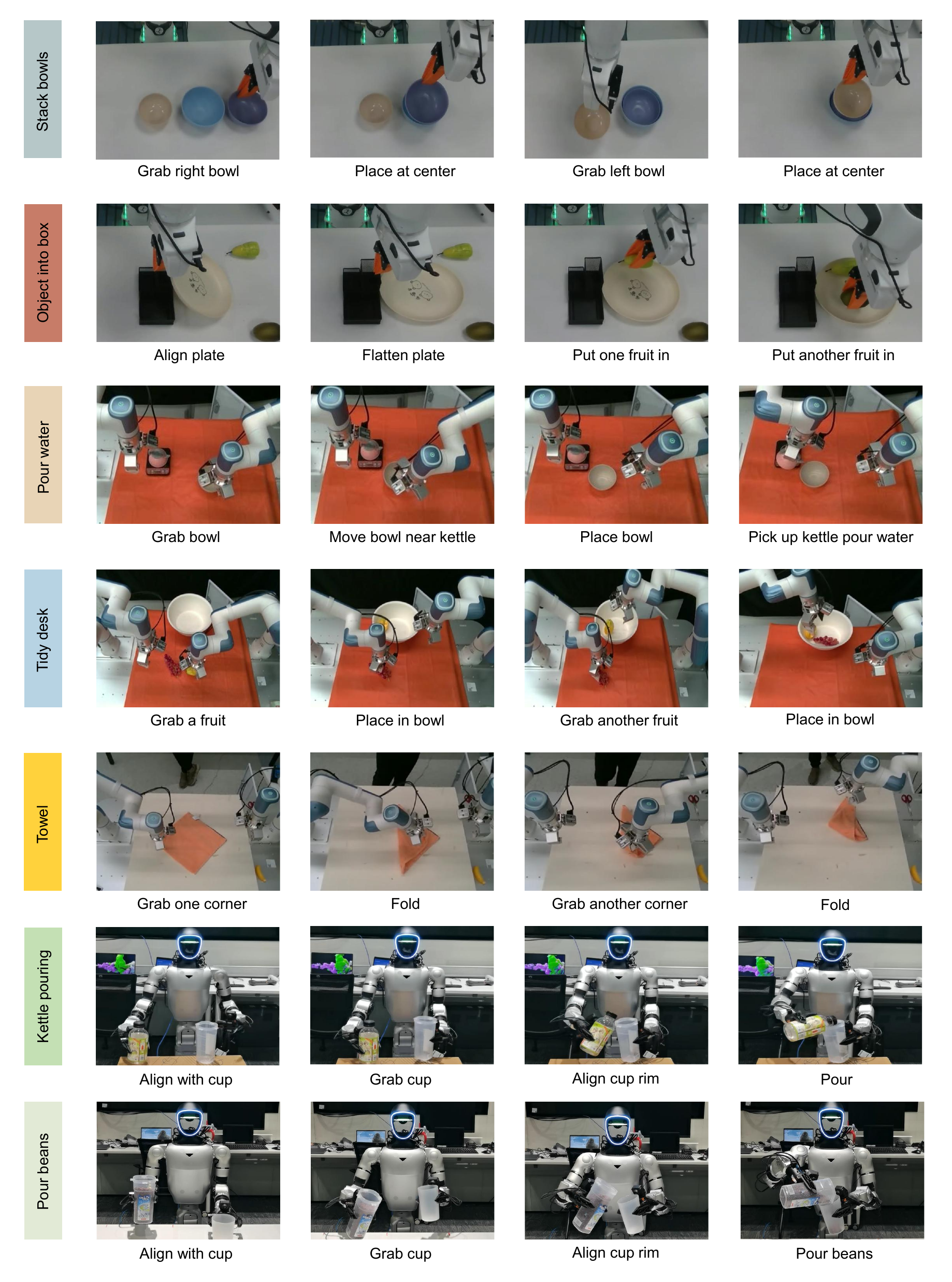}
  \caption{\textbf{Seven real-world manipulation tasks evaluated on Franka,
  Dobot, and Unitree G1-D platforms.}}
  \label{fig:RoboManip}
\end{figure}

We evaluate the same model family on a Franka single-arm platform, a Dobot dual-arm platform, and a Unitree~G1-D humanoid. The selected tasks cover single-arm object rearrangement, bimanual coordination, and contact-rich liquid transfer. Franka evaluation includes stacking bowls and placing an object into a box. Dobot evaluation includes pouring water, tidying a desk, and manipulating a towel. Unitree~G1-D evaluation includes kettle pouring and pouring beans. Together, these tasks test grasp stability, object rearrangement, deformable-object manipulation, bimanual coordination, and controlled pouring. Figure~\ref{fig:RoboManip} illustrates the real-world robot evaluation setup and corresponding tasks. Table~\ref{tab:app_real_protocol} summarizes the embodiments and tasks included in this evaluation.

\begin{table}[h]
  \caption{\textbf{Overview of the seven real-world evaluation tasks.} The
  evaluation spans single-arm, dual-arm, and humanoid dual-arm embodiments.}
  \label{tab:app_real_protocol}
  \centering
  \begin{tabular}{@{}llp{0.48\linewidth}@{}}
    \toprule
    Platform & Embodiment & Tasks \\
    \midrule
    Franka & Single arm & Stack bowls; object into box \\
    Dobot & Dual arm & Pour water; tidy desk; towel manipulation \\
    Unitree G1-D & Humanoid dual arm & Kettle pouring; pour beans \\
    \bottomrule
  \end{tabular}
\end{table}

Every method is evaluated with 10 trials per task on each platform, and all
real-world success rates reported in Tables~\ref{tab:real_robot_franka},
\ref{tab:real_robot_dual}, and~\ref{tab:g1d_cotraining} are computed over those
10 trials. An episode is counted as successful only when the task-level terminal
condition is satisfied, rather than when an intermediate grasp or motion
succeeds. Initial object poses are varied across trials within the reachable
workspace. We use the same task instruction for EWAM and the corresponding
baseline and do not manually intervene after execution begins.

\subsection{Long-Horizon Subtask Evaluation}
\label{app:long_horizon}

This appendix details the two long-horizon evaluations of
Section~\ref{sec:subtask_progress}: three instruction-conditioned block
tasks in simulation, and \textsc{Dual-Basket Sorting} on a Unitree G1-D.

\subsubsection{Instruction-Conditioned Block Tasks}
\label{app:instructed_blocks}

The three tasks used in Section~\ref{sec:subtask_progress} are adapted from
the \textsc{Blocks Ranking} and \textsc{Blocks Stacking} layouts of RoboTwin
and are distinct from the 50-task suite evaluated in
Sections~\ref{sec:robotwin_c2r} and~\ref{sec:robotwin_main}. In the original
designs, three colored blocks (red, green, and blue) are initialized at
random table positions and must be arranged in a fixed red--green--blue
order, from left to right for ranking and from bottom to top for stacking.
Our variants preserve the object set, workspace, and embodiment, and modify
only the language interface. Each instruction specifies a sequence of
local objectives that changes within the episode, so later stages remain
conditioned on completing the currently active subtask.

\paragraph{Instructed Ranking.}
The policy grasps the three blocks in one instructed order and places them
along a left-to-right line in a second instructed order. The two orders
need not coincide. For example, an instruction may require grasping
blue--red--green and placing red--green--blue from left to right.

\paragraph{Instructed Stacking.}
The policy grasps the three blocks in an instructed order and stacks them
bottom to top in that same order. With three blocks this admits only
$3!=6$ instruction variants. Because the stacking order is not specified
independently, an episode that completes the stack necessarily completes
the full task, so Stack and Full coincide in
Table~\ref{tab:subtask_success}. For example, an instruction may require
grasping blue--red--green and stacking blue--red--green from bottom to top.

\paragraph{Instructed Ranking \& Stacking.}
This task composes the two skills within a single episode. The policy first
completes an instructed ranking, grasping in one specified order and
placing from left to right in a second, and then restacks the same blocks
bottom to top in a third instructed order. For example, an instruction may
require grasping blue--red--green, placing red--green--blue from left to
right, and then stacking green--red--blue from bottom to top.

\paragraph{Evaluation protocol.}
Each task is evaluated for 200 episodes. Both variants share the same
unified backbone, pretraining, and post-training data, and differ only in
whether the subtask-phase heads are trained.
Grasp, Place, and Stack are the percentages of episodes that complete the
named stage, with Place and Stack counted only when the currently instructed
target is met; they are computed independently over the same 200 episodes
and are therefore not nested, so a later-stage rate may exceed an
earlier-stage rate. Full is complete-episode
success, counted only when the terminal spatial configuration satisfies
every specified order. Dashes in Table~\ref{tab:subtask_success} mark
stages a task does not define: Stack is absent from
instructed ranking, and Place is absent from
instructed stacking. Because the commanded orders
differ from the fixed convention of the original tasks, success requires
finishing each active subtask and then the remaining sequence, rather
than reproducing a memorized layout.

\subsubsection{Real-World Dual-Basket Sorting}
\label{app:dual_basket}

We evaluate the same with- versus without-supervision pair on a Unitree
G1-D \textsc{Dual-Basket Sorting} sequence. Four objects are placed on the table: a doll, a
marker, a crumpled tissue, and a used water bottle. The policy must place
useful objects (the doll and the marker) in the right basket and waste
(the tissue and the bottle) in the left basket, thereby clearing the
table. We score each trial by the number of objects placed in the
designated basket (out of four); a trial is fully successful only when
all four objects are cleared.

With subtask-phase supervision, ten trials complete
$4,3,0,1,3,3,2,3,1,2$ objects (mean $2.2/4$); one trial clears the table.
After repeated failures on the marker, the policy typically switches to
another remaining object and later returns, rather than persisting on the
same failed grasp. Without this supervision, ten trials complete
$0,1,1,1,2,1,2,1,1,0$ objects (mean $1.0/4$), with no full clearance.
After a failed grasp the policy typically retries the same object and
does not reallocate to an easier remaining target.

Together with the block tasks, these results indicate that subtask-phase
labels supply a current-goal signal. Training the policy to recognize and
complete the active phase increases later-stage completion and
full-sequence success on both the instructed block tasks and the physical
multi-object sequence.

\begin{figure}[h]
  \centering
  \includegraphics[width=0.95\linewidth]{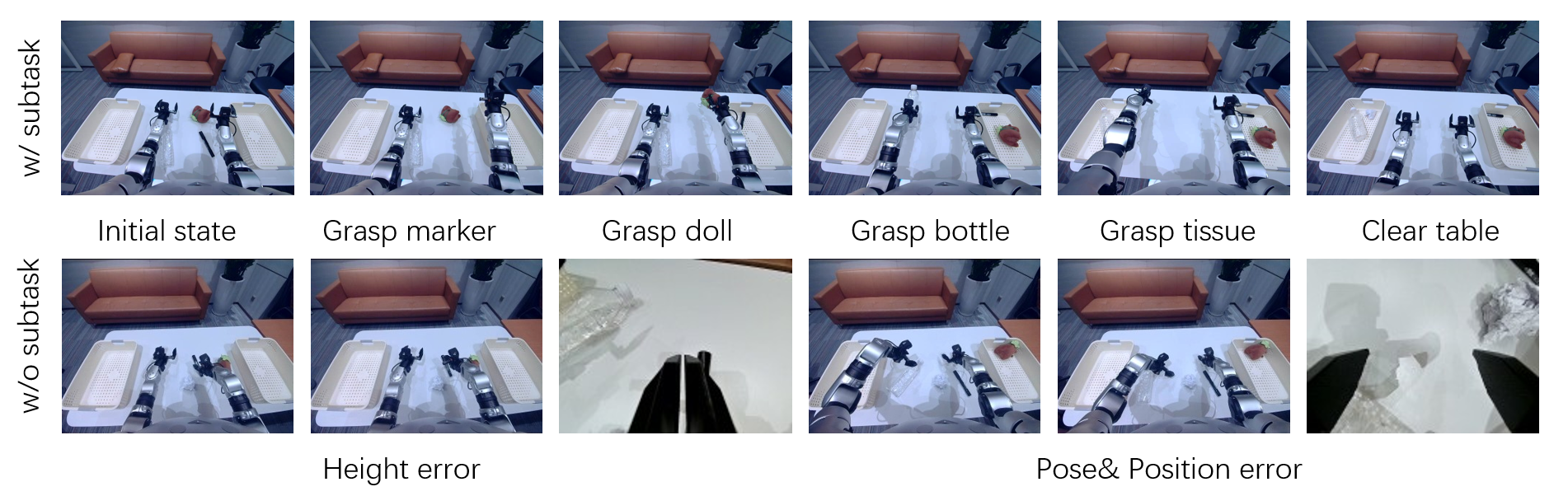}
  \caption{\textbf{Dual-Basket Sorting on Unitree G1-D.}
  Top: a successful rollout with subtask-phase supervision that clears
  the table (initial state, grasp marker, grasp doll, grasp bottle,
  grasp tissue, and clear table).
  Bottom: two failure modes that occur frequently without this
  supervision. The first three frames show grasp height errors; the
  last three show grasp pose-and-position errors. These errors lead to
  repeated attempts on the same object and stalled task progress.}
  \label{fig:dual_basket}
\end{figure}

Figure~\ref{fig:dual_basket} illustrates the behavioral contrast.
With subtask-phase supervision (top), the policy maintains a clear local
objective and clears the table: it successfully grasps each of the four
objects and places them in the designated baskets.
Without this supervision (bottom), rollouts frequently trigger grasp
height errors and grasp pose-and-position errors. The policy then
retries the same failed grasp rather than switching to another object,
so progress stalls and the table is not cleared.

\FloatBarrier
\subsection{Generalization under Tabletop and Cup Changes}
\label{app:ego_generalization}

\paragraph{Evaluation setting.}
This section details the real-robot pouring evaluation summarized in
\textit{Effect of Egocentric Co-Training Data}
(Section~\ref{sec:ego_data_scaling}), which changes table colours or
tablecloths and replaces the pouring cup. All seven configurations omit EWAM robot and human
pretraining and use end-effector (EEF) actions. We vary robot-only
demonstration counts or add egocentric data to 60 robot demonstrations.
Each configuration is evaluated over 10 physical trials, randomly
switching among two to three tablecloth and cup variants. Rates pool
these trials. This evaluation is separate from
Table~\ref{tab:g1d_cotraining}.

\paragraph{Observed results.}
Robot-only training remains at 0--20\% success
(Figure~\ref{fig:ego_scaling}).
With robot demonstrations fixed at 60, adding 1{,}000 egocentric
demonstrations raises success from 10\% to 60\%, a gain of 50 percentage
points. Two findings follow from these results. First, a small amount of egocentric data can substitute for an equal
amount of robot data: 60 robot plus 30 egocentric demonstrations reach
20\%, matching 90 robot-only demonstrations, while egocentric data are
far cheaper to collect. Second, performance continues to scale with
egocentric data: as egocentric demonstrations increase from 30 to 1{,}000,
success rises from 20\% to 30\%, 40\%, and 60\%, with no sign of
saturation in this range. Egocentric data thus
offer a cost-effective way to improve robustness.

\FloatBarrier

\raggedbottom
\FloatBarrier
\section{Inference Latency and Real-Time Deployment}
\label{app:latency}

\paragraph{Inference configuration and timing.}
EWAM generates a chunk of 16 actions, each with 16 dimensions, in 258~ms
on a single RTX~5090 using five denoising (flow-integration) steps, including
chunk-overlap correction (Table~\ref{tab:latency}). Timings are measured
at batch size one in bfloat16 after warm-up, with GPU synchronization
around each inference call. We report the minimum of three repetitions,
which differ by at most 1~ms.

\paragraph{Reducing inference cost.}
We omit VAE decoding of predicted future frames because robot control
requires only the action output. The video expert and the latent future
features used by action tokens remain active. Rewriting rotary position
embeddings to avoid graph breaks enables full-graph compilation of each
denoising step.
At ten denoising steps, these changes reduce chunk latency from
1{,}668 to 461~ms; overlap correction adds 3~ms, giving 464~ms in total
and a $3.6\times$ speedup over the original implementation. With five
steps, the corresponding timings are 254~ms before correction and
258~ms after correction. The reduction in denoising steps changes
the sampling procedure and is separate from the implementation
optimizations. Neither float16 nor the tested FP8 implementations
improved on compiled bfloat16.

\paragraph{Action execution and replanning.}
With a temporal stride of four, a 16-action prediction has a nominal
horizon of $16\times4/30=2.13$~s at a 30~Hz command rate.
The measured 258~ms inference time is approximately 12\% of this horizon,
providing a latency margin relative to the nominal prediction horizon.
Real-time chunking~\citep{black2025rtc} generates a new chunk while the
previous one executes and constrains generation using the remaining
actions to improve continuity. The command execution rate and the
observation-conditioned replanning rate are distinct: the latter depends
on inference latency and the execution schedule.

\paragraph{Comparison with other policies.}
Table~\ref{tab:latency_baselines} compares action-output inference on the
same GPU, with batch size one, bfloat16, and the same action-chunk shape.
Baseline models use randomly initialized weights, so the results
characterize their execution cost under this benchmark configuration.
At five denoising steps, EWAM takes 258~ms versus 901~ms for Motus,
while remaining slower than $\pi_{0.5}$, Xiaomi Robotics-1, and Fast-WAM.

\begin{table}[t]
  \caption{\textbf{Inference latency per action chunk on a single RTX~5090.}
  Each chunk contains 16 actions; step counts refer to denoising.
  The baseline includes future-frame VAE decoding and does not use
  full-graph compilation. Subsequent rows apply the listed changes
  cumulatively.}
  \label{tab:latency}
  \centering
  \small
  \begin{tabular}{lcc}
    \toprule
    Configuration & 10 steps (ms) & 5 steps (ms) \\
    \midrule
    Baseline inference
      & 1668 & -- \\
    Without future-frame VAE decoding
      & 1128 & 578 \\
    \quad + graph-compatible RoPE and full-graph compilation
      & 461 & 254 \\
    \quad + chunk-overlap correction
      & 464 & \textbf{258} \\
    \bottomrule
  \end{tabular}
\end{table}

\begin{table}[t]
  \caption{\textbf{Action-output inference latency under a common
  benchmark configuration.}
  All models use a single RTX~5090, batch size one, bfloat16, and chunks
  of 16 actions with 16 dimensions each. Baseline weights are randomly
  initialized. The final column reports the marginal cost of one
  denoising step.}
  \label{tab:latency_baselines}
  \centering
  \small
  \begin{tabular}{lccc}
    \toprule
    Model & 5 steps (ms) & 10 steps (ms) & Per step (ms) \\
    \midrule
    \multicolumn{4}{l}{\textit{VLA}} \\
    $\pi_{0.5}$~\citep{black2025pi05}
      & 50 & 60 & 1.9 \\
    Xiaomi Robotics-1~\citep{team2026xiaomirobotics1}
      & 186 & 321 & 27.1 \\
    LingBot-VLA-4B~\citep{added_wu2026lingbotvla}
      & 293 & 504 & 42.3 \\
    \midrule
    \multicolumn{4}{l}{\textit{WAM}} \\
    Fast-WAM~\citep{yuan2026fastwam}
      & 216 & 401 & 37.1 \\
    \midrule
    \multicolumn{4}{l}{\textit{VLA+WAM}} \\
    Motus~\citep{bi2026motus}
      & 901 & 1901 & 200.0 \\
    \textbf{EWAM (Ours)}
      & 258 & 464 & 41.2 \\
    \bottomrule
  \end{tabular}
\end{table}

\section{Pretraining Data}
\label{app:data_compute}
This section describes the data used in EWAM's two separate pretraining regimes, with further details provided in Sections~\ref{app:human_data} and~\ref{app:robot_data}. The human pretraining regime uses approximately 2,084 hours of egocentric video, in which wrist motion supplies the action target. Independently, the robot pretraining regime uses 300K cross‑embodiment trajectories, comprising approximately 1,800 hours of curated robotic manipulation data collected across multiple robot embodiments, and provides synchronized observations, language instructions, robot states, actions, and future visual observations. In each regime, heterogeneous state and action spaces are mapped into a shared padded representation, while validity masks ensure that unavailable dimensions do not contribute to the action loss.

\subsection{Human Egocentric Data}
\label{app:human_data}
The human stage of Section~\ref{sec:human_pretraining} draws on three egocentric corpora summarized in Table~\ref{tab:human_data}. Combined, they provide approximately 2,084 hours of first-person manipulation video paired with per-frame hand pose and camera geometry. The datasets offer complementary coverage: \textbf{VITRA‑1M}~\citep{added_li2025vitra}, compiled from unconstrained internet footage, provides the broadest diversity of objects, environments, and activities; \textbf{EgoDex}~\citep{added_hoque2025egodex} contains scripted tabletop manipulation with high-quality on-device hand tracking; \textbf{Xperience}~\citep{added_ropedia2026xperience} supplies SLAM-registered recordings captured via a head-mounted stereo rig.

\begin{table}[!htbp]
  \caption{\textbf{Human egocentric corpora used for human pretraining.} The last column names the quantity from which each source's anchor camera pose is obtained; all three then enter the identical computation of Eq.~\plaineqref{eq:camera_delta}. VITRA‑1M is used through four of its five subsets; its EgoExo4D subset is excluded.}
  \label{tab:human_data}
  \centering
  \small
  \setlength{\tabcolsep}{5pt}
  \begin{tabular}{llrl}
    \toprule
    Corpus & Source material & Hours & Anchor camera pose from \\
    \midrule
    VITRA‑1M~\citep{added_li2025vitra} & Ego4D, EPIC‑Kitchens, SSv2 & 283 & Inverted extrinsics \\
    EgoDex~\citep{added_hoque2025egodex} & Apple Vision Pro & 829 & Camera‑to‑world transform \\
    Xperience~\citep{added_ropedia2026xperience} & Head‑mounted stereo rig & 972 & SLAM pose and calibration \\
    \midrule
    Total & & 2,084 & \\
    \bottomrule
  \end{tabular}
\end{table}

\textbf{VITRA‑1M} provides MANO‑based hand reconstructions together with corrected camera intrinsics and extrinsics for episodes derived from Ego4D, EPIC‑Kitchens, and Something‑Something~v2~\citep{added_grauman2022ego4d,added_damen2018epickitchens,added_goyal2017something}. We use its \texttt{ego4d\_cooking\_and\_cleaning}, \texttt{ego4d\_other}, \texttt{epic}, and \texttt{ssv2} subsets, retaining 1{,}155{,}865 episodes; the EgoExo4D subset is excluded. Frames flagged as untracked by the released per‑hand masks are treated as invalid. \textbf{EgoDex} contributes 338{,}000 demonstrations across 194 tabletop tasks recorded at 30~Hz, with per‑frame hand tracking confidences; we discard hand observations whose confidence falls below 0.3. \textbf{Xperience} is a partially released corpus recorded with a head‑mounted stereo rig, for which hand poses are stored in the per‑frame rectified stereo‑left camera frame and must first be lifted to world coordinates using the SLAM trajectory before the shared computation applies. Episodes lacking a SLAM pose, lacking stereo calibration, or shorter than one training window are dropped at indexing time.

Every retained episode is reduced to world‑frame wrist poses plus an anchor camera pose and then converted by Eq.~\plaineqref{eq:camera_delta}, so the sixteen action dimensions carry identical physical meaning across the mixture. Sampling is weighted in proportion to the number of indexed windows in each corpus. Because the targets are camera‑relative displacements whose scale is already comparable across sources, no action normalization is applied during this stage.

\subsection{Robotic Manipulation Data}
\label{app:robot_data}
\begin{figure}[!htbp]
  \centering
  \includegraphics[width=0.83\linewidth]{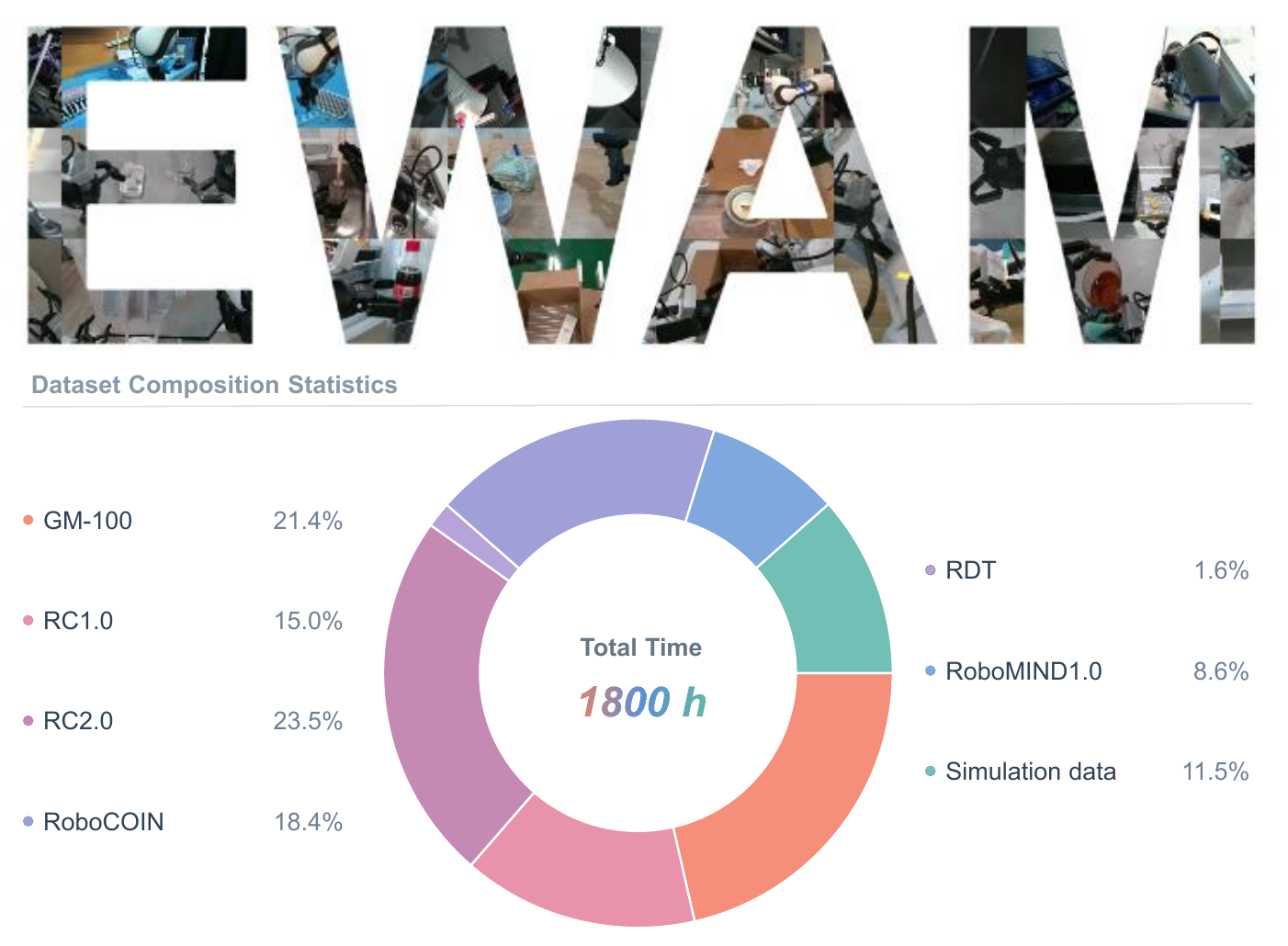}
  \caption{\textbf{Source composition and retained duration distribution of the high‑quality robotic manipulation data used for EWAM pretraining.}}
  \label{fig:RoboData}
\end{figure}

During pretraining, EWAM leverages multiple open-source robotic manipulation datasets containing approximately 3,250 hours of raw recordings, including \textbf{GM-100}~\citep{yu2026gm100}, \textbf{RoboCOIN}~\citep{wu2025robocoin}, \textbf{RoboMIND1.0}~\citep{wu2025robomind}, \textbf{RoboChallenge1.0}~\citep{yakefu2025robochallenge}, \textbf{RoboChallenge2.0}~\citep{ma2026table30}, \textbf{RDT}~\citep{liu2024rdt}, and \textbf{Simulation data}~\citep{chen2026robotwin2,hou2025robomind2,wu2025robomind}.
To focus on tabletop manipulation, we apply task‑relevance filtering, data cleaning, and quality assessment. After removing invalid, low‑quality, and task‑irrelevant samples, we obtain approximately 1,800 hours of high‑quality pretraining data. Figure~\ref{fig:RoboData} shows the retained duration and corresponding proportion of each source dataset.

\begin{figure}[!htbp]
  \centering
  \includegraphics[width=0.89\linewidth]{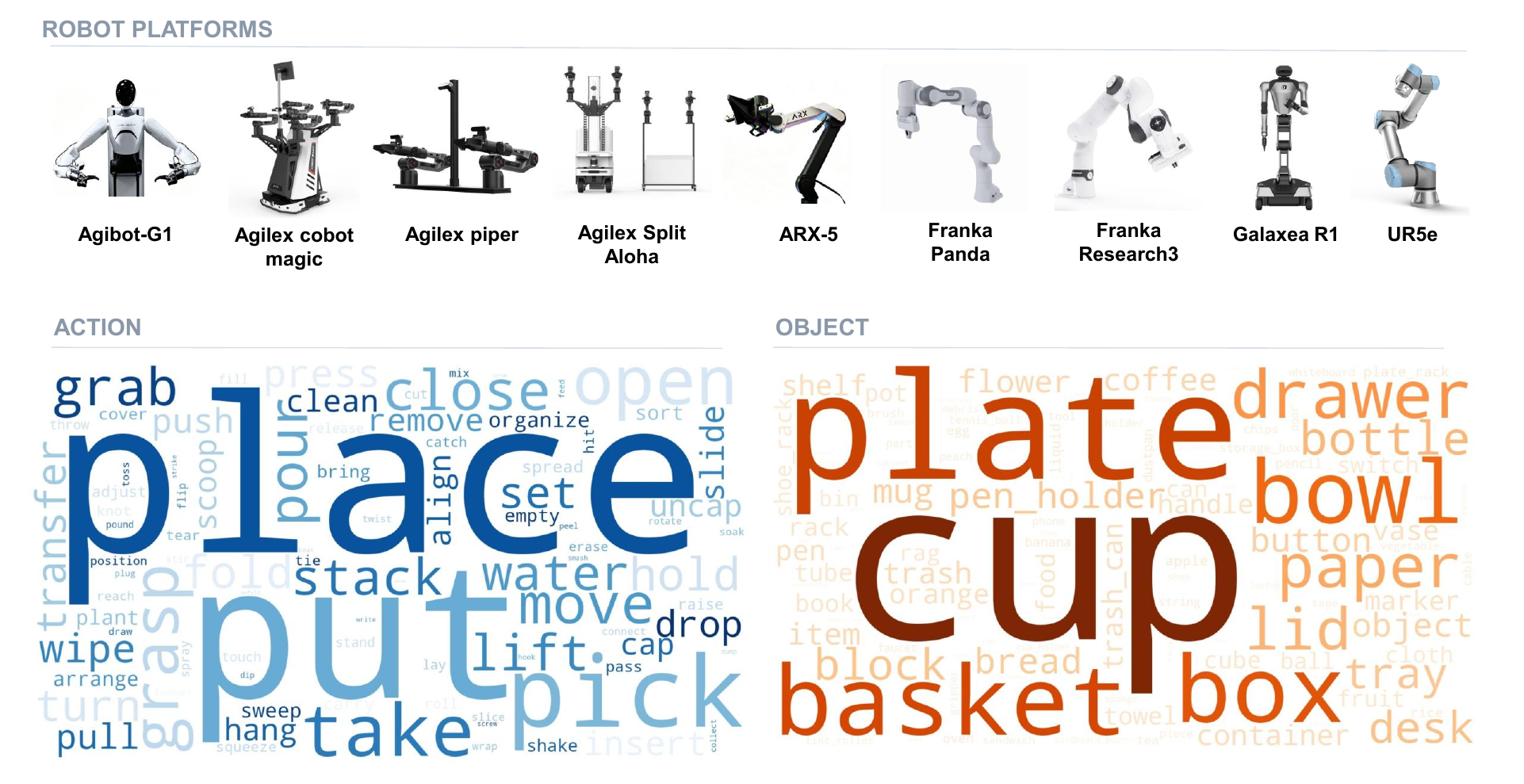}
  \caption{\textbf{Robot platforms, manipulation tasks, and interacted objects covered by the EWAM pretraining data.}}
  \label{fig:RoboPlat}
\end{figure}

The pretraining data of EWAM cover a wide range of robot platforms and manipulation tasks. Substantial disparities exist among robots in terms of embodiment, degrees of freedom, and manipulation approaches. The data encompass representative manipulation tasks, including grasping, placing, moving, organizing, switch operation, and cleaning, together with a wide variety of interactive objects. As shown in Figure~\ref{fig:RoboPlat}, the composition of robot platforms and the keyword distributions of manipulation tasks and interactive objects are presented.

To systematically characterize the composition of the pretraining data, we summarize the kinematic joint dimensions across various robot platforms in Table~\ref{tab:Robo_platform}. The reported statistics correspond to joint-state dimensionality. As several platforms support both single-arm and bimanual setups, joint dimensions differ accordingly.

\begin{table}[!htbp]
  \centering
  \caption{\textbf{Joint-state dimensions by robot platform.}}
  \label{tab:Robo_platform}
  \small
  \begin{tabular}{@{}llc@{}}
    \toprule
    Mode & Robot Platform & Joint Dim. \\
    \midrule
    \multirow{3}{*}{Single Arm}
    & Franka  & 8  \\
    & UR5     & 7  \\
    & ARX-5   & 7  \\
    \midrule
    \multirow{8}{*}{Dual Arm}
    & Franka     & 16  \\
    & UR5        & 14  \\
    & Agilex     & 14  \\
    & AgiBot G1  & 16  \\
    & Galaxea R1 & 16  \\
    & ALOHA      & 14  \\
    & ARX-5      & 14  \\
    & DOS-W1     & 14  \\
    \bottomrule
  \end{tabular}
\end{table}

During data processing, we find substantial differences among the open‑source datasets in terms of acquisition devices, data formats, annotation protocols, and data quality. Common defects include corrupted videos, black frames, severe occlusion, prolonged inactivity, and missing frames. Certain episodes also exhibit state‑action misalignment, abrupt signal changes, and numerical outliers. These issues may disrupt the temporal continuity of action sequences and the logical consistency between states and actions, thereby impairing model pretraining.

According to the overall quality inspection results, most datasets exhibit varying degrees of state‑action misalignment, abrupt numerical changes, and anomalous values. Numerical outliers have the most detrimental effect on model optimization. After the model has fitted the normal data, anomalous samples substantially increase the difficulty of optimization, causing the loss to rise sharply and potentially preventing training from converging. Several datasets contain a substantial number of anomalies, including extreme outliers with magnitudes as high as tens of billions. Regarding state‑action alignment, some samples exhibit state signals that precede action signals or motion trends that are inconsistent between the two. Further analysis indicates that these problems typically arise from inconsistent sampling frequencies or temporal synchronization errors among different signals. Moreover, some episodes contain prolonged static segments, predominantly near their beginning and end, with occasional occurrences during task execution. Mid‑trajectory inactivity often indicates hesitation or pauses during data collection, potentially encouraging the model to learn uninformative or suboptimal action patterns.

To address these issues, we design a two‑stage data cleaning and quality inspection pipeline. The first stage detects file corruption, black frames, occlusions, static frames, and state‑action misalignment, and removes unqualified data at the trajectory level. The second stage detects abrupt numerical changes and anomalous values, and removes abnormal data at the frame level. This pipeline adopts a coarse‑to‑fine filtering strategy that preserves the temporal continuity and state‑action consistency of the cleaned data while maintaining a stable data retention rate.

Overall, systematic curation, format standardization, and two‑stage quality assurance yield approximately 1,800 hours of high‑quality robotic manipulation data for pretraining EWAM on tabletop manipulation. Spanning diverse robot platforms, objects, and task categories, the resulting corpus combines reliable state‑action synchronization with broad behavioral coverage. It therefore provides a solid foundation for learning generalizable visuomotor representations and supports transfer to downstream tasks.

\FloatBarrier
\section{Egocentric Reconstruction and Retargeting Details}
\label{app:ego_geometry}

This appendix gives the coordinate definitions and implementation details for
the demonstration construction in Section~\ref{sec:ego_collection}.

\paragraph{Hand representation and interaction reference.}
Let $s\in\{\mathrm{L},\mathrm{R}\}$ index the hands, $W$ the reconstruction
world frame, $H_s$ the wrist frame, and $G$ the robot base frame. The recovered
wrist pose is
$\mathbf T^W_{H_s}(t)=[\mathbf R^W_{H_s}(t),\mathbf p^W_{H_s}(t)]$;
the articulated joints $\mathbf J^{H_s}(t)$ are expressed locally. A local
joint is placed in the world as
$\mathbf j^W=\mathbf R^W_{H_s}\mathbf j^{H_s}+\mathbf p^W_{H_s}$.
We estimate a fixed local interaction offset $\bar{\mathbf c}_s^H$ as the
median thumb--index midpoint in valid closed-hand frames, giving
\begin{equation}
  \mathbf c_s^W(t)
  =\mathbf R^W_{H_s}(t)\bar{\mathbf c}_s^H
   +\mathbf p^W_{H_s}(t).
  \label{eq:ego_interaction_reference}
\end{equation}
where $t$ is the frame index and $\mathbf c_s^W(t)$ is the interaction
reference point of hand $s$ in the world frame.
The point follows wrist translation and rotation while remaining independent
of instantaneous finger articulation. Its use assumes that one wrist-local
interaction reference is appropriate for the episode.

\paragraph{Shared robot-frame mapping.}
One episode-level rigid transform
$(\mathbf C_{G\leftarrow W},\mathbf b_{G\leftarrow W})$ places both hands
in the robot workspace. The tool targets are
\begin{equation}
\begin{aligned}
 \mathbf p^{\star}_{t,s}
 &=\mathbf C_{G\leftarrow W}\mathbf c_s^W(t)
   +\mathbf b_{G\leftarrow W},\\
 \mathbf R^{\star}_{t,s}
 &=\mathbf C_{G\leftarrow W}\mathbf R^W_{H_s}(t)\mathbf B_s.
\end{aligned}
\label{eq:human_robot_target}
\end{equation}
where $\mathbf p^{\star}_{t,s}$ and $\mathbf R^{\star}_{t,s}$ are the target
TCP position and orientation of hand $s$ at frame $t$ in the robot base frame.
Here $\mathbf C_{G\leftarrow W}$ aligns the axes and
$\mathbf b_{G\leftarrow W}$ sets the placement using a real-robot reference
trajectory. The transform is shared across hands and has unit scale.
The fixed basis $\mathbf B_s$ maps the wrist orientation to the tool
orientation using an episode-consistent closing axis and an orthogonalized
approach axis. The targets refer to the nominal pinch-center TCP and are
held fixed during joint optimization.

\paragraph{Sequence optimization.}
Let $Q=\{\mathbf q_t\}_{t=0}^{T-1}$, with
$\mathbf q_t\in\mathbb R^{14}$ containing seven joints per arm. The
retargeter seeks to minimize
\begin{equation}
 \underset{Q\in\mathcal F_{\mathrm{solve}}}{\operatorname{minimize}}\;
 \sum_{t=0}^{T-1}
 \left[
 \lambda_{\mathrm{task}}E_{\mathrm{task}}(\mathbf q_t)
 +\lambda_{\mathrm{post}}E_{\mathrm{post}}(\mathbf q_t)
 \right]
 +\lambda_{\mathrm{temp}}E_{\mathrm{temp}}(Q).
\label{eq:joint_refinement_objective}
\end{equation}
where $T$ is the number of frames and $\lambda_{\mathrm{task}}$,
$\lambda_{\mathrm{post}}$, and $\lambda_{\mathrm{temp}}$ are the weights of
the corresponding terms.
The task term $E_{\mathrm{task}}$ groups TCP position error and closing- and approach-axis
errors outside phase-dependent tolerances. The posture term $E_{\mathrm{post}}$ regularizes
redundant shoulder--elbow and wrist configurations around a real-robot
reference posture. The temporal term $E_{\mathrm{temp}}$ penalizes velocity, acceleration, and
jerk through finite differences on the original 30~Hz timeline.
$\mathcal F_{\mathrm{solve}}$ denotes the constraints enforced during the
solve, including joint, velocity, and acceleration bounds. Workspace and
collision conditions are also checked on the returned trajectory. This is a
grouped description of the objective; the returned reference need not be a
global minimizer. Output admissibility is reported separately in
Section~\ref{sec:ego_pipeline_evaluation}.

\paragraph{Visual tracking and temporal processing.}
The visual branch applies WiLoR hand detection~\citep{added_potamias2025wilor}
and ByteTrack association~\citep{added_zhang2022bytetrack} to fisheye-rectified
images. We retain detections for each anatomical hand, reject tracks dominated
by image boundaries, and
interpolate short detection gaps only when the endpoint detections have
sufficient overlap. The resulting contiguous clips are supplied to HaWoR.
Short, bounded gaps in wrist-local articulation are interpolated, and valid
articulation segments are smoothed with a Savitzky--Golay
filter~\citep{added_savitzky1964smoothing}. World-frame wrist poses, wrist-local
geometry, and reconstruction-source and validity indicators are retained.

\paragraph{Staged optimization and initialization.}
We solve sparse quadratic subproblems within a trust region after local
linearization, using multiple initial inverse-kinematics candidates and staged
warm starts. The selected refinement first regularizes the full trajectory and then
increases position-tracking priority. A final prefix refinement frees the
initial joint pose and uses a stable later segment to initialize the first
1.2~s, keeping the remainder fixed. This avoids locking the sequence to an
unfavorable first-frame solution. Later frames are available because the
procedure operates offline. Throughout refinement, the task-space targets,
gripper commands, and original timeline are held fixed.

\paragraph{Articulated-hand adaptation.}
For articulated robot hands, an additional branch aligns palm frames and hand
scales, then solves fingertip-position inverse kinematics with joint limits and
temporal regularization. The G1-D two-finger interface instead uses the inferred
binary gripper commands alongside the arm trajectory.

\paragraph{Output audit.}
Before export, we check finite values, joint and derivative limits, workspace
conditions, collision clearance at recorded and interpolated configurations,
FK consistency, and preservation of target, gripper, and time signals. The
exported trajectories are reference actions, distinct from controller commands
and hardware measurements. Grasp states are inferred from image-space evidence
and do not measure contact forces. Mechanical admissibility is evaluated under
the tested robot model and sampled configurations, separately from physical
task performance.

\end{document}